\documentclass{article}

\PassOptionsToPackage{compress}{natbib}
\usepackage[final]{corl_2022} 

\usepackage{textcomp}

\usepackage[utf8]{inputenc} 
\usepackage[T1]{fontenc}    
\usepackage{hyperref}       
\usepackage{url}            
\usepackage{booktabs}       
\usepackage{amsfonts}       
\usepackage{nicefrac}       
\usepackage{microtype}      
\usepackage{colortbl}
\usepackage{placeins}

\usepackage{tikz}
\usetikzlibrary{positioning}
\usetikzlibrary{calc}
\usepackage{wrapfig}

\usepackage{times}
\usepackage{epsfig}
\usepackage{graphicx}

\usepackage{amsmath}
\usepackage{amssymb}
\usepackage{subcaption}
\usepackage{listings}
\usepackage{siunitx}
\usepackage{enumitem}
\usepackage{multirow}
\usepackage{xspace}
\usepackage{float}
\usepackage[export]{adjustbox}
\usepackage{longtable}
\usepackage{afterpage}

\usepackage{geometry}
\usepackage{pdflscape}

\usepackage{amssymb}
\usepackage{pifont}
\usepackage{fontawesome}

\newcommand\blfootnote[1]{%
  \begingroup
  \renewcommand\thefootnote{}\footnote{#1}%
  \addtocounter{footnote}{-1}%
  \endgroup
}
\usepackage{lipsum}

\usepackage{adjustbox}
\usepackage{array}
\newcolumntype{R}[2]{%
    >{\adjustbox{angle=#1,lap=\width-(#2)}\bgroup}%
    l%
    <{\egroup}%
}
\newcommand*\rot{\multicolumn{1}{R{30}{1em}}}

\usepackage{rotating}
\definecolor{forestgreen}{rgb}{0.13, 0.55, 0.13}
\definecolor{indiagreen}{rgb}{0.07, 0.53, 0.03}

\usepackage[small]{titlesec}
\titlespacing*{\section}{0pt}{4pt plus 2pt minus 2pt}{2pt plus 1pt minus 1pt}
\titlespacing*{\subsection}{0pt}{1pt plus 1pt minus 1pt}{0pt plus 1pt minus 1pt}
\titlespacing*{\paragraph}{0pt}{0pt plus 0pt minus 2pt}{5pt}

\newbox{\bigpicturebox}

\definecolor{MyDarkBlue}{rgb}{0,0.08,1}
\definecolor{MyDarkGreen}{rgb}{0.02,0.6,0.02}
\definecolor{MyDarkRed}{rgb}{0.8,0.02,0.02}
\definecolor{MyDarkOrange}{rgb}{0.40,0.2,0.02}
\definecolor{MyPurple}{RGB}{111,0,255}
\definecolor{MyRed}{rgb}{1.0,0.0,0.0}
\definecolor{MyGold}{rgb}{0.75,0.6,0.12}
\definecolor{MyDarkgray}{rgb}{0.66, 0.66, 0.66}

\newcommand{\benchmark}{BEHAVIOR-1K\xspace}
\newcommand{\oldbenchmark}{BEHAVIOR-100\xspace}
\newcommand{\dataset}{\textsc{BEHAVIOR-1K Dataset}\xspace}
\newcommand{\simulator}{\textsc{OmniGibson}\xspace}
\newcommand{\bddlfull}{BEHAVIOR Domain Definition Language\xspace} 
\newcommand{\bddl}{BDDL\xspace} 
\newcommand{\rlvmc}{RL-VMC\xspace}
\newcommand{\rlprim}{RL-Prim.\xspace}
\newcommand{\rlprimhist}{RL-Prim.\-Hist.\xspace}

\newcommand{\revision}[1]{\textcolor{black}{#1}}

\usepackage[resetlabels,labeled]{multibib}
\newcites{A}{Appendix References}

\usepackage{xparse}
\let\oriCiteA\citeA

\RenewDocumentCommand{\citeA}{O{} O{} m}{%
  \renewcommand{\citenumfont}[1]{A##1}%
  \oriCiteA[#1][#2]{#3}%
  \renewcommand{\citenumfont}[1]{##1}%
}

\title{\benchmark: A Human-Centered, Embodied AI Benchmark with 1,000 Everyday Activities and Realistic Simulation}

\author{%
Chengshu Li*$^{\scriptscriptstyle 1}$,
Ruohan Zhang*$^{\scriptscriptstyle 1}$,
Josiah Wong*$^{\scriptscriptstyle 2}$,
\\\textbf{
Cem Gokmen*$^{\scriptscriptstyle 1}$,
Sanjana Srivastava*$^{\scriptscriptstyle 1}$,
Roberto Martín-Martín*$^{\scriptscriptstyle 9}$,
}
\\\textbf{
Chen Wang*$^{\scriptscriptstyle 1}$,
Gabrael Levine*$^{\scriptscriptstyle 1}$,
Wensi Ai*$^{\scriptscriptstyle 1}$,
}
\\\textbf{
Benjamin Martinez$^{\scriptscriptstyle 1}$,
Hang Yin$^{\scriptscriptstyle 1}$,
Michael Lingelbach$^{\scriptscriptstyle 3}$,
Minjune Hwang$^{\scriptscriptstyle 1}$,
}
\\\textbf{
Ayano Hiranaka$^{\scriptscriptstyle 1}$,
Sujay Garlanka$^{\scriptscriptstyle 11}$,
Arman Aydin$^{\scriptscriptstyle 1}$,
Sharon Lee$^{\scriptscriptstyle 1}$,
}
\\\textbf{
Jiankai Sun$^{\scriptscriptstyle 4}$, 
Mona Anvari$^{\scriptscriptstyle 1}$,
Manasi Sharma$^{\scriptscriptstyle 1}$,
Dhruva Bansal$^{\scriptscriptstyle 1}$,
}
\\\textbf{
Samuel Hunter$^{\scriptscriptstyle 1}$,
Kyu-Young Kim$^{\scriptscriptstyle 1}$,
Alan Lou$^{\scriptscriptstyle 5}$,
Caleb R Matthews$^{\scriptscriptstyle 1}$,
}
\\\textbf{
Ivan Villa-Renteria$^{\scriptscriptstyle 1}$, 
Jerry Huayang Tang$^{\scriptscriptstyle 1}$,
Claire Tang$^{\scriptscriptstyle 1}$,
Fei Xia$^{\scriptscriptstyle 6}$,
}
\\\textbf{
Yunzhu Li$^{\scriptscriptstyle 10}$,
Silvio Savarese$^{\scriptscriptstyle 1,8,12}$,
Hyowon Gweon$^{\scriptscriptstyle 7,8}$,
}
\\\textbf{
C. Karen Liu$^{\scriptscriptstyle 1,8}$, 
Jiajun Wu$^{\scriptscriptstyle 1,8}$, 
Li Fei-Fei$^{\scriptscriptstyle 1,8}$
}
\\\\
Department of Computer Science$^{\scriptscriptstyle 1}$, Department of Mechanical Engineering$^{\scriptscriptstyle 2}$
\\Neurosciences IDP$^{\scriptscriptstyle 3}$, Department of Aeronautics and Astronautics$^{\scriptscriptstyle 4}$
\\Institute for Computational and Mathematical Engineering$^{\scriptscriptstyle 5}$
\\Department of Electrical Engineering$^{\scriptscriptstyle 6}$, Department of Psychology$^{\scriptscriptstyle 7}$
\\Institute for Human-Centered Artificial Intelligence (HAI)$^{\scriptscriptstyle 8}$
\\Stanford University
\\
\\The University of Texas at Austin$^{\scriptscriptstyle 9}$ 
\\The University of Illinois Urbana-Champaign$^{\scriptscriptstyle 10}$
\\University of Southern California$^{\scriptscriptstyle 11}$
\\ Salesforce Research$^{\scriptscriptstyle 12}$
}

\begin{document}
\maketitle


\begin{abstract}
We present \benchmark, a comprehensive simulation benchmark for human-centered robotics.
\benchmark includes two components, guided and motivated by the results of an extensive survey on `\textit{what do you want robots to do for you?}'. The first is the definition of 1,000 everyday activities, grounded in 50 scenes (houses, gardens, restaurants, offices, etc.) with more than 9,000 objects annotated with rich physical and semantic properties. 
The second is \simulator, a novel simulation environment that supports these activities via realistic physics simulation and rendering of rigid bodies, deformable bodies, and liquids. 
Our experiments indicate that the activities in \benchmark are long-horizon and dependent on complex manipulation skills, both of which remain a challenge for even state-of-the-art robot learning solutions. To calibrate the simulation-to-reality gap of \benchmark, we provide an initial study on transferring solutions learned with a mobile manipulator in a simulated apartment to its real-world counterpart. 
We hope that \benchmark's human-grounded nature, diversity, and realism make it valuable for embodied AI and robot learning research.
Project website: \url{https://behavior.stanford.edu}.

\blfootnote{* indicates equal contribution\newline correspondence to \href{mailto:chengshu@stanford.edu,zharu@stanford.edu,jdwong@stanford.edu,cgokmen@stanford.edu,sanjana2@stanford.edu}{\{chengshu,zharu,jdwong,cgokmen,sanjana2\}@stanford.edu} }

\end{abstract}




\section{Introduction}
\label{s:intro}

Inspired by the progress that benchmarking brought to computer vision~\cite{deng2009imagenet,lin2014microsoft,everingham2010pascal,krishna2017visual,geiger2012we,goyal2017something,sigurdsson2018actor,xiang2017posecnn,martin2021jrdb,caba2015activitynet,gurari2018vizwiz} and natural language processing~\cite{marcinkiewicz1994building,wang2018glue,rajpurkar2016squad,socher2013recursive,antol2015vqa}, the robotics community has developed several benchmarks in simulation~\cite{batra2020rearrangement,weihs2021visual,gan2021threedworld,puig2018virtualhome,shridhar2020alfred,xia2020interactive,yu2020meta,james2019rlbench,savva2019habitat,szot2021habitat,srivastava2021behavior,zhu2020robosuite,tassa2018deepmind,brockman2016openai}.
The broader goal of these benchmarks is to fuel the development of general, effective robots that bring major benefits to people's daily lives---human-centered AI that ``serves human needs, goals, and values''~\cite{littman2021gathering, riedl2019human, xu2019toward,shneiderman2020bridging}. Inspiring as they are, the tasks and activities in those benchmarks are designed by researchers; it remains unclear if they are addressing the actual needs of humans. 

We observe that a human-centered robotic benchmark should not only be designed \emph{for} human needs, but also originated \emph{from} human needs: \textit{what everyday activities do humans want robots to do for them?} 
To this end, we conduct an extensive survey with 1,461 participants (see Sec.~\ref{sec:survey}) to rank a wide range of daily activities based on participants' desire to delegate these activities to robots. We also ask layperson annotators to provide definitions of those activities. The survey reveals systematicity in what activities people want robots to do, but more importantly, highlights two key factors that we should prioritize when designing robotic benchmarks: \textbf{diversity} in the type of scenes, objects, and activities, and \textbf{realism} of the underlying simulation environments. 

The most needed activities indicated by the survey range from `wash floor' to `clean bathtub.' Clearly, the diversity of these activities is far beyond what real-world robotics challenges may offer~\cite{kitano1997robocup,wisspeintner2009robocup,iocchi2015robocup,buehler2009darpa,krotkov2017darpa,correll2016analysis,eppner2017lessons,roa2021mobile}. 
Developing simulation environments is a natural alternative: one can train and test robotic agents in many activities with diverse scenes, objects, and conditions efficiently and safely. 
However, for this paradigm to work, the activities have to be simulated realistically, reproducing accurately the circumstances that a robot may encounter in the real world. 
While significant progress in realism has been made in specific domains~\cite{heiden2021disect,urakami2019doorgym,lin2020softgym}, achieving realism for a diverse set of activities remains a tremendous challenge, due to the effort required to provide realistic models and simulation features. 



In this work, we present \textbf{\benchmark}, a Benchmark of 1,000 Everyday Household Activities in Virtual, Interactive, and Ecological Environments---the next generation of \oldbenchmark~\cite{srivastava2021behavior}. 
\benchmark includes two novel components to address the demands for diversity and realism: the diverse \textbf{\dataset} and the realistic \textbf{\simulator} simulation environment. The \dataset is a large-scale dataset comprising 1) a commonsense knowledge base for 1,000 activities with definitions in predicate logic (initial and goal conditions), as well as the objects involved, their properties, and their state transitions, and 2) high-quality 3D assets including 50 scenes and 9,000+ object models with rich physical and semantic annotations. 

All activities in the \dataset are instantiated in a novel simulation environment, \simulator, which we build on top of Nvidia's Omniverse and PhysX 5~\cite{physx} to provide realistic physics simulation and rendering of rigid bodies, deformable bodies, and fluids. 
\simulator expands beyond Omniverse's capabilities with a set of extended object states like temperature, toggled, soaked, and dirtiness. It also includes capabilities to generate valid initial activity configurations and discriminate valid goal solutions based on activity definitions. 
With all these realistic simulation features, \simulator supports the 1,000 diverse activities in the \dataset.

We evaluate state-of-the-art reinforcement learning algorithms~\cite{schulman2017proximal,haarnoja2018soft} in several activities of \benchmark, both with visuomotor control in the original action space, and with action primitives that leverage sampling-based motion planning~\cite{Jordan.Perez.ea:CSAIL13}. Our analysis indicates that even a single activity in \benchmark is extremely challenging for current AI algorithms, and the baselines can only solve it with a significant injection of domain knowledge. Concretely, the difficulties derive in part from the length of \benchmark's activities and the complexity of the physical manipulation required. To calibrate the simulation-to-real gap of \benchmark, we provide an initial study on transferring solutions learned with a mobile manipulator in a simulated apartment to its real-world counterpart. 
We hope that the \benchmark benchmark, our survey, and our analysis will serve to support and guide the development of future embodied AI agents and robots.      

\vspace{1cm}

\begin{figure}[t]
    \centering
\includegraphics[width=0.99\textwidth]{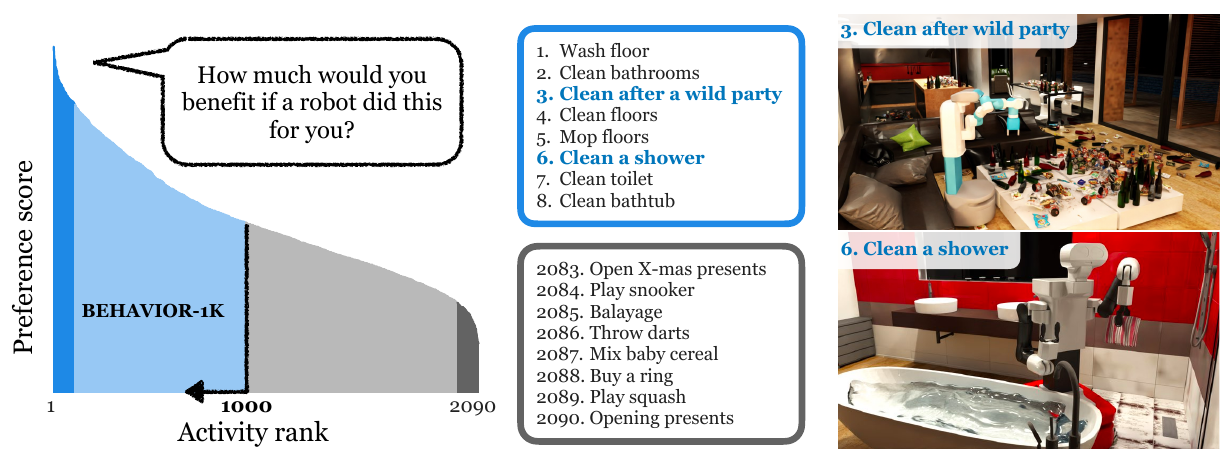}
\begin{minipage}{0.4\textwidth}
\end{minipage}
\caption{\textbf{Developing a Human-Centered Benchmark for Embodied AI.} Left: human preference score over 2,090 activities, ranked based on a survey on 1,461 participants. The distribution indicates the high \textbf{diversity} of needs and preferences of humans that should be reflected in a comprehensive benchmark. Middle: Example activities. Laborious activities are ranked the highest, while pleasurable ones are ranked the lowest. Right: visualization of two of the top 8 activities generated by our \textbf{realistic} \simulator simulation environment.}
\label{fig:activity_ranking}
\end{figure}

\section{Creating a Benchmark Grounded in Human Needs: A Survey Study}
\label{sec:survey}



A significant amount of robotics research aspires to satisfy human needs, but those needs are typically assumed or speculated. Human-centric development requires direct information about what humans want from autonomous agents~\cite{littman2021gathering}. To create a benchmark that reflects these needs, we conduct a survey targeting the general U.S. population that asks: \textit{what do you want robots to do for you?} The survey sources around 2,000 activities from time-use surveys~\cite{atus, hetus, mtus}, which record how people spend their time, and from WikiHow articles~\cite{wikihow}. We conduct the survey on Amazon Mechanical Turk with a total of 1,461 respondents (demographics in Appendix~\ref{sec:demographics}) and fifty 10-point Likert scale responses per activity.

Survey results are summarized in Fig.~\ref{fig:activity_ranking} (left), in which we rank the activities based on their human preference score. The full list of ranked activities can be found on our website. The distribution shows large statistical dispersion (Gini index$=0.158$): humans want robots to perform a wide range of activities, from cleaning chores to cooking large feasts. Tedious tasks like ``scrubbing the bathroom floor'' score the highest, while recreational activities like game-play score the lowest. There are around 200 cleaning activities and over 200 cooking activities, among many other categories.


\benchmark activities include the 909 activities with the highest human preference scores and 91 activities from \oldbenchmark~\cite{srivastava2021behavior}, which total the top-ranked 1,000 activities. \benchmark sets itself apart from other embodied AI benchmarks by being sourced from time-use surveys and using survey data to prioritize activities considered most important and useful by humans, and by including a tremendously diverse set of activities. 

\section{Related Work: Embodied AI Benchmarks}
\label{sec:related}


\begin{table}[!t]
    \definecolor{myGrayCell}{rgb}{.8,.8,.8}
    \centering
    \resizebox{\linewidth}{!}{
    \begin{tabular}{|c|c|c|cccccccccccccccccccccc|}
        \rot{} & 
        \rot{} & 
        \rot{\textbf{\benchmark}} & 
        \rot{BEHAVIOR-100} & 
        \rot{AI2THOR Vis. Room Rearr.} & 
        \rot{TDW Transport} & 
        \rot{Rearr. T5 (Habitat)} &
        \rot{ManipulaTHOR ArmPointNav} & 
        \rot{Interactive Gibson Benchmark} & 
        \rot{VirtualHome} & 
        \rot{ALFRED} & 
        \rot{Habitat 2.0 HAB} & 
        \rot{SAPIEN ManiSkill} & 
        \rot{Watch-And-Help} & 
        \rot{RFUniverse} & 
        \rot{Rearr. T2 (OCRTOC)} & 
        \rot{IKEA Furniture Assembly} & 
        \rot{RLBench} & 
        \rot{Metaworld} & 
        \rot{Robosuite} & 
        \rot{SoftGym} & 
        \rot{DeepMind Control Suite} & 
        \rot{OpenAIGym} & 
        \rot{Habitat 1.0} & 
        \rot{Gibson}
        \\\cline{3-25}
        \multicolumn{2}{c}{} & 
        \multicolumn{13}{|c}{Mobile manipulation} & 
        \multicolumn{8}{|c}{Static manipulation} & 
        \multicolumn{2}{|c|}{Navigation}
        \\\cline{2-25}
        
        \rot{} & 
        \multicolumn{1}{|c|}{\begin{tabular}{@{}c@{}}Activities from\\ human preference survey\end{tabular}} &
         \textcolor{indiagreen}\faCheck & 
         \textcolor{red}\faTimes & 
         \textcolor{red}\faTimes & 
         \textcolor{red}\faTimes & 
         \textcolor{red}\faTimes &
         \textcolor{red}\faTimes &
         \textcolor{red}\faTimes & 
         \textcolor{red}\faTimes & 
         \textcolor{red}\faTimes & 
         \textcolor{red}\faTimes & 
         \textcolor{red}\faTimes &  
         \textcolor{red}\faTimes & 
         \textcolor{red}\faTimes & 
         \textcolor{red}\faTimes & 
         \textcolor{red}\faTimes & 
         \textcolor{red}\faTimes & 
         \textcolor{red}\faTimes & 
         \textcolor{red}\faTimes & 
         \textcolor{red}\faTimes & 
         \textcolor{red}\faTimes & 
         \textcolor{red}\faTimes & 
         \textcolor{red}\faTimes & 
         \textcolor{red}\faTimes 
        \\       
        
        \hline
        \rule{0pt}{12pt} & 
         \begin{tabular}{@{}c@{}} Activities\end{tabular} & 
         \textbf{1000} & 
         100 & 
         1 & 
         1 & 
         1 &
         1 & 
         2 & 
         549 & 
         7 & 
         3 &
         4 &
         5 & 
         5 &
         5 & 
         100 & 
         50 & 
         1 & 
         5 & 
         10 & 
         28 & 
         8 & 
         2 & 
         3
        \\

        & 
        \begin{tabular}{@{}c@{}} Scene types \end{tabular} & 
        \textbf{8} & 
        1 & 
        1 &
        1 &
        1 & 
        1 &
        1 &
        1 &
        1 & 
        1 & 
        1 & 
        1 & 
        1 & 
        1 & 
        1 & 
        1 & 
        1 & 
        1 & 
        1 & 
        1 & 
        1 & 
        1 & 
        1
        \\
        & 
         Scenes / rooms &
        \textbf{50/373} & 
        15/100 & 
        -/120 & 
        15/105 & 
        55 static/- & 
        -/30 & 
        10/- & 
        6/24 &
        -/120 & 
        1/6 & 
        1/- & 
        7/29 & 
        Gibson &
        1/- & 
        1/- & 
        1/- & 
        1/- & 
        1/- & 
        1/- & 
        1/- & 
        1/- & 
        HM3D &
        572 static
        \\

        \multirow{1}*{\rotatebox{90}{Diversity}}
         & 
         \begin{tabular}{@{}c@{}} Object categories\end{tabular} & 
         \textbf{1949} & 
         391 & 
         118 & 
         ~50 & 
         YCB & 
         150 & 
         5 & 
         308 & 
         84 & 
         41+YCB &
         4 &
         117 &
         UNK &
         12+YCB & 
         73+ & 
         ~28 & 
         7 & 
         10 & 
         4 & 
         4 & 
         4 & 
         Mtpt. & 
         N/A
        \\
        & 
        \begin{tabular}{@{}c@{}} Object models\end{tabular} & 
        \textbf{9318} & 
        1217 & 
        118 &
        112 & 
        YCB &
        150 & 
        152 & 
        UNK &
        84 &
        92+YCB & 
        162 & 
        UNK & 
        UNK &
        101+YCB & 
        73+ & 
        28 & 
        80 & 
        10 & 
        4 & 
        4 & 
        4 & 
        N/A & 
        N/A
        \\
        %
        &
        \begin{tabular}{@{}c@{}} Objs. per activity\end{tabular} & 
        \textbf{3-47} & 
        3-34 & 
        5 &
        7-9 &
        2-5 &
        2-3 & 
        ~10 & 
        1-24 &
        2 &
        5 &
        1 &
        2-8 &
        1-6 &
        5-10 &
        1-2 &
        1-2 &
        1 &
        1-3 &
        1-3 &
        1-3 &
        1 &
        0-1 &
        N/A
        \\        
        
        & 
        \begin{tabular}{@{}c@{}}Diff. state changes \\required per activity \end{tabular} & 
        \textbf{2-11} & 
        2-8 & 
        4 & 
        4 &
        4 &
        2 &
        1-3 & 
        1-7 & 
        2-3 & 
        1-2 &
        1 & 
        1-3 &
        1 &
        1 &
        1-3 & 
        1-4 & 
        4 & 
        1 & 
        1-3 & 
        1-2 & 
        1-2 & 
        1 & 
        1
        \\
        
         & 
        \begin{tabular}{@{}c@{}}Infinite scene-\\agnostic instantiation\end{tabular} & 
        \textcolor{indiagreen}\faCheck & 
        \textcolor{indiagreen}\faCheck & 
        \textcolor{red}\faTimes & 
        \textcolor{red}\faTimes & 
        \textcolor{red}\faTimes & 
        \textcolor{red}\faTimes & 
        \textcolor{red}\faTimes & 
        \textcolor{red}\faTimes & 
        \textcolor{indiagreen}\faCheck & 
        \textcolor{indiagreen}\faCheck & 
        \textcolor{red}\faTimes & 
        \textcolor{red}\faTimes & 
        \textcolor{red}\faTimes & 
        \textcolor{red}\faTimes & 
        \textcolor{red}\faTimes & 
        \textcolor{red}\faTimes & 
        \textcolor{red}\faTimes & 
        \textcolor{red}\faTimes & 
        \textcolor{red}\faTimes & 
        \textcolor{red}\faTimes & 
        \textcolor{red}\faTimes & 
        N/A & 
        N/A
        \\ 
        \cline{1-25}

        \rule{0pt}{12pt} 
        & 
        \begin{tabular}{@{}c@{}}Visual quality score\end{tabular} & 
        $\mathbf{3.20}$&
        $1.69$ &
        $1.73$ &
        $1.65$ &
        - &
        $1.73$ &
        - &
        - &
        $1.73$ &
        $1.74$ &
        - &
        - &
        - &
        - &
        - &
        - &
        - &
        - &
        - &
        - &
        - &
        - &
        - 
        \\ 
        & 
        \begin{tabular}{@{}c@{}}Kinematics, dynamics\end{tabular} & 
        \textcolor{indiagreen}\faCheck &
        \textcolor{indiagreen}\faCheck &
        \textcolor{indiagreen}\faCheck &
        \textcolor{indiagreen}\faCheck &
        \textcolor{indiagreen}\faCheck &
        \textcolor{indiagreen}\faCheck &
        \textcolor{indiagreen}\faCheck &
        \textcolor{red}\faTimes &
        \textcolor{indiagreen}\faCheck &
        \textcolor{indiagreen}\faCheck &
        \textcolor{indiagreen}\faCheck &
        \textcolor{red}\faTimes &
        \textcolor{indiagreen}\faCheck &
        \textcolor{indiagreen}\faCheck &
        \textcolor{indiagreen}\faCheck &
        \textcolor{indiagreen}\faCheck &
        \textcolor{indiagreen}\faCheck &
        \textcolor{indiagreen}\faCheck &
        \textcolor{indiagreen}\faCheck &
        \textcolor{indiagreen}\faCheck &
        \textcolor{indiagreen}\faCheck &
        \textcolor{indiagreen}\faCheck &
        \textcolor{indiagreen}\faCheck 
        \\ 
        
        \multirow{1}*{\rotatebox{90}{Realism}} & 
         
        \begin{tabular}{@{}c@{}}Continuous extended\\states (temp., wetness)\end{tabular} & 
        \textcolor{indiagreen}\faCheck & 
        \textcolor{indiagreen}\faCheck & 
        \textcolor{red}\faTimes & 
        \textcolor{red}\faTimes & 
        \textcolor{red}\faTimes & 
        \textcolor{red}\faTimes & 
        \textcolor{red}\faTimes & 
        \textcolor{red}\faTimes & 
        \textcolor{red}\faTimes & 
        \textcolor{red}\faTimes &
        \textcolor{red}\faTimes & 
        \textcolor{red}\faTimes & 
        \textcolor{red}\faTimes & 
        \textcolor{red}\faTimes & 
        \textcolor{red}\faTimes & 
        \textcolor{red}\faTimes & 
        \textcolor{red}\faTimes & 
        \textcolor{red}\faTimes & 
        \textcolor{red}\faTimes & 
        \textcolor{red}\faTimes & 
        \textcolor{red}\faTimes & 
        \textcolor{red}\faTimes & 
        \textcolor{red}\faTimes 
        \\ 
        & 
        \begin{tabular}{@{}c@{}} Flexible materials\end{tabular} & 
        \textcolor{indiagreen}\faCheck & 
        \textcolor{red}\faTimes & 
        \textcolor{red}\faTimes & 
        \textcolor{red}\faTimes & 
        \textcolor{red}\faTimes & 
        \textcolor{red}\faTimes & 
        \textcolor{red}\faTimes &
        \textcolor{red}\faTimes &
        \textcolor{red}\faTimes & 
        \textcolor{red}\faTimes & 
        \textcolor{red}\faTimes & 
        \textcolor{red}\faTimes & 
        \textcolor{indiagreen}\faCheck &
        \textcolor{red}\faTimes & 
        \textcolor{red}\faTimes & 
        \textcolor{red}\faTimes & 
        \textcolor{red}\faTimes & 
        \textcolor{red}\faTimes & 
        \textcolor{indiagreen}\faCheck & 
        \textcolor{red}\faTimes & 
        \textcolor{red}\faTimes & 
        \textcolor{red}\faTimes & 
        \textcolor{red}\faTimes 
        \\ 
        & 
        \begin{tabular}{@{}c@{}}Deformable bodies\end{tabular} & 
        \textcolor{indiagreen}\faCheck & 
        \textcolor{red}\faTimes & 
        \textcolor{red}\faTimes & 
        \textcolor{red}\faTimes & 
        \textcolor{red}\faTimes & 
        \textcolor{red}\faTimes & 
        \textcolor{red}\faTimes &
        \textcolor{red}\faTimes & 
        \textcolor{red}\faTimes & 
        \textcolor{red}\faTimes & 
        \textcolor{red}\faTimes & 
        \textcolor{red}\faTimes & 
        \textcolor{indiagreen}\faCheck &
        \textcolor{red}\faTimes & 
        \textcolor{red}\faTimes & 
        \textcolor{red}\faTimes & 
        \textcolor{red}\faTimes & 
        \textcolor{red}\faTimes & 
        \textcolor{red}\faTimes & 
        \textcolor{red}\faTimes & 
        \textcolor{red}\faTimes & 
        \textcolor{red}\faTimes & 
        \textcolor{red}\faTimes 
        \\ 
        & 
        \begin{tabular}{@{}c@{}}Realistic Fluid\end{tabular} & 
        \textcolor{indiagreen}\faCheck & 
        \textcolor{red}\faTimes & 
        \textcolor{red}\faTimes & 
        \textcolor{red}\faTimes & 
        \textcolor{red}\faTimes & 
        \textcolor{red}\faTimes & 
        \textcolor{red}\faTimes & 
        \textcolor{red}\faTimes & 
        \textcolor{red}\faTimes & 
        \textcolor{red}\faTimes & 
        \textcolor{red}\faTimes &
        \textcolor{red}\faTimes &
        \textcolor{indiagreen}\faCheck &
        \textcolor{red}\faTimes &
        \textcolor{red}\faTimes & 
        \textcolor{red}\faTimes & 
        \textcolor{red}\faTimes & 
        \textcolor{red}\faTimes & 
        \textcolor{indiagreen}\faCheck & 
        \textcolor{red}\faTimes & 
        \textcolor{red}\faTimes & 
        \textcolor{red}\faTimes & 
        \textcolor{red}\faTimes 
        \\ 
        & 
        \begin{tabular}{@{}c@{}}Thermal effects (steam/fire)\end{tabular} & 
        \textcolor{indiagreen}\faCheck & 
        \textcolor{red}\faTimes & 
        \textcolor{red}\faTimes & 
        \textcolor{red}\faTimes & 
        \textcolor{red}\faTimes & 
        \textcolor{red}\faTimes & 
        \textcolor{red}\faTimes & 
        \textcolor{red}\faTimes & 
        \textcolor{red}\faTimes & 
        \textcolor{red}\faTimes & 
        \textcolor{red}\faTimes &
        \textcolor{red}\faTimes &
        \textcolor{indiagreen}\faCheck &
        \textcolor{red}\faTimes &
        \textcolor{red}\faTimes & 
        \textcolor{red}\faTimes & 
        \textcolor{red}\faTimes & 
        \textcolor{red}\faTimes & 
        \textcolor{red}\faTimes & 
        \textcolor{red}\faTimes & 
        \textcolor{red}\faTimes & 
        \textcolor{red}\faTimes & 
        \textcolor{red}\faTimes 
        \\ 
         &
         \begin{tabular}{@{}c@{}}Realistic action execution\end{tabular} & 
         \textcolor{indiagreen}\faCheck &
         \textcolor{indiagreen}\faCheck &
         \textcolor{red}\faTimes &
         \textcolor{indiagreen}\faCheck &
         \textcolor{red}\faTimes &
         \textcolor{indiagreen}\faCheck &
         \textcolor{indiagreen}\faCheck &
         \textcolor{red}\faTimes &
         \textcolor{red}\faTimes &
         \textcolor{indiagreen}\faCheck &
         \textcolor{indiagreen}\faCheck &
         \textcolor{red}\faTimes &
         \textcolor{indiagreen}\faCheck &
         \textcolor{indiagreen}\faCheck &
         \textcolor{indiagreen}\faCheck &
         \textcolor{indiagreen}\faCheck &
         \textcolor{indiagreen}\faCheck &
         \textcolor{indiagreen}\faCheck &
         \textcolor{indiagreen}\faCheck &
         \textcolor{indiagreen}\faCheck &
         \textcolor{indiagreen}\faCheck &
         \textcolor{indiagreen}\faCheck &
         \textcolor{indiagreen}\faCheck 
        \\\cline{1-25}


        \rot{} 
        \rule{0pt}{15pt} & 
        \multicolumn{1}{|c|}{\begin{tabular}{@{}c@{}}Benchmark focus: Task-\\Planning and/or Control\end{tabular}} & 
        TP+C & 
        TP+C & 
        TP & 
        TP+C & 
        TP+C & 
        TP+C & 
        C & 
        TP &         
        TP & 
        TP+C & 
        C & 
        TP &
        TP+C &
        TP+C & 
        C & 
        TP+C & 
        C & 
        C & 
        C & 
        C & 
        C & 
        C & 
        C 
        \\\cline{2-25}
    \end{tabular}
  }
  \caption{\textbf{Comparison of Embodied AI Benchmarks:} 
  \benchmark contains 1,000 diverse activities that are grounded by human needs. It achieves a new level of diversity in scenes, objects, and state changes involved. \simulator provides realistic simulation of these 1,000 activities, including some of the most advanced simulation and rendering features such as fluid and deformable bodies. This table is extended from~\cite{srivastava2021behavior}.}

  \label{tab:comparingbenchmarks}
\end{table}


We provide an extensive comparison between \benchmark, and other embodied AI benchmarks in simulation~\cite{batra2020rearrangement,weihs2021visual,gan2021threedworld,puig2018virtualhome,shridhar2020alfred,xia2020interactive,yu2020meta,james2019rlbench,savva2019habitat,szot2021habitat} in Table~\ref{tab:comparingbenchmarks}. We include a number of factors that contribute to diversity and realism and observe a significant step forward with \benchmark.
First, no other benchmark grounds their activity set on the needs of lay people.  
Other benchmarks often target a relatively restricted set of activities, and their simulators are realistic only in the relevant aspects for those tasks. In fact, we often observe a diversity-realism tradeoff. For instance, instruction-following benchmarks such as VirtualHome~\cite{puig2018virtualhome} and ALFRED~\cite{puig2018virtualhome, shridhar2020alfred} are diverse in the number of scenes, objects, and state changes, but offer a limited low-level physical realism. On the other hand, household rearrangement benchmarks such as Habitat 2.0 HAB~\cite{szot2021habitat}, TDW Transport~\cite{gan2021threedworld}, and SAPIEN ManiSkill~\cite{xiang2020sapien,mu2021maniskill} support realistic action execution and accurate physics simulation for rigid bodies, but only include a handful of tasks. Similarly, SoftGym~\cite{lin2020softgym} and RFUniverse~\cite{fu2022rfuniverse} have the closest simulation features and hence realism to \simulator, but they also lack the task diversity needed to support the development of human-centered general robots. 

The most similar benchmark to us is the previous generation \oldbenchmark~\cite{srivastava2021behavior}. \oldbenchmark brought forward several beneficial design choices that we inherit in \benchmark such as the activity sources (ATUS~\cite{atus}), activity definition logic language, and evaluation metrics. However, it fell short in the diversity and realism necessary to support a human-centered embodied AI benchmark in simulation, dimensions where \benchmark achieves unmatched levels.
While \oldbenchmark comprises 100 activities selected by researchers, our \benchmark increases diversity by one order of magnitude, to 1,000 activities, that are grounded in human needs thanks to our unique survey. 
Furthermore, \oldbenchmark includes only 15 scenes (all houses) and 300+ object categories, while \benchmark increases to 50 scenes (houses, stores, restaurants, offices, etc.) and 1,900+ object categories. In terms of realism, \benchmark extends the simulatable physical states and processes with \simulator: fluids, flexible materials, mixing substances, etc. The realism achieved in rendering by \simulator for \benchmark is also significantly higher than what was possible in \oldbenchmark and other benchmarks (see Fig.~\ref{fig:sim_render}).


\section{\dataset}
\label{sec:activity}

\begin{figure}[t]
    \centering
    \includegraphics[width=\textwidth]{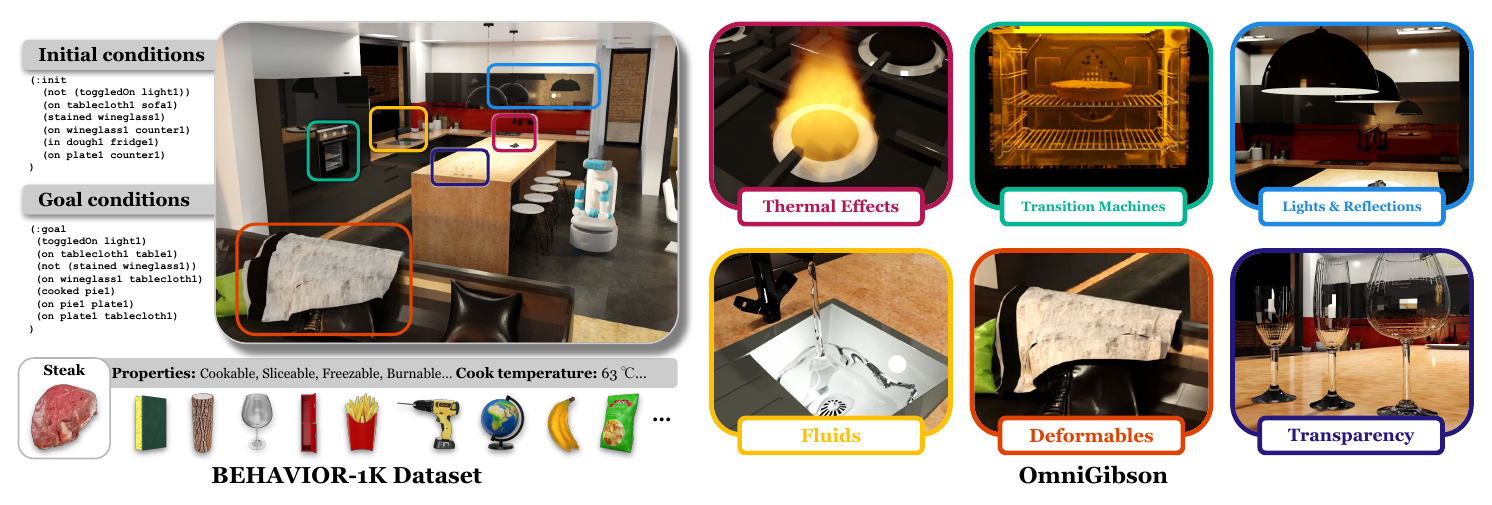}
    \caption{\textbf{Elements of \benchmark.} Our benchmark comprises two elements: \dataset and \simulator. Left: \dataset includes 1,000 BDDL activity definitions (top left), 50 realistic and diverse scenes (top right), and 9,000+ objects with properties annotated in the knowledge base (bottom). Right: \simulator provides the necessary functionalities to realistically simulate the 1000 activities, including thermal effects such as fire/steam/smoke (top left), fluid dynamics (bottom left), functional machines for transition rules (top center), deformable bodies/cloths (bottom center), realistic lighting and reflections (top right), and transparency rendering (bottom right). Together, they constitute a concrete, realistic instantiation of an everyday activity like \texttt{CookingDinner} in simulation.}
    \label{fig:model_vis}
\end{figure}

Once activities have been sourced to reflect human needs, they need to be concretely defined and instantiated the way they would occur in the real world. We build the \dataset, which includes a knowledge base of crowdsourced activity definitions with relevant objects and object states, and a large-scale repository of high-quality, interactive 3D models.


We crowdsource concrete definitions of activities in the form of \bddlfull (\bddl)~\cite{srivastava2021behavior}. \bddl is based on predicate logic and designed to be accessible for laypeople to describe concrete initial and goal conditions for a given activity. Unlike geometric, image/video, or experience goal specifications~\cite{batra2020rearrangement, weihs2021visual}, \bddl definitions are in terms of objects and object states, allowing annotators to define at an intuitive semantic level. The semantic symbols also capture the fact that multiple physical states might be valid initializations and solutions to an activity. See Listings~\ref{lst:bddl3},~\ref{lst:bddl4} and~\ref{lst:bddl5} in Appendix for example definitions.


The object and object state spaces that activity definitions are built upon are annotated to be ecologically plausible. The object spaces are derived from 5,000 WikiHow articles for the 1,000 activities and mapped to 2,964 leaf-level synsets either from WordNet~\cite{wordnet} or custom-designed. Through crowdworkers, students, and GPT-3~\cite{gpt3}, we also associate each object with our fully simulatable object states: for example, \texttt{apple} is associated with \texttt{cooked} and \texttt{sliced}, but not \texttt{toggledOn}.
Many object-property pairs are augmented with parameters, e.g., ``cooked temperature for apples'', taking advantage of \simulator's continuous extended states to make activities especially realistic. Finally, annotators and researchers also create transition rules, e.g., turning tomatoes and salt into sauces, or requiring sandpaper to remove rust. The result is a knowledge base of tens of thousands of elements underlying 1000 ecologically plausible activity definitions. We ensure annotation quality by having five experienced machine learning annotators verify a subset of all types of annotations and receive extremely high approval rates ($>$96.8\%). See Appendix~\ref{as:act_ann} for more details about the knowledge base.

The diversity of these activity definitions requires diverse object and scene models. 
On top of the 15 house scenes from \oldbenchmark~\cite{srivastava2021behavior}, we acquire 35 fully interactive scenes across diverse scene types, such as gardens, offices, restaurants, and stores, that are essential for everyday activities. This is unprecedented compared to other benchmarks (see Table~\ref{tab:comparingbenchmarks}). We also acquire 9,000+ object instances across 1,900+ categories required by the activities, and annotate rich physical (e.g., friction, mass, articulation) and semantic properties (e.g., category) for each object. Representative scene and object models can be seen in Fig.~\ref{fig:model_vis}. More details of the 3D models can be found in Appendix~\ref{as:modeling}.

\section{\simulator: Instantiating \benchmark with Realistic Simulation}
\label{sec:simulation}

\begin{figure}[t]
\captionsetup[subfloat]{labelformat=empty}
\captionsetup[subfloat]{justification=centering}
\centering
\subfloat[][\textbf{\simulator\\ $\mathbf{3.20\pm1.23}$}]{\includegraphics[width=.199\textwidth]{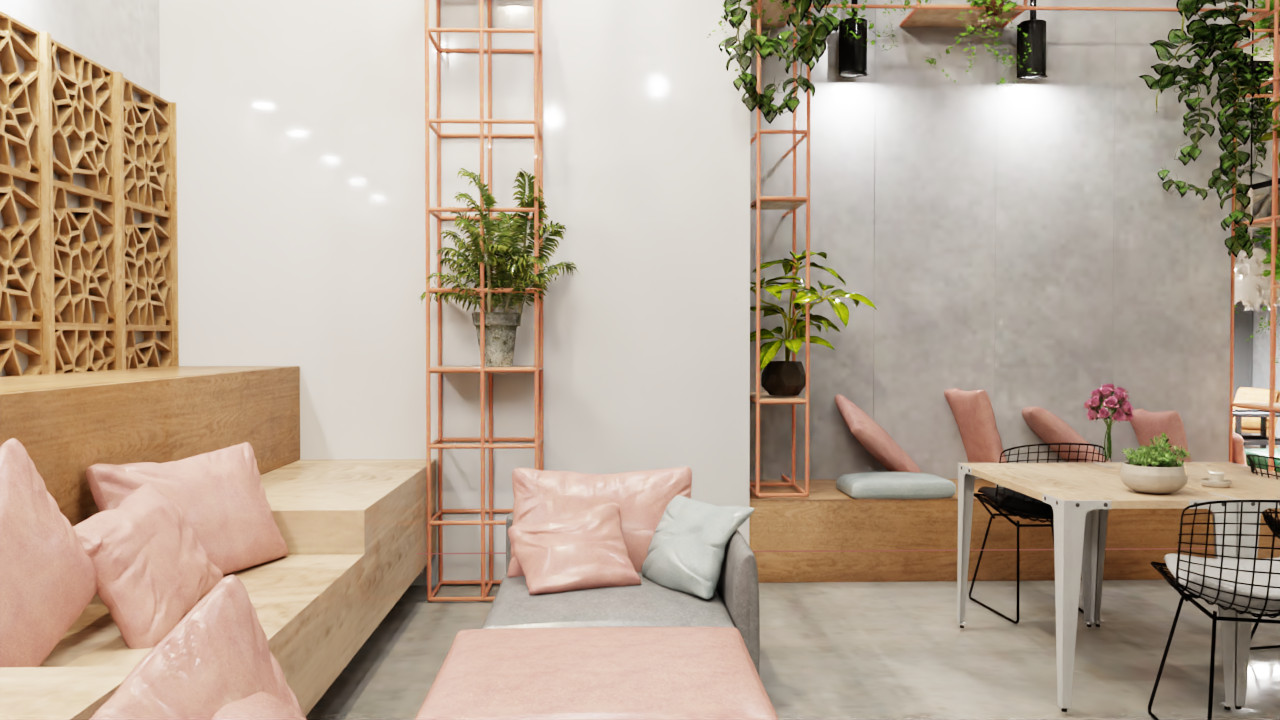}}
\hfill
\subfloat[][Habitat 2.0\\ $1.74\pm1.33$]{\includegraphics[width=.199\textwidth]{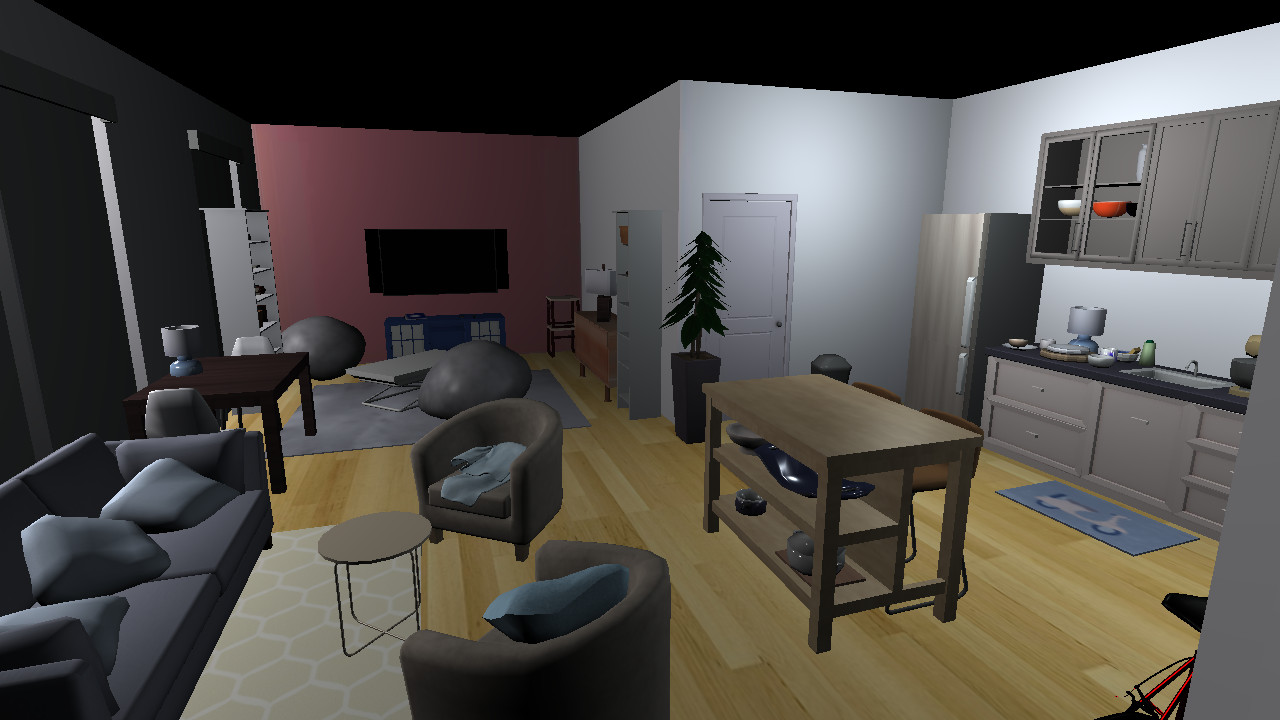}}
\hfill
\subfloat[][AI2-THOR\\ $1.73\pm1.37$]{\includegraphics[width=.199\textwidth]{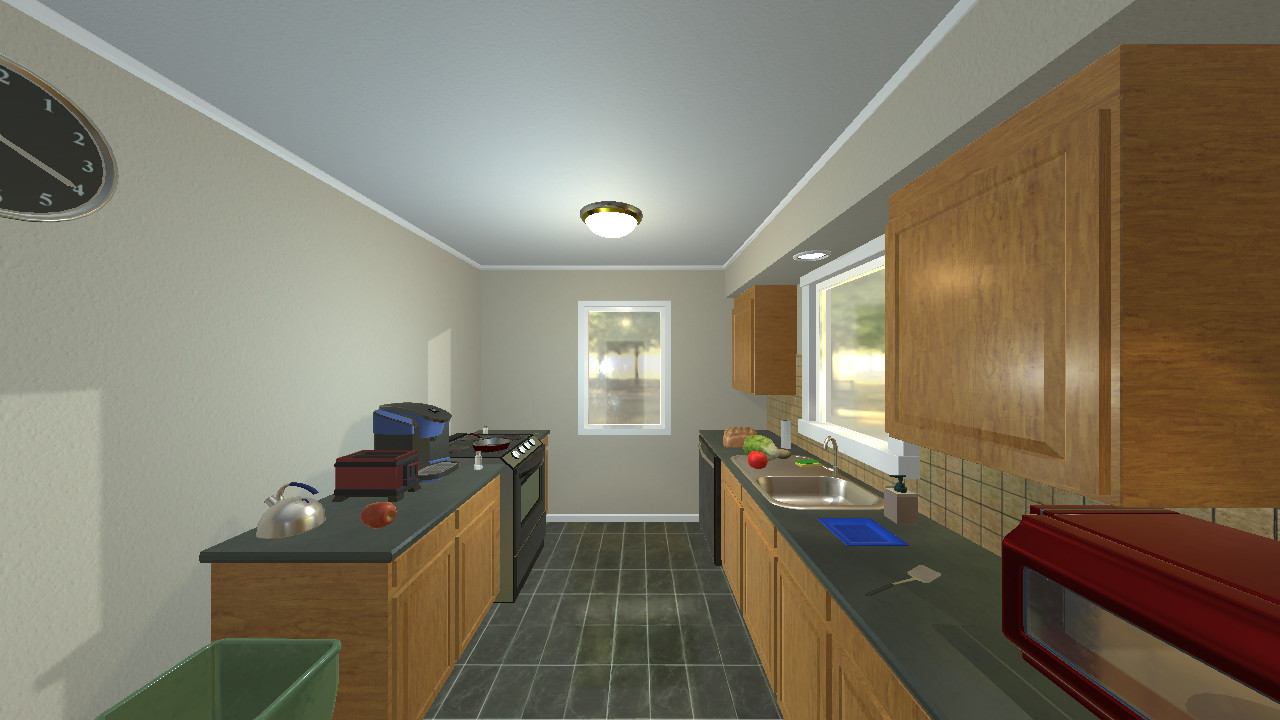}}
\hfill
\subfloat[][iGibson 2.0\\ $1.69\pm1.24$]{\includegraphics[width=.199\textwidth]{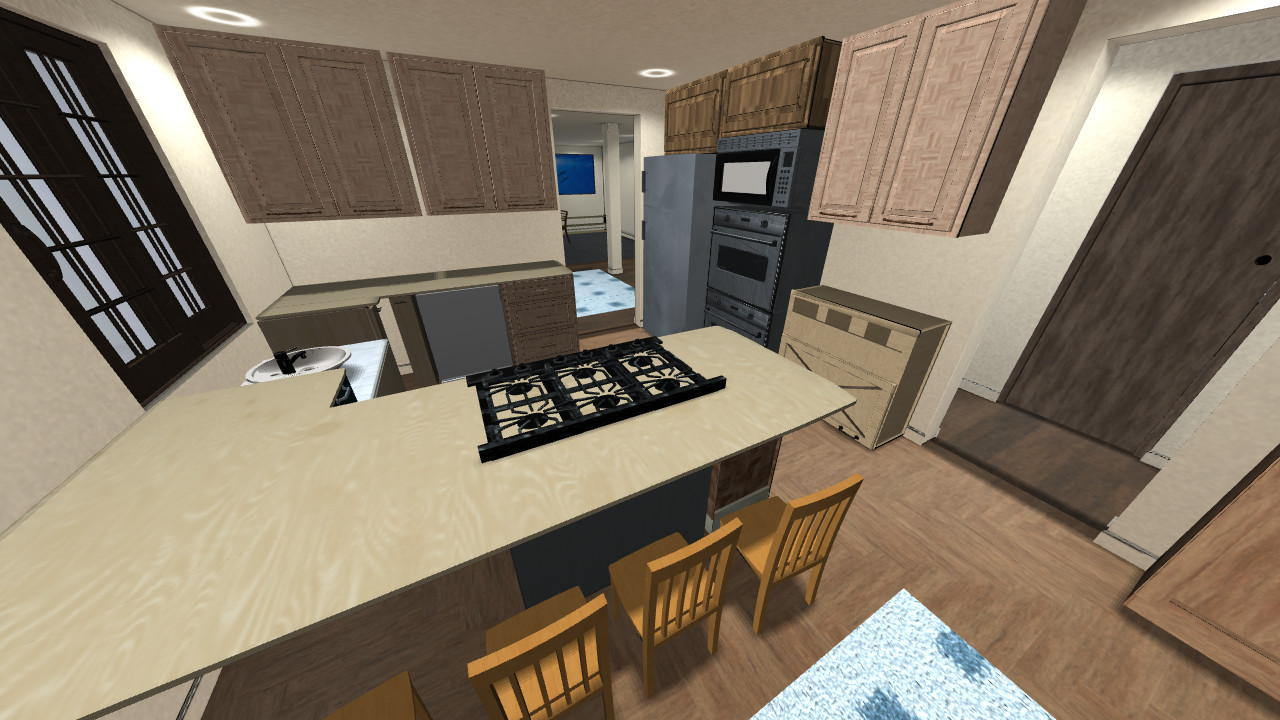}}
\hfill
\subfloat[][ThreeDWorld\\ $1.65\pm1.23$]{\includegraphics[width=.199\textwidth]{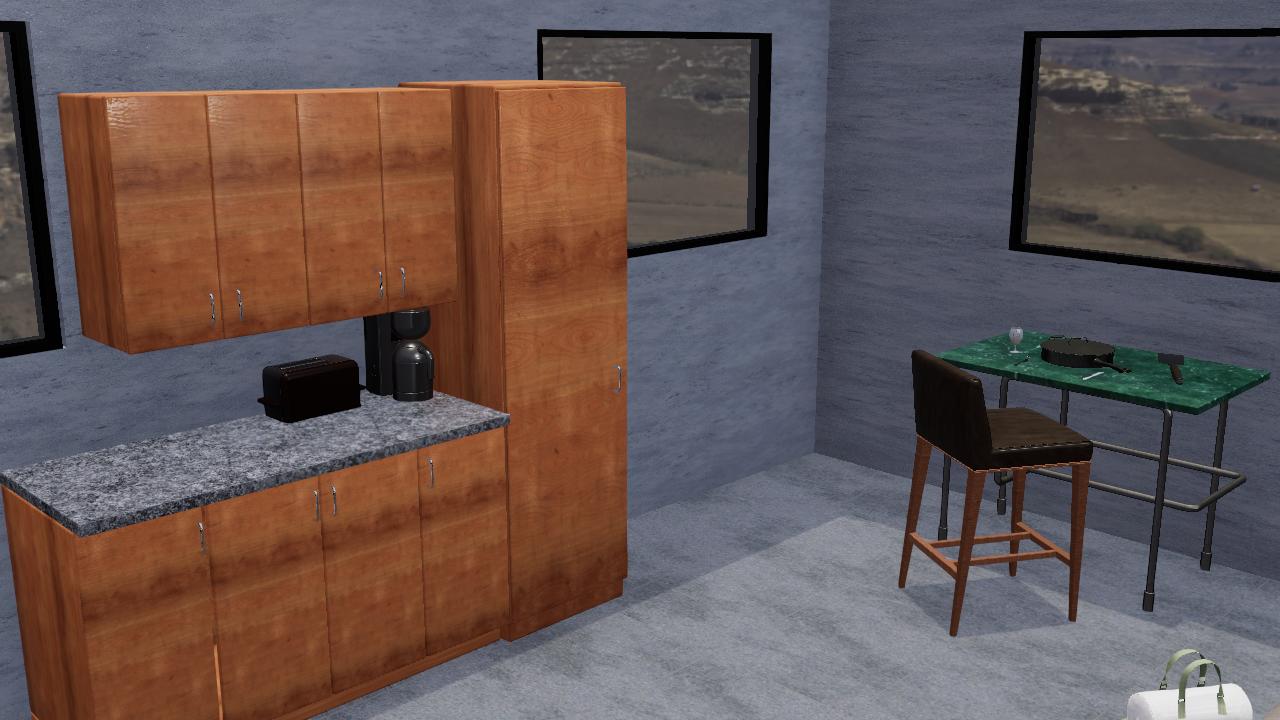}}
\vspace{15pt}
\caption{\textbf{Comparison of Visual Realism:} We evaluate \simulator's visual realism against other simulation environments by running a survey with 60 human subjects. We ask them to rank the realism of sampled images from each environment with a score of 5 (most realistic) to 1 (least realistic). We report the mean and standard deviation and show a sampled image from the study. We observe that the participants consider \simulator to be significantly more visually realistic than all other environments. See Appendix~\ref{amt_study_appx} for more info.}
\label{fig:sim_render}
\end{figure}

\begin{figure}[t]
    \centering
    \includegraphics[width=1\textwidth]{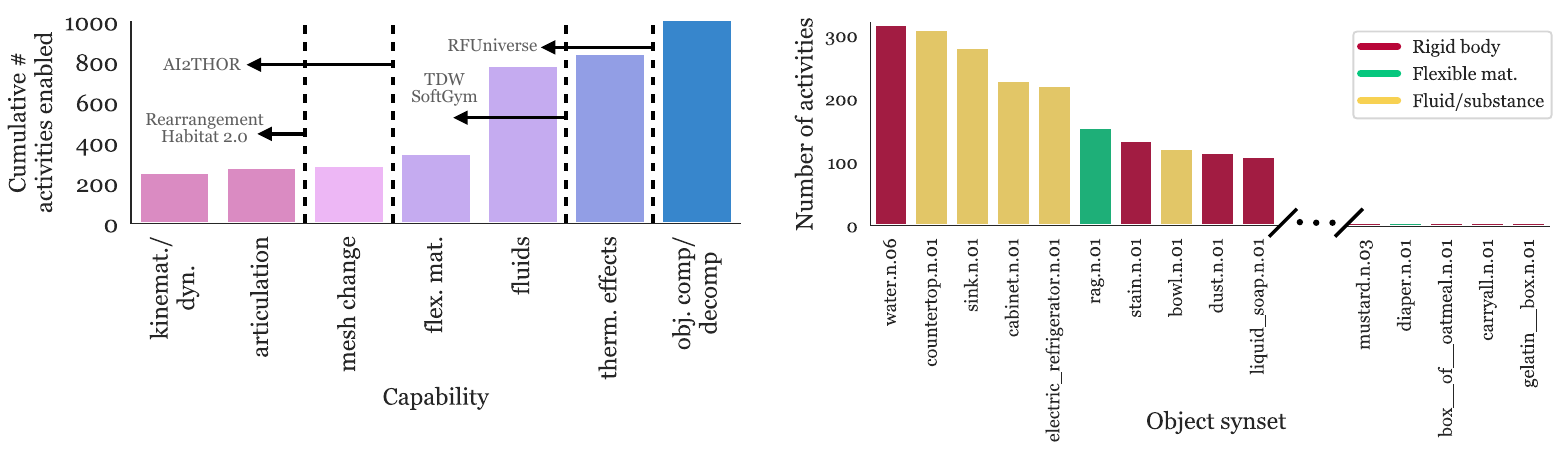}\\
    \caption{\textbf{Objects and States in Activity Definitions:} Left: the number of activities unlocked by each simulation capability that \simulator has. None of the other simulation environments are sufficient to fully support \benchmark, e.g. Habitat 2.0 can support only 23\% of the activities. Right: the number of activities that require each object synset (category). Several top-10 object synsets are fluids and flexible materials, necessitating the development of \simulator. As we expect, the object synsets also follow a long-tail distribution: most objects are involved in only a few activities.}
    \vspace{10pt}
    \label{fig:sim_obj_prop}
\end{figure}

\oldbenchmark is implemented in iGibson 2.0~\cite{li2021ig2}; however, realistic simulation of the diverse activities in \benchmark is beyond the capability of iGibson 2.0. We present a novel simulation environment, \simulator, that provides the necessary functionalities to support and instantiate \benchmark.
\simulator is built on top of Nvidia Omniverse and PhysX 5, providing the simulation of not only rigid bodies, but also deformable objects, fluids, and flexible materials (see Fig.~\ref{fig:sim_obj_prop}), while generating highly realistic ray-traced or path-traced virtual images (see Fig.~\ref{fig:sim_render}).
These features significantly boost the realism of \benchmark compared to other benchmarks.

Similar to \oldbenchmark, \simulator also simulates additional, non-kinematic extended object states (e.g., temperature, soaked level) based on heuristics (e.g., temperature increases when being next to a heat source that is toggled on). \simulator also implements the functionalities to generate infinite valid physical configurations that satisfy the activities' initial conditions as logical predicates (e.g. \texttt{food} is \texttt{frozen}), and to evaluate their goal conditions (e.g. \texttt{food} is \texttt{cooked} and \texttt{onTop} of a \texttt{plate}, the \texttt{cloth} is \texttt{folded}) based on the object's physical states (pose and joint configuration) and extended states. \revision{\simulator natively supports randomization during scene initialization and can sample amongst object models and their poses/states.} The full details of extended object states and logical predicates that \simulator supports can be found in Appendix~\ref{as:object_states}. 



Many everyday tasks are difficult to simulate because they require modeling complex physical processes, such as folding a towel or pouring a glass of water. \simulator unlocks them by supporting realistic simulation of fluids, deformable bodies, and cloths (see Fig.~\ref{fig:model_vis}). Indeed, without these features, over half of \benchmark activities would not be simulatable, highlighting how crucial these features are for capturing everyday activities. \simulator also captures multiple physical processes that are not natively simulatable by Omniverse, such as baking pies or pureeing vegetables. Aside from the extended states mentioned above, we also design a modular \textit{Transition Machine}, which specifies custom transitions between groups of objects when specified conditions are met. For example, a dough placed inside an oven that reaches a certain temperature threshold will turn into a pie. This further expands \simulator's capacity to simulate complex, realistic activities that would otherwise be intractable to fully simulate physically.

\section{Experiments: Evaluating Embodied AI Solutions in \benchmark}
\label{sec:baseline}

In our experiments, we aim to answer three questions: How do existing vision-based robot learning algorithms perform in \benchmark, and what assumptions have to be made to improve their success? What elements of the activities are the most problematic for current AI? What are the main sources of the sim-real gap in \benchmark/\simulator? Our goal is to indicate promising research directions to improve AI's performance in \benchmark activities in simulation and, ultimately, in the real world.


\subsection{Evaluating \benchmark Solutions in \simulator}
\label{ss:ev_in_sim}

\paragraph{Experimental Setups.} We selected three paradigmatic activities for our experiments: 
\texttt{CollectTrash}, where the agent gathers empty bottles and cups, and throws them into a trash bin (rigid body manipulation); \texttt{StoreDecoration}, where the agent stores items into a drawer (articulated object manipulation); and \texttt{CleanTable}, where the agent wipes a dirty table with a soaked piece of cloth (manipulation of flexible materials and fluids). 
We evaluate three different baselines based on state-of-the-art reinforcement learning algorithms (RL)~\cite{barto2003recent}: 
\begin{itemize}
    \item \rlvmc, a visuomotor control (from image to low-level joint commands) RL solution based on Soft Actor-Critic (SAC)~\cite{haarnoja2018soft};
    \item \rlprim, a RL solution based on PPO~\cite{schulman2017proximal} that leverages a set of action primitives based on a sampling-based motion planner~\cite{lavalle2006planning,kuffner2000rrt,Jordan.Perez.ea:CSAIL13} (\texttt{pick}, \texttt{place}, \texttt{push}, \texttt{navigate}, \texttt{dip} and \texttt{wipe}). The policy outputs a discrete selection of a primitive applied on an object;
    \item \rlprimhist, a variant of \rlprim that takes in the history observations (3 steps) as additional inputs to help disentangle similar-looking states.    
\end{itemize} 

All agents are trained with a sparse task success reward without any reward engineering. Following the metrics proposed in \oldbenchmark~\cite{srivastava2021behavior}, we report the success rate and efficiency metrics (distance traveled, time invested, and disarrangement caused) in Table~\ref{tab:baseline_performance} and~\ref{tab:baseline_metric}, and the success score Q in Table~\ref{tab:baseline_q_score} in Appendix. 

\begin{table}[t]
    \centering
    \resizebox{\textwidth}{!}{\begin{tabular}{l|cc|ccc}
    \toprule
    \multicolumn{1}{c|}{\multirow{2}{*}{Method}} & \multicolumn{2}{c|}{Policy Features} & \multicolumn{3}{c}{Task success rate} \\
    \multicolumn{1}{c|}{} & Primitives & History & \texttt{StoreDecoration} & \texttt{CollectTrash} & \texttt{CleanTable} \\ \midrule
    \rlvmc & \textcolor{red}\faTimes & \textcolor{red}\faTimes & $0.0 \pm 0.0$ & $0.0 \pm 0.0$ & $0.0 \pm 0.0$ \\
    \rlprim & \textcolor{indiagreen}\faCheck & \textcolor{red}\faTimes & $0.48 \pm 0.06$ & $0.42 \pm 0.02$ & $0.77 \pm 0.08$ \\
    \rlprimhist & \textcolor{indiagreen}\faCheck & \textcolor{indiagreen}\faCheck & $0.55 \pm 0.05$ & $0.63 \pm 0.03$ & $0.88 \pm 0.02$ \\ \bottomrule
    \end{tabular}
    }
    \caption{Task success rates across three baseline methods. \rlvmc with end-to-end visuomotor control completely fails to solve any of the activities, whereas \rlprim and \rlprimhist with action primitives are able achieve decent performance. Memory of observations helps in longer horizon activities (e.g. \texttt{CollectTrash}).}
    \label{tab:baseline_performance}
\end{table}

Grasping is a challenging research topic on its own. To facilitate our experiments, we adopt an assistive \texttt{pick} primitive that creates a rigid connection between the object and the gripper if grasping is requested when all fingers are in contact with the object, a stricter form of \textit{StickyMitten} used in prior works~\cite{szot2021habitat, ehsani2021manipulathor, li2020hrl4in}. Furthermore, to accelerate training, 
the action primitives check only the feasibility (e.g., reachability, collisions) of the final configuration, e.g. the grasping pose for \texttt{pick} or the desired location for \texttt{navigate}. If kinematically feasible, the action primitives will directly set the robot state to the final configuration, and continue to simulate from then on. We include an ablation analysis of the effect of these assumptions and simplifications in our evaluation (see Table~\ref{tab:baseline_ablation}).
Further details about our training and evaluation setup can be found in Appendix~\ref{sec:baseline_appendix}.


\paragraph{Results: Task Completion.} Table~\ref{tab:baseline_performance} contains task success rates across our baseline methods. The extreme long-horizon in our activities causes the visuomotor control (RL-VMC) policy to fail in all three activities, potentially due to problems such as credit assignment~\cite{alipov2021towards}, deep exploration~\cite{yang2021exploration,JMLR:v20:18-339}, and vanishing gradients~\cite{10.1142/S0218488598000094} as reported by prior works.
Our baselines with time-extended action primitives (\rlprim and \rlprimhist) obtain better success, achieving over 40\% success rates across all three activities. We observe that longer-horizon activities are more challenging: while \texttt{CleanTable} can be accomplished by executing the optimal sequence of 6 primitive steps, \texttt{CollectTrash} requires at least $16$.
This supports the idea that some form of action-space abstraction must be necessary to solve long-horizon activities of \benchmark, as others reported~\cite{srivastava2021behavior, szot2021habitat,xia2020relmogen}.
When analyzing the role of memory, we observe a sizable performance gain from \rlprim to \rlprimhist, especially in long-horizon activities with aliased observations such as \texttt{CollectTrash}. In this task, when the robot is looking at the trash bin, it needs additional information to know what location has been cleaned already in order to proceed to other locations.
Our results indicate that memory will play a critical role for embodied AI in long-horizon \benchmark activities.

\paragraph{Results: Efficiency.} In addition to success, efficiency is also critical in the evaluation of embodied AI: a successful policy in simulation may be infeasible in the real world if it takes a long time or wastes too much energy. In Table~\ref{tab:baseline_metric}, we report the results with three efficiency metrics proposed by Srivastava et al.~\cite{srivastava2021behavior}. We observe that the use of memory (\rlprimhist) improves efficiency across all metrics: distance navigated (Dist.~Nav.), simulated time (Sim.~Time), and kinematic object disarrangement (Kin.~Dis.), i.e., amount of object displacement due to robot motion.

We also evaluate to what extent the simplifications we introduce in physics and actuation (grasping, motion execution) during training impact the performance of \rlprim during evaluation when these simplifications are removed. We report the results in Table~\ref{tab:baseline_ablation}. 
We observe a radical performance drop after enabling fully physics-based grasping during evaluation. Grasping is thus a critical component of any embodied AI task, and researchers should be careful when simplifying its execution during training. While \simulator supports fully physics-based grasping,  designing a \texttt{pick} action primitive for arbitrary objects that leverages fully physics-based grasping is by itself an open research problem that we leave for future work. In contrast, there is much less performance drop after enabling full trajectory motion execution during evaluation. This result supports our hypothesis that it is reasonable to accelerate the training process by assuming that motion planning is likely to provide viable paths in free space during evaluation.

\begin{table}[t]
\centering
\begin{minipage}[t]{.41\linewidth}%
\resizebox{\textwidth}{!}{%
\begin{tabular}{l|ccc}%
\toprule
\multicolumn{1}{c|}{\multirow{2}{*}{Method}} & \multicolumn{3}{c}{Metrics in \texttt{CollectTrash}} \\
\multicolumn{1}{c|}{} & Dist. Nav. [m] & Sim. Time [s] & Kin. Dis. [m] \\ \midrule
\rlvmc & $27.58 \pm 5.95$  & $16.67 \pm 0.00$ & $0.00 \pm 0.00$ \\
\rlprim & $17.98 \pm 2.35$ & $13.95 \pm 5.14$ & $12.34 \pm 5.01$  \\
\rlprimhist & $15.33 \pm 2.70$  & $12.48 \pm 3.68$  & $10.82 \pm 3.90$  \\ \bottomrule
\end{tabular}}
\caption{Efficiency metrics across three baseline methods. \rlvmc has low spatial and temporal efficiency because it fails to learn, whereas history information helps remove redundant actions and improve efficiency.}
\label{tab:baseline_metric}%
\end{minipage}%
\hfill
\begin{minipage}[t]{.56\linewidth}%
\resizebox{\textwidth}{!}{%
\begin{tabular}{cc|ccc}%
\toprule
\multicolumn{2}{c|}{Phys. Realism} & \multicolumn{3}{c}{Task success rate} \\
Grasping & Full Motion & \texttt{StoreDecoration} & \texttt{CollectTrash} & \texttt{CleanTable} \\ \midrule
\textcolor{indiagreen}\faCheck & \textcolor{indiagreen}\faCheck & $0.0 \pm 0.0$ & $0.0 \pm 0.0$ & $0.0 \pm 0.0$ \\
\textcolor{red}\faTimes & \textcolor{indiagreen}\faCheck & $0.46 \pm 0.04$ & $0.36 \pm 0.08$ & $0.73 \pm 0.03$ \\ 
\textcolor{red}\faTimes & \textcolor{red}\faTimes & $0.48 \pm 0.06$ & $0.42 \pm 0.02$ & $0.77 \pm 0.08$ \\ \bottomrule
\end{tabular}}
\caption{Ablation study of \rlprim on the impact of removing the simplifying assumptions of grasping and motion execution during evaluation. We observe a large drop in performance when enabling fully physics-based grasping, but not when enabling full trajectory motion execution.}
\label{tab:baseline_ablation}
\end{minipage}
\vspace{10pt}
\end{table}

\subsection{Evaluating \benchmark Solutions on a Real Robot}
\label{sec:sim2real}

\begin{figure}[t!]
    \centering
    \includegraphics[width=0.27\textwidth,valign=c]{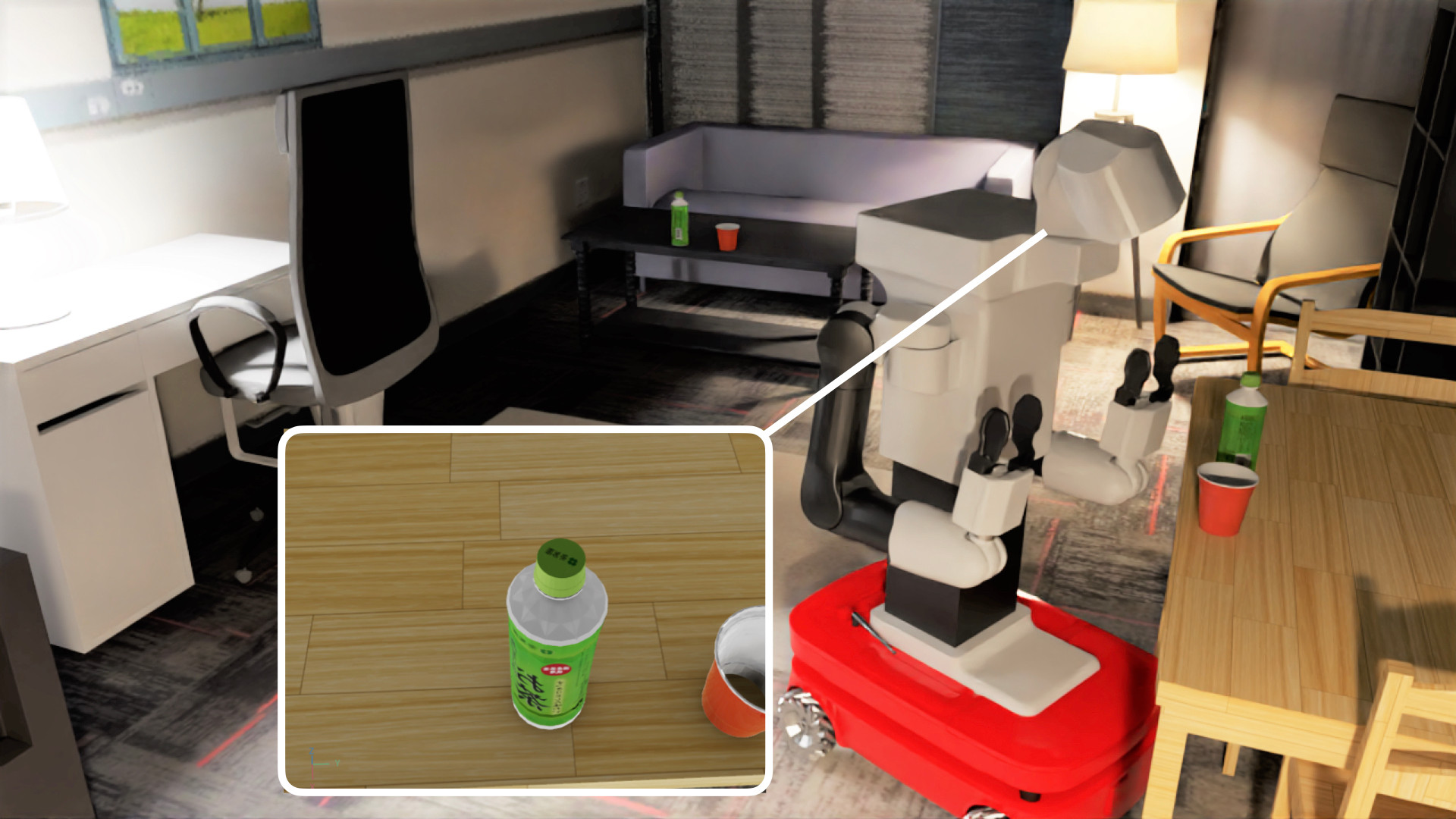}
    \hfill
    \includegraphics[width=0.27\textwidth,valign=c]{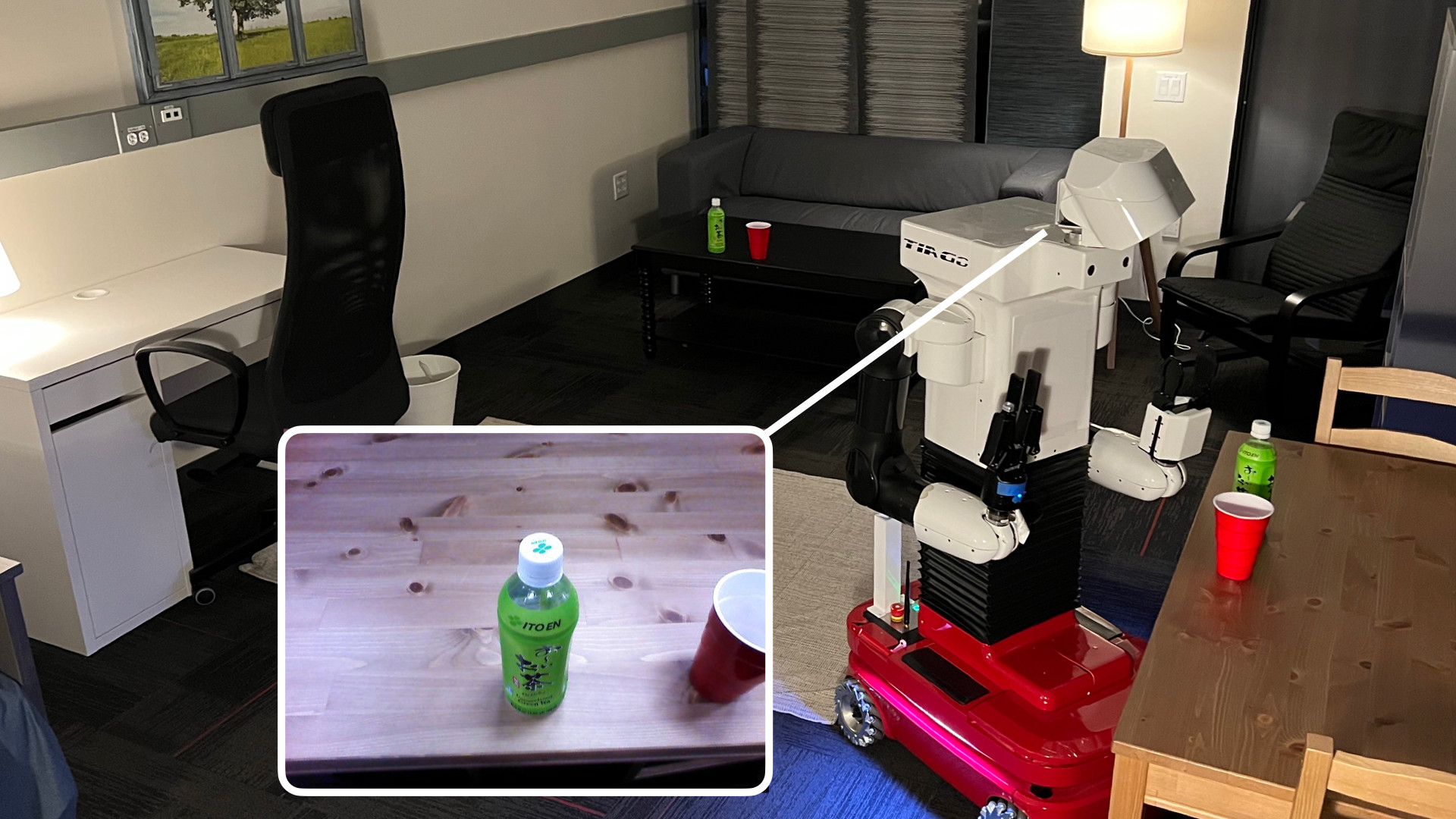}
    \hfill
    \includegraphics[width=0.44\textwidth,valign=c]{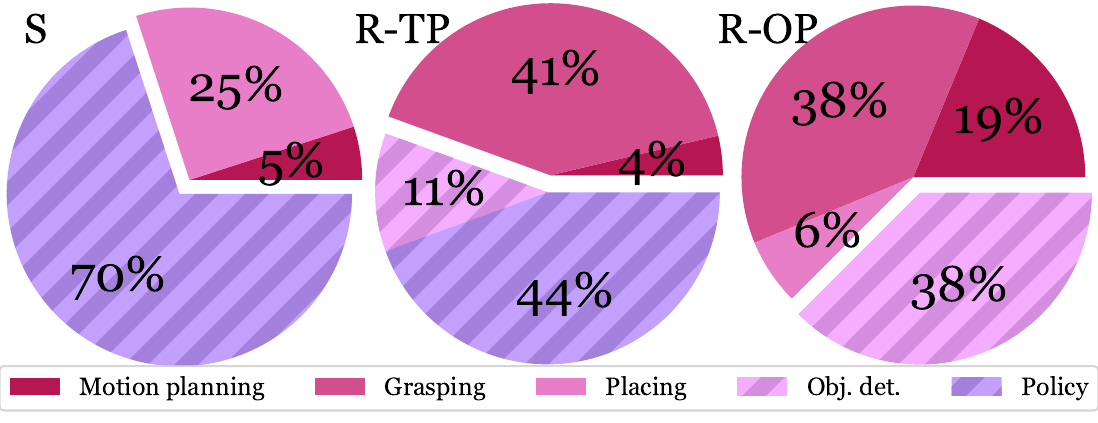}
    \caption{\textbf{Characterizing the Sim-Real Gap:} Left: a side-by-side comparison of the simulated and the real scene, including virtual and real images obtained by the robot. While the high-resolution images are extremely similar, the mismatch in wooden texture and camera properties causes a sizable gap in the visual input to the agent. Right: source of failure in \texttt{S}imulation (\texttt{S}, left) and in \texttt{R}eal-world with a \texttt{T}rained \texttt{P}olicy in \simulator (\texttt{R-TP}, middle) and an \texttt{O}ptimal \texttt{P}olicy (\texttt{R-OP}, right) due to actuation (solid color) or perception (striped). In simulation, without full simulation of grasping (see Sec.~\ref{ss:ev_in_sim}), policy failures (i.e., selecting the wrong action primitive) dominate. On the real robot, grasping is one of the main causes of error, as well as perception issues (policy errors with the trained visual policy, object detection errors with the optimal policy).}
    \label{fig:sim2real_setup}
\end{figure}


We performed a series of experiments with a real robot to answer the question: \textit{what are the main sources of discrepancy between our realistic simulation and the real world?}
To that end, we used a real-world counterpart of the simulated scene of a mockup apartment for the \texttt{CollectTrash} activity. We scanned the apartment and converted it into a virtual, interactive scene. 
We use a real bi-manual mobile manipulator Tiago, and leverage the RGB-D images from its onboard sensors and a YOLOv3 object detector~\cite{redmon2018yolov3,bjelonicYolo2018} to localize the objects in 3D space for manipulation. 
For navigation, the robot localizes with a particle filter~\cite{thrun2005probabilistic} based on two LiDAR sensors and a map of the apartment.
The action primitives are implemented with the same sampling-based motion planning algorithm as in simulation~\cite{kuffner2000rrt,Jordan.Perez.ea:CSAIL13} with additional tuning. 
We evaluate two strategies for selecting action primitives in the real world: an optimal policy based on human input, and a vision-based policy (\rlprim) trained in \simulator. To facilitate sim-to-real transfer, during training we additionally applied image-based data augmentation to the observations based on a prior work~\cite{fan2021secant} (see Appendix~\ref{as:s2r} for further details). With the optimal policy, we evaluate the gap in actuation between the simulated and the real robot; with the learned policy, we also evaluate the gap in visual perception. We achieve different success rates in simulation (50 runs, $\sim$40\% success) and in the real world with optimal (27 runs, $\sim$22\%) and trained policies (26 runs, 0\%), hinting at a sim-real gap that we analyze below.

The failure cases are depicted in Fig.~\ref{fig:sim2real_setup} (right). We observe that the majority of failures in simulation are due to the visual policy (perception), while others are caused by stochasticity in the \texttt{place} primitive and the sampling-based motion planner. The reason why none of the failures are due to grasping is that, in simulation, we evaluate with the assistive \texttt{pick} primitive. 
Grasping is fairly difficult in the real world, contributing to around 40\% of the failures for both the trained and the optimal policy. 
For the learned policy, 44\% of the errors come from the visual policy selecting the wrong action primitive due to the differences between the simulated and the real images. The visual discrepancy results from unmodeled effects such as the real camera's poor dynamic range (see Fig.~\ref{fig:sim2real_setup}, left and middle) 
and imperfect object modeling (e.g. the exact wooden texture and the surface reflectivity of the tables), which can be alleviated by more targeted domain randomization.
Interestingly, several manipulation failures on the real robot are caused by unfavorable robot base placement resulting from navigation inaccuracies in the previous timestep. This compounding source of error is not present in simulation because we assume perfect localization and execution. 

We believe this analysis provides relevant information about the main sources and severity of the sim-real gap in \benchmark in \simulator, and provide insights for future research avenues.
Our plan is to use some of these insights to create novel sim-to-real solutions that make progress on \benchmark.


\section{Discussion and Limitations}
\label{sec:discuss}
We have presented \benchmark, a benchmark for embodied AI and robotics research with realistic simulation of 1000 diverse activities grounded in human needs. \benchmark comprises two elements: \dataset, a semantic knowledge base of everyday activities, and a large-scale 3D model library; \simulator, a simulation environment that provides realistic rendering and physics for rigid/deformable objects, flexible materials and fluids. In our evaluation, we observed that \benchmark is an extremely challenging benchmark: solving these 1,000 activities autonomously is beyond the capability of current state-of-the-art AI algorithms. We studied and attempted to solve a handful of the activities with action primitives in order to gain insights into the most challenging components, providing a starting point for other researchers to work on our benchmark. Similarly, we explored the sources of the sim-real gap by creating a digital twin of a real-world mock apartment, and by performing rigorous evaluation and analysis of policies in both simulation and the real world with a simulated and real mobile manipulator. 


\paragraph{Limitations:} We inherit several limitations from our underlying physics and rendering engine, Nvidia's Omniverse. In \simulator, we trade off rendering speed for visual realism (ray-traced), reaching around 60 fps for a house scene of around 60 objects (v.s. around 100 fps in iGibson 2.0~\cite{li2021ig2}). We are actively working on performance optimization. 
Another limitation is that we only include activities that do not require interactions with humans. Realistic simulation of humans (behavior, motion, appearance) is extremely challenging and an open research area. We plan to include simulated humans as technology matures. Finally, there is still room for improvement in \simulator to further facilitate sim2real transfer, such as incorporating noise models of perception and actuation.



\acknowledgments{This work was done in part when Chengshu Li, Josiah Wong, and Michael Lingelbach were interns at Nvidia Research. The work is in part supported by the Stanford Institute for Human-Centered AI (HAI), the Toyota Research Institute (TRI), NSF CCRI \#2120095, NSF RI \#2211258, NSF NRI \#2024247, ONR MURI N00014-22-1-2740, ONR MURI N00014-21-1-2801, Amazon, Bosch, Salesforce, and Samsung. Ruohan Zhang is partially supported by the Wu Tsai Human Performance Alliance Fellowship. Sanjana Srivastava is partially supported by the National Science Foundation Graduate Research Fellowship Program (NSF GRFP).}


\bibliography{ref}  

\begin{thebibliography}{85}
\providecommand{\natexlab}[1]{#1}
\providecommand{\url}[1]{\texttt{#1}}
\expandafter\ifx\csname urlstyle\endcsname\relax
  \providecommand{\doi}[1]{doi: #1}\else
  \providecommand{\doi}{doi: \begingroup \urlstyle{rm}\Url}\fi

\bibitem[Deng et~al.(2009)Deng, Dong, Socher, Li, Li, and Fei-Fei]{deng2009imagenet}
J.~Deng, W.~Dong, R.~Socher, L.-J. Li, K.~Li, and L.~Fei-Fei.
\newblock Imagenet: A large-scale hierarchical image database.
\newblock In \emph{IEEE Conference on Computer Vision and Pattern Recognition}, pages 248--255, 2009.

\bibitem[Lin et~al.(2014)Lin, Maire, Belongie, Hays, Perona, Ramanan, Doll{\'a}r, and Zitnick]{lin2014microsoft}
T.-Y. Lin, M.~Maire, S.~Belongie, J.~Hays, P.~Perona, D.~Ramanan, P.~Doll{\'a}r, and C.~L. Zitnick.
\newblock Microsoft coco: Common objects in context.
\newblock In \emph{European Conference on Computer Vision}, pages 740--755. Springer, 2014.

\bibitem[Everingham et~al.(2010)Everingham, Van~Gool, Williams, Winn, and Zisserman]{everingham2010pascal}
M.~Everingham, L.~Van~Gool, C.~K. Williams, J.~Winn, and A.~Zisserman.
\newblock The pascal visual object classes (voc) challenge.
\newblock \emph{International Journal of Computer Vision}, 88\penalty0 (2):\penalty0 303--338, 2010.

\bibitem[Krishna et~al.(2017)Krishna, Zhu, Groth, Johnson, Hata, Kravitz, Chen, Kalantidis, Li, Shamma, et~al.]{krishna2017visual}
R.~Krishna, Y.~Zhu, O.~Groth, J.~Johnson, K.~Hata, J.~Kravitz, S.~Chen, Y.~Kalantidis, L.-J. Li, D.~A. Shamma, et~al.
\newblock Visual genome: Connecting language and vision using crowdsourced dense image annotations.
\newblock \emph{International Journal of Computer Vision}, 123\penalty0 (1):\penalty0 32--73, 2017.

\bibitem[Geiger et~al.(2012)Geiger, Lenz, and Urtasun]{geiger2012we}
A.~Geiger, P.~Lenz, and R.~Urtasun.
\newblock Are we ready for autonomous driving? the kitti vision benchmark suite.
\newblock In \emph{IEEE Conference on Computer Vision and Pattern Recognition}, pages 3354--3361. IEEE, 2012.

\bibitem[Goyal et~al.(2017)Goyal, Ebrahimi~Kahou, Michalski, Materzynska, Westphal, Kim, Haenel, Fruend, Yianilos, Mueller-Freitag, et~al.]{goyal2017something}
R.~Goyal, S.~Ebrahimi~Kahou, V.~Michalski, J.~Materzynska, S.~Westphal, H.~Kim, V.~Haenel, I.~Fruend, P.~Yianilos, M.~Mueller-Freitag, et~al.
\newblock The" something something" video database for learning and evaluating visual common sense.
\newblock In \emph{Proceedings of the IEEE International Conference on Computer Vision}, pages 5842--5850, 2017.

\bibitem[Sigurdsson et~al.(2018)Sigurdsson, Gupta, Schmid, Farhadi, and Alahari]{sigurdsson2018actor}
G.~A. Sigurdsson, A.~Gupta, C.~Schmid, A.~Farhadi, and K.~Alahari.
\newblock Actor and observer: Joint modeling of first and third-person videos.
\newblock In \emph{Proceedings of the IEEE Conference on Computer Vision and Pattern Recognition}, pages 7396--7404, 2018.

\bibitem[Xiang et~al.(2017)Xiang, Schmidt, Narayanan, and Fox]{xiang2017posecnn}
Y.~Xiang, T.~Schmidt, V.~Narayanan, and D.~Fox.
\newblock Posecnn: A convolutional neural network for 6d object pose estimation in cluttered scenes.
\newblock \emph{arXiv preprint arXiv:1711.00199}, 2017.

\bibitem[Mart\'in-Mart\'in et~al.(2021)Mart\'in-Mart\'in, Patel, Rezatofighi, Shenoi, Gwak, Frankel, Sadeghian, and Savarese]{martin2021jrdb}
R.~Mart\'in-Mart\'in, M.~Patel, H.~Rezatofighi, A.~Shenoi, J.~Gwak, E.~Frankel, A.~Sadeghian, and S.~Savarese.
\newblock Jrdb: A dataset and benchmark of egocentric robot visual perception of humans in built environments.
\newblock \emph{IEEE Transactions on Pattern Analysis and Machine Intelligence}, 2021.

\bibitem[Caba~Heilbron et~al.(2015)Caba~Heilbron, Escorcia, Ghanem, and Carlos~Niebles]{caba2015activitynet}
F.~Caba~Heilbron, V.~Escorcia, B.~Ghanem, and J.~Carlos~Niebles.
\newblock Activitynet: A large-scale video benchmark for human activity understanding.
\newblock In \emph{Proceedings of the IEEE Conference on Computer Vision and Pattern Recognition}, pages 961--970, 2015.

\bibitem[Gurari et~al.(2018)Gurari, Li, Stangl, Guo, Lin, Grauman, Luo, and Bigham]{gurari2018vizwiz}
D.~Gurari, Q.~Li, A.~J. Stangl, A.~Guo, C.~Lin, K.~Grauman, J.~Luo, and J.~P. Bigham.
\newblock Vizwiz grand challenge: Answering visual questions from blind people.
\newblock In \emph{Proceedings of the IEEE Conference on Computer Vision and Pattern Recognition}, pages 3608--3617, 2018.

\bibitem[Marcinkiewicz(1994)]{marcinkiewicz1994building}
M.~A. Marcinkiewicz.
\newblock Building a large annotated corpus of english: The penn treebank.
\newblock \emph{Using Large Corpora}, page 273, 1994.

\bibitem[Wang et~al.(2018)Wang, Singh, Michael, Hill, Levy, and Bowman]{wang2018glue}
A.~Wang, A.~Singh, J.~Michael, F.~Hill, O.~Levy, and S.~R. Bowman.
\newblock Glue: A multi-task benchmark and analysis platform for natural language understanding.
\newblock \emph{arXiv preprint arXiv:1804.07461}, 2018.

\bibitem[Rajpurkar et~al.(2016)Rajpurkar, Zhang, Lopyrev, and Liang]{rajpurkar2016squad}
P.~Rajpurkar, J.~Zhang, K.~Lopyrev, and P.~Liang.
\newblock Squad: 100,000+ questions for machine comprehension of text.
\newblock \emph{arXiv preprint arXiv:1606.05250}, 2016.

\bibitem[Socher et~al.(2013)Socher, Perelygin, Wu, Chuang, Manning, Ng, and Potts]{socher2013recursive}
R.~Socher, A.~Perelygin, J.~Wu, J.~Chuang, C.~D. Manning, A.~Y. Ng, and C.~Potts.
\newblock Recursive deep models for semantic compositionality over a sentiment treebank.
\newblock In \emph{Proceedings of the Conference on Empirical Methods in Natural Language Processing}, pages 1631--1642, 2013.

\bibitem[Antol et~al.(2015)Antol, Agrawal, Lu, Mitchell, Batra, Zitnick, and Parikh]{antol2015vqa}
S.~Antol, A.~Agrawal, J.~Lu, M.~Mitchell, D.~Batra, C.~L. Zitnick, and D.~Parikh.
\newblock Vqa: Visual question answering.
\newblock In \emph{Proceedings of the IEEE International Conference on Computer Vision}, pages 2425--2433, 2015.

\bibitem[Batra et~al.(2020)Batra, Chang, Chernova, Davison, Deng, Koltun, Levine, Malik, Mordatch, Mottaghi, Savva, and Su]{batra2020rearrangement}
D.~Batra, A.~X. Chang, S.~Chernova, A.~J. Davison, J.~Deng, V.~Koltun, S.~Levine, J.~Malik, I.~Mordatch, R.~Mottaghi, M.~Savva, and H.~Su.
\newblock Rearrangement: A challenge for embodied ai.
\newblock \emph{arXiv preprint arXiv:2011.01975}, 2020.

\bibitem[Weihs et~al.(2021)Weihs, Deitke, Kembhavi, and Mottaghi]{weihs2021visual}
L.~Weihs, M.~Deitke, A.~Kembhavi, and R.~Mottaghi.
\newblock Visual room rearrangement.
\newblock \emph{arXiv preprint arXiv:2103.16544}, 2021.

\bibitem[Gan et~al.(2021)Gan, Zhou, Schwartz, Alter, Bhandwaldar, Gutfreund, Yamins, DiCarlo, McDermott, Torralba, et~al.]{gan2021threedworld}
C.~Gan, S.~Zhou, J.~Schwartz, S.~Alter, A.~Bhandwaldar, D.~Gutfreund, D.~L. Yamins, J.~J. DiCarlo, J.~McDermott, A.~Torralba, et~al.
\newblock The threedworld transport challenge: A visually guided task-and-motion planning benchmark for physically realistic embodied ai.
\newblock \emph{arXiv preprint arXiv:2103.14025}, 2021.

\bibitem[Puig et~al.(2018)]{puig2018virtualhome}
X.~Puig et~al.
\newblock Virtualhome: Simulating household activities via programs.
\newblock In \emph{IEEE/CVF Conference on Computer Vision and Pattern Recognition}, 2018.

\bibitem[Shridhar et~al.(2020)Shridhar, Thomason, Gordon, Bisk, Han, Mottaghi, Zettlemoyer, and Fox]{shridhar2020alfred}
M.~Shridhar, J.~Thomason, D.~Gordon, Y.~Bisk, W.~Han, R.~Mottaghi, L.~Zettlemoyer, and D.~Fox.
\newblock Alfred: A benchmark for interpreting grounded instructions for everyday tasks.
\newblock In \emph{Proceedings of the IEEE/CVF conference on Computer Vision and Pattern Recognition}, pages 10740--10749, 2020.

\bibitem[Xia et~al.(2020)Xia, Shen, Li, Kasimbeg, Tchapmi, Toshev, Mart{\'i}n-Mart{\'i}n, and Savarese]{xia2020interactive}
F.~Xia, W.~B. Shen, C.~Li, P.~Kasimbeg, M.~E. Tchapmi, A.~Toshev, R.~Mart{\'i}n-Mart{\'i}n, and S.~Savarese.
\newblock Interactive gibson benchmark: A benchmark for interactive navigation in cluttered environments.
\newblock \emph{IEEE Robotics and Automation Letters}, 5\penalty0 (2):\penalty0 713--720, 2020.

\bibitem[Yu et~al.(2020)Yu, Quillen, He, Julian, Hausman, Finn, and Levine]{yu2020meta}
T.~Yu, D.~Quillen, Z.~He, R.~Julian, K.~Hausman, C.~Finn, and S.~Levine.
\newblock Meta-world: A benchmark and evaluation for multi-task and meta reinforcement learning.
\newblock In \emph{Conference on Robot Learning}, pages 1094--1100. PMLR, 2020.

\bibitem[James et~al.(2020)James, Ma, Rovick~Arrojo, and Davison]{james2019rlbench}
S.~James, Z.~Ma, D.~Rovick~Arrojo, and A.~J. Davison.
\newblock Rlbench: The robot learning benchmark \& learning environment.
\newblock \emph{IEEE Robotics and Automation Letters}, 2020.

\bibitem[Savva et~al.(2019)Savva, Kadian, Maksymets, Zhao, Wijmans, Jain, Straub, Liu, Koltun, Malik, et~al.]{savva2019habitat}
M.~Savva, A.~Kadian, O.~Maksymets, Y.~Zhao, E.~Wijmans, B.~Jain, J.~Straub, J.~Liu, V.~Koltun, J.~Malik, et~al.
\newblock Habitat: A platform for embodied ai research.
\newblock In \emph{Proceedings of the IEEE/CVF International Conference on Computer Vision}, pages 9339--9347, 2019.

\bibitem[Szot et~al.(2021)Szot, Clegg, Undersander, Wijmans, Zhao, Turner, Maestre, Mukadam, Chaplot, Maksymets, et~al.]{szot2021habitat}
A.~Szot, A.~Clegg, E.~Undersander, E.~Wijmans, Y.~Zhao, J.~Turner, N.~Maestre, M.~Mukadam, D.~S. Chaplot, O.~Maksymets, et~al.
\newblock Habitat 2.0: Training home assistants to rearrange their habitat.
\newblock In \emph{Advances in Neural Information Processing Systems}, volume~34, 2021.

\bibitem[Srivastava et~al.(2022)Srivastava, Li, Lingelbach, Mart{\'i}n-Mart{\'i}n, Xia, Vainio, Lian, Gokmen, Buch, Liu, et~al.]{srivastava2021behavior}
S.~Srivastava, C.~Li, M.~Lingelbach, R.~Mart{\'i}n-Mart{\'i}n, F.~Xia, K.~E. Vainio, Z.~Lian, C.~Gokmen, S.~Buch, K.~Liu, et~al.
\newblock Behavior: Benchmark for everyday household activities in virtual, interactive, and ecological environments.
\newblock In \emph{Conference on Robot Learning}, pages 477--490. PMLR, 2022.

\bibitem[Zhu et~al.(2020)Zhu, Wong, Mandlekar, and Mart{\'i}n-Mart{\'i}n]{zhu2020robosuite}
Y.~Zhu, J.~Wong, A.~Mandlekar, and R.~Mart{\'i}n-Mart{\'i}n.
\newblock robosuite: A modular simulation framework and benchmark for robot learning.
\newblock \emph{arXiv preprint arXiv:2009.12293}, 2020.

\bibitem[Tassa et~al.(2018)Tassa, Doron, Muldal, Erez, Li, Casas, Budden, Abdolmaleki, Merel, Lefrancq, et~al.]{tassa2018deepmind}
Y.~Tassa, Y.~Doron, A.~Muldal, T.~Erez, Y.~Li, D.~d.~L. Casas, D.~Budden, A.~Abdolmaleki, J.~Merel, A.~Lefrancq, et~al.
\newblock Deepmind control suite.
\newblock \emph{arXiv preprint arXiv:1801.00690}, 2018.

\bibitem[Brockman et~al.(2016)Brockman, Cheung, Pettersson, Schneider, Schulman, Tang, and Zaremba]{brockman2016openai}
G.~Brockman, V.~Cheung, L.~Pettersson, J.~Schneider, J.~Schulman, J.~Tang, and W.~Zaremba.
\newblock Openai gym.
\newblock \emph{arXiv preprint arXiv:1606.01540}, 2016.

\bibitem[Littman et~al.(2021)Littman, Ajunwa, Berger, Boutilier, Currie, Doshi-Velez, Hadfield, Horowitz, Isbell, Kitano, et~al.]{littman2021gathering}
M.~L. Littman, I.~Ajunwa, G.~Berger, C.~Boutilier, M.~Currie, F.~Doshi-Velez, G.~Hadfield, M.~C. Horowitz, C.~Isbell, H.~Kitano, et~al.
\newblock Gathering strength, gathering storms: The one hundred year study on artificial intelligence ({AI100}) 2021 study panel report.
\newblock Technical report, Stanford University, 2021.

\bibitem[Riedl(2019)]{riedl2019human}
M.~O. Riedl.
\newblock Human-centered artificial intelligence and machine learning.
\newblock \emph{Human Behavior and Emerging Technologies}, 1\penalty0 (1):\penalty0 33--36, 2019.

\bibitem[Xu(2019)]{xu2019toward}
W.~Xu.
\newblock Toward human-centered ai: a perspective from human-computer interaction.
\newblock \emph{Interactions}, 26\penalty0 (4):\penalty0 42--46, 2019.

\bibitem[Shneiderman(2020)]{shneiderman2020bridging}
B.~Shneiderman.
\newblock Bridging the gap between ethics and practice: guidelines for reliable, safe, and trustworthy human-centered ai systems.
\newblock \emph{ACM Transactions on Interactive Intelligent Systems (TiiS)}, 10\penalty0 (4):\penalty0 1--31, 2020.

\bibitem[Kitano et~al.(1997)Kitano, Asada, Kuniyoshi, Noda, Osawa, and Matsubara]{kitano1997robocup}
H.~Kitano, M.~Asada, Y.~Kuniyoshi, I.~Noda, E.~Osawa, and H.~Matsubara.
\newblock Robocup: A challenge problem for ai.
\newblock \emph{AI magazine}, 18\penalty0 (1):\penalty0 73--73, 1997.

\bibitem[Wisspeintner et~al.(2009)Wisspeintner, Van Der~Zant, Iocchi, and Schiffer]{wisspeintner2009robocup}
T.~Wisspeintner, T.~Van Der~Zant, L.~Iocchi, and S.~Schiffer.
\newblock Robocup{@}home: Scientific competition and benchmarking for domestic service robots.
\newblock \emph{Interaction Studies}, 10\penalty0 (3):\penalty0 392--426, 2009.

\bibitem[Iocchi et~al.(2015)Iocchi, Holz, Ruiz-del Solar, Sugiura, and Van Der~Zant]{iocchi2015robocup}
L.~Iocchi, D.~Holz, J.~Ruiz-del Solar, K.~Sugiura, and T.~Van Der~Zant.
\newblock Robocup@ home: Analysis and results of evolving competitions for domestic and service robots.
\newblock \emph{Artificial Intelligence}, 229:\penalty0 258--281, 2015.

\bibitem[Buehler et~al.(2009)Buehler, Iagnemma, and Singh]{buehler2009darpa}
M.~Buehler, K.~Iagnemma, and S.~Singh.
\newblock \emph{The DARPA Urban Challenge: Autonomous Vehicles in City Traffic}, volume~56.
\newblock springer, 2009.

\bibitem[Krotkov et~al.(2017)Krotkov, Hackett, Jackel, Perschbacher, Pippine, Strauss, Pratt, and Orlowski]{krotkov2017darpa}
E.~Krotkov, D.~Hackett, L.~Jackel, M.~Perschbacher, J.~Pippine, J.~Strauss, G.~Pratt, and C.~Orlowski.
\newblock The darpa robotics challenge finals: Results and perspectives.
\newblock \emph{Journal of Field Robotics}, 34\penalty0 (2):\penalty0 229--240, 2017.

\bibitem[Correll et~al.(2016)Correll, Bekris, Berenson, Brock, Causo, Hauser, Okada, Rodriguez, Romano, and Wurman]{correll2016analysis}
N.~Correll, K.~E. Bekris, D.~Berenson, O.~Brock, A.~Causo, K.~Hauser, K.~Okada, A.~Rodriguez, J.~M. Romano, and P.~R. Wurman.
\newblock Analysis and observations from the first amazon picking challenge.
\newblock \emph{IEEE Transactions on Automation Science and Engineering}, 15\penalty0 (1):\penalty0 172--188, 2016.

\bibitem[Eppner et~al.(2017)Eppner, H{\"o}fer, Jonschkowski, Mart{\'i}n-Mart{\'i}n, Sieverling, Wall, and Brock]{eppner2017lessons}
C.~Eppner, S.~H{\"o}fer, R.~Jonschkowski, R.~Mart{\'i}n-Mart{\'i}n, A.~Sieverling, V.~Wall, and O.~Brock.
\newblock Lessons from the amazon picking challenge: four aspects of building robotic systems.
\newblock In \emph{Proceedings of the 26th International Joint Conference on Artificial Intelligence}, pages 4831--4835, 2017.

\bibitem[Roa et~al.(2021)Roa, Dogar, Vivas, Morales, Correll, Gorner, Rosell, Foix, Memmesheimer, Ferro, et~al.]{roa2021mobile}
M.~A. Roa, M.~Dogar, C.~Vivas, A.~Morales, N.~Correll, M.~Gorner, J.~Rosell, S.~Foix, R.~Memmesheimer, F.~Ferro, et~al.
\newblock Mobile manipulation hackathon: Moving into real world applications.
\newblock \emph{IEEE Robotics \& Automation Magazine}, pages 2--14, 2021.

\bibitem[Heiden et~al.(2021)Heiden, Macklin, Narang, Fox, Garg, and Ramos]{heiden2021disect}
E.~Heiden, M.~Macklin, Y.~Narang, D.~Fox, A.~Garg, and F.~Ramos.
\newblock Disect: A differentiable simulation engine for autonomous robotic cutting.
\newblock \emph{arXiv preprint arXiv:2105.12244}, 2021.

\bibitem[Urakami et~al.(2019)Urakami, Hodgkinson, Carlin, Leu, Rigazio, and Abbeel]{urakami2019doorgym}
Y.~Urakami, A.~Hodgkinson, C.~Carlin, R.~Leu, L.~Rigazio, and P.~Abbeel.
\newblock Doorgym: A scalable door opening environment and baseline agent.
\newblock \emph{arXiv preprint arXiv:1908.01887}, 2019.

\bibitem[Lin et~al.(2020)Lin, Wang, Olkin, and Held]{lin2020softgym}
X.~Lin, Y.~Wang, J.~Olkin, and D.~Held.
\newblock Softgym: Benchmarking deep reinforcement learning for deformable object manipulation.
\newblock In \emph{Conference on Robot Learning}, 2020.

\bibitem[{Nvidia, Corp.}(2022)]{physx}
{Nvidia, Corp.}
\newblock Physx.
\newblock \url{https://developer.nvidia.com/physx-sdk}, 2022.
\newblock Accessed: 2022-06-10.

\bibitem[Schulman et~al.(2017)Schulman, Wolski, Dhariwal, Radford, and Klimov]{schulman2017proximal}
J.~Schulman, F.~Wolski, P.~Dhariwal, A.~Radford, and O.~Klimov.
\newblock Proximal policy optimization algorithms.
\newblock \emph{arXiv preprint arXiv:1707.06347}, 2017.

\bibitem[Haarnoja et~al.(2018)Haarnoja, Zhou, Abbeel, and Levine]{haarnoja2018soft}
T.~Haarnoja, A.~Zhou, P.~Abbeel, and S.~Levine.
\newblock Soft actor-critic: Off-policy maximum entropy deep reinforcement learning with a stochastic actor.
\newblock In \emph{International Conference on Machine Learning}, pages 1861--1870. PMLR, 2018.

\bibitem[Jordan and Perez(2013)]{Jordan.Perez.ea:CSAIL13}
M.~Jordan and A.~Perez.
\newblock Optimal bidirectional rapidly-exploring random trees.
\newblock Technical Report MIT-CSAIL-TR-2013-021, Computer Science and Artificial Intelligence Laboratory, Massachusetts Institute of Technology, Cambridge, MA, 2013.

\bibitem[{U.S. Bureau of Labor Statistics}(2019)]{atus}
{U.S. Bureau of Labor Statistics}.
\newblock {American Time Use Survey}.
\newblock \url{https://www.bls.gov/tus/}, 2019.

\bibitem[{European Commission}(2010)]{hetus}
{European Commission}.
\newblock Harmonised european time use surveys.
\newblock \url{https://ec.europa.eu/eurostat/web/time-use-surveys}, 2010.

\bibitem[Gershuny et~al.(2020)Gershuny, Vega-Rapun, and Lamote]{mtus}
J.~Gershuny, M.~Vega-Rapun, and J.~Lamote.
\newblock Multinational time use study.
\newblock \url{https://www.timeuse.org/mtus}, 2020.

\bibitem[{wikiHow, Inc.}(2021)]{wikihow}
{wikiHow, Inc.}
\newblock wikihow.
\newblock \url{https://www.wikihow.com}, 2021.
\newblock Accessed: 2021-06-16.

\bibitem[Xiang et~al.(2020)Xiang, Qin, Mo, Xia, Zhu, Liu, Liu, Jiang, Yuan, Wang, et~al.]{xiang2020sapien}
F.~Xiang, Y.~Qin, K.~Mo, Y.~Xia, H.~Zhu, F.~Liu, M.~Liu, H.~Jiang, Y.~Yuan, H.~Wang, et~al.
\newblock {SAPIEN}: A simulated part-based interactive environment.
\newblock In \emph{Proceedings of the IEEE/CVF Conference on Computer Vision and Pattern Recognition}, pages 11097--11107, 2020.

\bibitem[Mu et~al.(2021)Mu, Ling, Xiang, Yang, Li, Tao, Huang, Jia, and Su]{mu2021maniskill}
T.~Mu, Z.~Ling, F.~Xiang, D.~Yang, X.~Li, S.~Tao, Z.~Huang, Z.~Jia, and H.~Su.
\newblock Maniskill: Generalizable manipulation skill benchmark with large-scale demonstrations.
\newblock \emph{arXiv preprint arXiv:2107.14483}, 2021.

\bibitem[Fu et~al.(2022)Fu, Xu, Xue, Yang, Ye, Huang, Xue, Wang, and Lu]{fu2022rfuniverse}
H.~Fu, W.~Xu, H.~Xue, H.~Yang, R.~Ye, Y.~Huang, Z.~Xue, Y.~Wang, and C.~Lu.
\newblock Rfuniverse: A physics-based action-centric interactive environment for everyday household tasks.
\newblock \emph{arXiv preprint arXiv:2202.00199}, 2022.

\bibitem[Miller(1995)]{wordnet}
G.~A. Miller.
\newblock Wordnet: a lexical database for english.
\newblock \emph{Communications of the ACM}, 38\penalty0 (11):\penalty0 39--41, 1995.

\bibitem[Brown et~al.(2020)Brown, Mann, Ryder, Subbiah, Kaplan, Dhariwal, Neelakantan, Shyam, Sastry, Askell, et~al.]{gpt3}
T.~Brown, B.~Mann, N.~Ryder, M.~Subbiah, J.~D. Kaplan, P.~Dhariwal, A.~Neelakantan, P.~Shyam, G.~Sastry, A.~Askell, et~al.
\newblock Language models are few-shot learners.
\newblock In \emph{Advances in Neural Information Processing Systems}, volume~33, pages 1877--1901, 2020.

\bibitem[Li et~al.(2021)Li, Xia, Mart{\'i}n-Mart{\'i}n, Lingelbach, Srivastava, Shen, Vainio, Gokmen, Dharan, Jain, Kurenkov, Liu, Gweon, Wu, Fei-Fei, and Savarese]{li2021ig2}
C.~Li, F.~Xia, R.~Mart{\'i}n-Mart{\'i}n, M.~Lingelbach, S.~Srivastava, B.~Shen, K.~E. Vainio, C.~Gokmen, G.~Dharan, T.~Jain, A.~Kurenkov, K.~Liu, H.~Gweon, J.~Wu, L.~Fei-Fei, and S.~Savarese.
\newblock igibson 2.0: Object-centric simulation for robot learning of everyday household tasks.
\newblock In \emph{Annual Conference on Robot Learning}, 2021.

\bibitem[Barto and Mahadevan(2003)]{barto2003recent}
A.~G. Barto and S.~Mahadevan.
\newblock Recent advances in hierarchical reinforcement learning.
\newblock \emph{Discrete Event Dynamic Systems}, 13\penalty0 (1):\penalty0 41--77, 2003.

\bibitem[LaValle(2006)]{lavalle2006planning}
S.~M. LaValle.
\newblock \emph{Planning Algorithms}.
\newblock Cambridge University Press, 2006.

\bibitem[Kuffner and LaValle(2000)]{kuffner2000rrt}
J.~J. Kuffner and S.~M. LaValle.
\newblock Rrt-connect: An efficient approach to single-query path planning.
\newblock In \emph{Proceedings IEEE International Conference on Robotics and Automation}, volume~2, pages 995--1001. IEEE, 2000.

\bibitem[Ehsani et~al.(2021)Ehsani, Han, Herrasti, VanderBilt, Weihs, Kolve, Kembhavi, and Mottaghi]{ehsani2021manipulathor}
K.~Ehsani, W.~Han, A.~Herrasti, E.~VanderBilt, L.~Weihs, E.~Kolve, A.~Kembhavi, and R.~Mottaghi.
\newblock Manipulathor: A framework for visual object manipulation.
\newblock In \emph{Proceedings of the IEEE/CVF conference on Computer Vision and Pattern Recognition}, pages 4497--4506, 2021.

\bibitem[Li et~al.(2020)Li, Xia, Mart{\'i}n-Mart{\'i}n, and Savarese]{li2020hrl4in}
C.~Li, F.~Xia, R.~Mart{\'i}n-Mart{\'i}n, and S.~Savarese.
\newblock Hrl4in: Hierarchical reinforcement learning for interactive navigation with mobile manipulators.
\newblock In \emph{Conference on Robot Learning}, pages 603--616. PMLR, 2020.

\bibitem[Alipov et~al.(2021)Alipov, Simmons-Edler, Putintsev, Kalinin, and Vetrov]{alipov2021towards}
V.~Alipov, R.~Simmons-Edler, N.~Putintsev, P.~Kalinin, and D.~Vetrov.
\newblock Towards practical credit assignment for deep reinforcement learning.
\newblock \emph{arXiv preprint arXiv:2106.04499}, 2021.

\bibitem[Yang et~al.(2021)Yang, Tang, Bai, Liu, Hao, Meng, and Liu]{yang2021exploration}
T.~Yang, H.~Tang, C.~Bai, J.~Liu, J.~Hao, Z.~Meng, and P.~Liu.
\newblock Exploration in deep reinforcement learning: a comprehensive survey.
\newblock \emph{arXiv preprint arXiv:2109.06668}, 2021.

\bibitem[Osband et~al.(2019)Osband, Roy, Russo, and Wen]{JMLR:v20:18-339}
I.~Osband, B.~V. Roy, D.~J. Russo, and Z.~Wen.
\newblock Deep exploration via randomized value functions.
\newblock \emph{Journal of Machine Learning Research}, 20\penalty0 (124):\penalty0 1--62, 2019.

\bibitem[Hochreiter(1998)]{10.1142/S0218488598000094}
S.~Hochreiter.
\newblock The vanishing gradient problem during learning recurrent neural nets and problem solutions.
\newblock \emph{Int. J. Uncertain. Fuzziness Knowl.-Based Syst.}, 6\penalty0 (2):\penalty0 107–116, 1998.

\bibitem[Xia et~al.(2020)Xia, Li, Mart{\'i}n-Mart{\'i}n, Litany, Toshev, and Savarese]{xia2020relmogen}
F.~Xia, C.~Li, R.~Mart{\'i}n-Mart{\'i}n, O.~Litany, A.~Toshev, and S.~Savarese.
\newblock {ReLMoGen}: Leveraging motion generation in reinforcement learning for mobile manipulation.
\newblock In \emph{IEEE International Conference on Robotics and Automation (ICRA)}. IEEE, 2020.

\bibitem[Redmon and Farhadi(2018)]{redmon2018yolov3}
J.~Redmon and A.~Farhadi.
\newblock Yolov3: An incremental improvement.
\newblock \emph{arXiv preprint arXiv:1804.02767}, 2018.

\bibitem[Bjelonic(2016--2018)]{bjelonicYolo2018}
M.~Bjelonic.
\newblock {YOLO ROS}: Real-time object detection for {ROS}.
\newblock \url{https://github.com/leggedrobotics/darknet_ros}, 2016--2018.

\bibitem[Thrun et~al.(2005)Thrun, Burgard, and Fox]{thrun2005probabilistic}
S.~Thrun, W.~Burgard, and D.~Fox.
\newblock \emph{Probabilistic Robotics}.
\newblock MIT Press, 2005.

\bibitem[Fan et~al.(2021)Fan, Wang, Huang, Yu, Fei-Fei, Zhu, and Anandkumar]{fan2021secant}
L.~Fan, G.~Wang, D.-A. Huang, Z.~Yu, L.~Fei-Fei, Y.~Zhu, and A.~Anandkumar.
\newblock Secant: Self-expert cloning for zero-shot generalization of visual policies.
\newblock \emph{arXiv preprint arXiv:2106.09678}, 2021.

\bibitem[Koupaee and Wang(2018)]{koupaee2018wikihow}
M.~Koupaee and W.~Y. Wang.
\newblock Wikihow: A large scale text summarization dataset.
\newblock \emph{arXiv preprint arXiv:1810.09305}, 2018.

\bibitem[Likert(1932)]{likert1932technique}
R.~Likert.
\newblock A technique for the measurement of attitudes.
\newblock \emph{Archives of psychology}, 1932.

\bibitem[Montgomery(2013)]{stats2013montgomery}
D.~C. Montgomery.
\newblock \emph{Design and analysis of experiments}.
\newblock John Wiley \& Sons, Inc., Hoboken, NJ, eighth edition, 2013.

\bibitem[Paolacci et~al.(2010)Paolacci, Chandler, and Ipeirotis]{paolacci2010running}
G.~Paolacci, J.~Chandler, and P.~G. Ipeirotis.
\newblock Running experiments on amazon mechanical turk.
\newblock \emph{Judgment and Decision making}, 5\penalty0 (5):\penalty0 411--419, 2010.

\bibitem[Center(2016)]{pew2016mturkdemog}
P.~R. Center.
\newblock Research in the crowdsourcing age, a case study.
\newblock Technical report, Washington, D.C., July 2016.
\newblock URL \url{https://www.pewresearch.org/internet/2016/07/11/research-in-the-crowdsourcing-age-a-case-study/}.

\bibitem[Akbik et~al.(2018)Akbik, Blythe, and Vollgraf]{akbik2018coling}
A.~Akbik, D.~Blythe, and R.~Vollgraf.
\newblock Contextual string embeddings for sequence labeling.
\newblock In \emph{{COLING} 2018, 27th International Conference on Computational Linguistics}, pages 1638--1649, 2018.

\bibitem[Fox and Long(2003)]{fox2003pddl21}
M.~Fox and D.~Long.
\newblock {PDDL}2.1: An extension to {PDDL} for expressing temporal planning domains.
\newblock \emph{Journal of Artificial Intelligence Research}, 20:\penalty0 61--124, dec 2003.
\newblock \doi{10.1613/jair.1129}.
\newblock URL \url{https://doi.org/10.1613%2Fjair.1129}.

\bibitem[Wichlacz et~al.(2019)Wichlacz, \'{A}lvaro Torralba, and Hoffmann]{wichlacz2019ghost}
J.~Wichlacz, \'{A}lvaro Torralba, and J.~Hoffmann.
\newblock Construction-planning models in minecraft.
\newblock In \emph{Proceedings of the 2nd ICAPS Workshop on Hierarchical Planning (HPlan 2019)}, pages 1--5, 2019.

\bibitem[{TurboSquid, Inc.}(2022)]{turbosquid}
{TurboSquid, Inc.}
\newblock Turbosquid.
\newblock \url{https://www.turbosquid.com/}, 2022.
\newblock Accessed: 2022-06-24.

\bibitem[{Niantic, Inc.}(2022)]{scaniverse}
{Niantic, Inc.}
\newblock Scaniverse.
\newblock \url{https://scaniverse.com}, 2022.
\newblock Accessed: 2022-06-24.

\bibitem[Quigley et~al.(2009)Quigley, Conley, Gerkey, Faust, Foote, Leibs, Wheeler, Ng, et~al.]{quigley2009ros}
M.~Quigley, K.~Conley, B.~Gerkey, J.~Faust, T.~Foote, J.~Leibs, R.~Wheeler, A.~Y. Ng, et~al.
\newblock Ros: an open-source robot operating system.
\newblock In \emph{ICRA workshop on open source software}, volume~3, page~5. Kobe, Japan, 2009.

\bibitem[Baboolall et~al.(2019)Baboolall, Pinder, and Stewart]{mckinsey2019automation}
D.~Baboolall, D.~Pinder, and S.~Stewart.
\newblock How automation could affect employment for women in the united kingdom and minorities in the united states.
\newblock \emph{McKinsey Digital}, 2019.
\newblock URL \url{tinyurl.com/h883dm9n}.

\end{thebibliography}


\appendix
\setcounter{figure}{0} 
\setcounter{table}{0} 
\renewcommand{\thetable}{A.\arabic{table}}
\renewcommand{\thefigure}{A.\arabic{figure}}

\section{Survey}
\label{as:survey}

In this section, we provide further information about the survey presented in Sec.~\ref{sec:survey} that guided our benchmark development. We'll discuss how we selected activities that are part of the survey, the survey design and execution, and the demographic information about the survey participants.

\subsection{Activity Sources} 
We source activities from a combination of \textit{time-use surveys} and WikiHow~\cite{koupaee2018wikihow}. Time-use surveys are studies that inquire about the daily time use of a population~\cite{atus}, making them a good proxy for daily requirements of embodied intelligence. We combine information from three time-use surveys\textemdash American~\cite{atus}, Harmonized European~\cite{hetus}, and Multinational~\cite{mtus} Time Use Surveys\textemdash and obtain an initial set of 540 activities. 
The time-use surveys focus on activities that happen with enough frequency and require a significant amount of time. However, there are other activities that are essential for everyday human life, but not reflected in the time-use surveys.
WikiHow articles include activities that are important for humans where they seek guidance, even if they are not as frequent as the ones included in the time-use surveys. 
This indicates a great potential to source additional useful and relevant daily activities. 
We complement the activities from the time use surveys with WikiHow article titles, of which there are 180,000+. These constitute a raw set of activities that are filtered down by feasibility, as explained in the next paragraph. 

\paragraph{Activity filtering for feasibility:} Simulation constrains which activities collected from time-use surveys and WikiHow can actually be used in \benchmark. We used the following criteria based on \simulator and \bddl constraints to filter the activity pool prior to the survey: 

\begin{table}[h]
    \centering
    \label{tab:xxxx}
    \resizebox{\textwidth}{!}{%
\begin{tabular}{|l|l|}
    \hline
    \textbf{Filtering principle} & 
    \textbf{Example activity filtered out} \\
    \hline
    Activity requires physics or chemistry not supported in simulation & 
        \texttt{SteamingClothes}, \texttt{MakingSoap} \\
    Activity involves creating or consuming media & 
        \texttt{ReadingABook} \\
    Activity requires more than a day in real time & 
        \texttt{DryingSeedsOvernight} \\
    Activity requires non-visual perceptual modalities & 
        \texttt{SweeteningFood} \\
    Activity requires geometric configurations too fine-grained for \bddl &
        \texttt{SettingUpANativityScene} \\
    Activity is predicated on branded items & 
        \texttt{SprayingWindex} \\
    Activity involves other people or live animals & 
        \texttt{AskingForARaise}\\
    \hline
\end{tabular}
}
\end{table}
After filtering and eliminating duplicates, and including the activities from~\cite{srivastava2021behavior}, 2,090 activities remain to be surveyed.

\subsection{Survey Design} 
Our survey is structured as follows:
\vspace{-2mm}
\begin{itemize}[leftmargin=1.5em]
    \item \textbf{Demographic questions} requesting information about the number of people in the household, occupation, general location, and relationship between household work, automation, and livelihood (see Fig.~\ref{fig:activity_survey_demo} for results). 
    \item \textbf{Activity survey questions} requesting the value of automation of a batch of activities to the respondent. Specifically, this section comprises:
    \begin{itemize}
        \item 50 questions, one question per activity 
        \item Question text: \textit{On a scale of 1 (left) to 10 (right), rate how much you want a robot to do this activity for you.}
        \item Each response uses an independent Likert score~\cite{likert1932technique} on a scale from 1 (less beneficial) to 10 (most beneficial)
    \end{itemize}
\end{itemize}

We piloted alternatives for the survey about two design decisions: 1) question wording, and 2) question format. With respect to the wording for the question, we piloted three alternatives to specify the \textit{agent}: \textit{robot}, \textit{assistant}, and \textit{automation}. Our goal was to study any possible (mis)conceptions about robots that might bias the study. By way of 30 pairwise T-tests and 10 ANOVAS~\cite{stats2013montgomery}, we did not find any statistically significant difference between the results obtained using these three wordings. With respect to the format of the question, we piloted two alternative formats: 10-point Likert, and three-element best-worst scaling. By way of Kendall's tau~\cite{stats2013montgomery}, there was a strong correlation between the rankings determined by Likert scores and standard metrics of best-worst scaling. We, therefore, conclude that either method will give us a similar ranking, and select the more resource-efficient option (Likert). 

\paragraph{Survey collection:}
We deployed our survey on Amazon Mechanical Turk~\cite{paolacci2010running} and collected 50 unique responses per activity. There were a total of 1,461 different respondents and their average scores ranged from 1.9 to 9.3, showing high diversity. The average score was 5.16. To ensure response quality, we repeated four questions in every survey as an attention check. If responses to a pair of repeats differed by more than two points, we considered it failed. Two or more failures led to rejection. We also rejected survey responses that had no significant variance across their responses to 50 different activities.


\subsection{Demographic Information}
\label{sec:demographics}

\begin{figure}[t!]
\centering
\includegraphics[width=0.33\textwidth,valign=T]{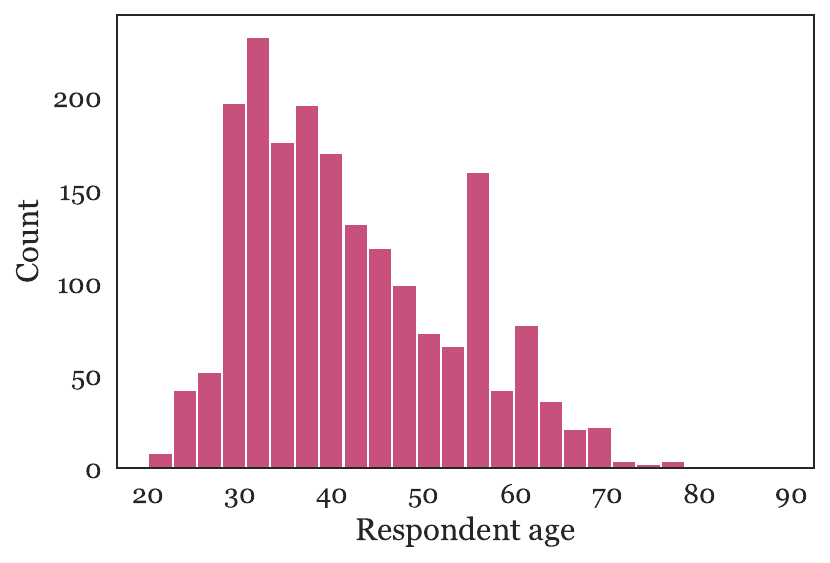}%
\hfill%
\includegraphics[width=0.33\textwidth,valign=T]{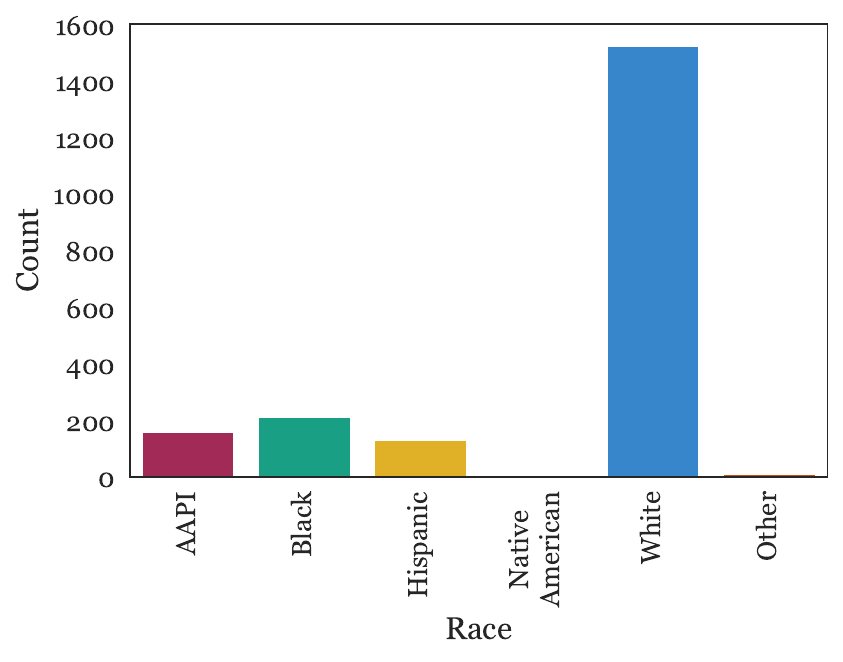}%
\hfill%
\includegraphics[width=0.33\textwidth,valign=T]{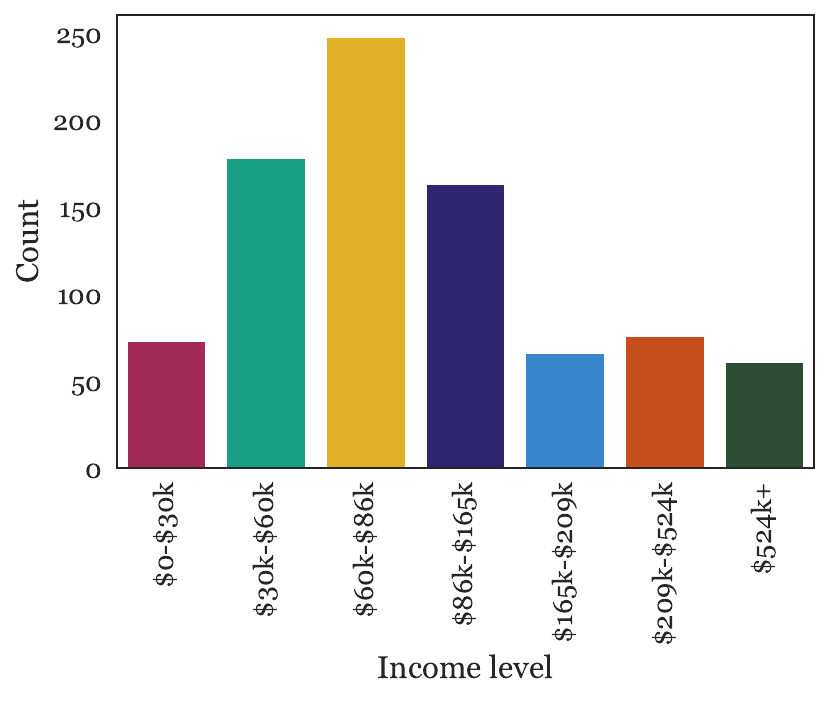}%
\caption{\textbf{Demographic distribution of the participants in our survey.} Respondents appear to have similar demographics to the wider U.S. Mechanical Turk worker population~\cite{pew2016mturkdemog}. Not pictured: respondent gender (43.41\% women, 55.50\% men, 0.83\% non-binary, 0.26\% other), disability status (92.56\% no, 5.74\% yes, 1.70\% prefer not to answer).}
\label{fig:activity_survey_demo}
\end{figure}

Fig.~\ref{fig:activity_survey_demo} depicts the results of our demographic questions on the participants of the survey. We observe that most adult age groups have good representation, particularly pre-retirement age groups, and are concentrated in the 30-40 group. Racially, the respondents are around 75\% white, a larger proportion than the U.S. population but similar to the Mechanical Turk proportion~\cite{pew2016mturkdemog}. Other races appear to have proportions similar to but not the same as their presence on both Mechanical Turk and in the general population; this may be due to their overall small numbers in all three~\cite{pew2016mturkdemog}. There is some Native American representation, though not statistically significant. Income-wise, respondents tend to be in the \$30,000-\$150,000 range, particularly in the lower half. 

The participants' gender is distributed by 43.41\% women, 55.50\% men, 0.83\% non-binary, 0.26\% other. Gender representation is not as even as the U.S. population but more so than the Mechanical Turk worker population~\cite{pew2016mturkdemog}. There is a small representation of non-binary individuals. When it comes to disability status, our participants reported 92.56\% no-disability, 5.74\% disability, and 1.70\%
prefer not to answer. Thus, our survey contained some but limited representation of people with disabilities. This group, together with the elderly, are groups commonly assumed to potentially benefit from robotic efforts; they could be subject to future targeted surveys.

\section{Activity Annotation}
\label{as:act_ann}
\begin{figure}[t]
    \centering
    \includegraphics[width=\textwidth]{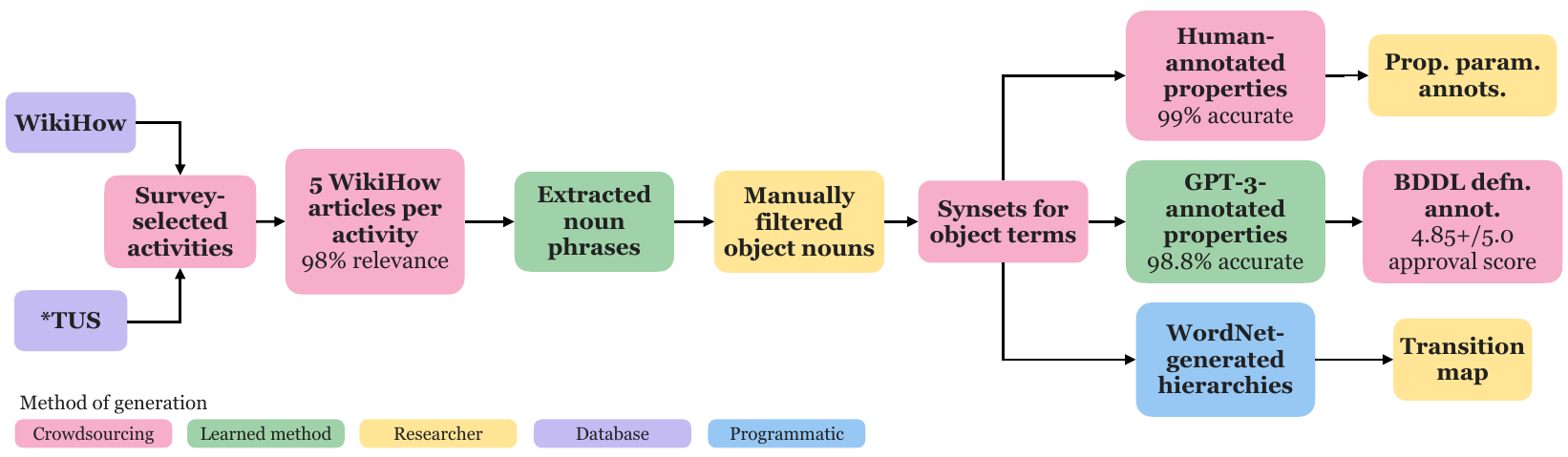}
    \caption{\textbf{Flow chart of the annotation pipeline:} activity selection, activity article collection, object list extraction, object property and parameter annotation, \bddl annotation.}
    \label{fig:activity_pipeline}
\end{figure}

Obtaining the \bddl definitions for 1,000 activities requires multiple preparatory steps to gather the knowledge needed. This knowledge also becomes part of the~\dataset. The pipeline is outlined in Fig.~\ref{fig:activity_pipeline}. We now detail each annotation other than the survey. Furthermore, quality assessment statistics for these annotations are found in Sec.~\ref{sec:annot_qa}. We see that experienced annotators find our resulting \benchmark knowledge base highly accurate.

\paragraph{Obtaining list of objects per activity (WikiHow articles, noun phrase extraction, and manual object filtering):} Our activity definition requires first defining a set of objects and properties that the annotators could use to describe an activity. We would like these domains to be natural and ecological, containing all relevant objects that humans may consider necessary for a task. To obtain such an ecologically plausible object space per activity, we parse WikiHow articles. Since WikiHow is a how-to database with a wide following and community as well as expert- and peer-review, the article texts document objects that are used and acted on in an activity. We therefore asked crowdworkers from the Upwork platform to collect articles; for robustness, we had five articles collected for each activity. We used a chunking model~\cite{akbik2018coling} to extract noun phrases from the article text, then we manually filtered the noun phrases into tangible objects. 

\paragraph{Synsets for object terms and WordNet hierarchy generation:} After obtaining an object space for each activity, we take their union and crowdworkers match each one to a WordNet~\cite{wordnet} synset. This eliminates word-sense ambiguity in the knowledge base and creates a hierarchical structure in the overall object space via the WordNet hierarchy. This process yielded 1,538 total WordNet leaf synsets. Based on various property annotations and needs of the activity set, there are also 1,426 custom synsets for a total of 2,964 leaf synsets.

\paragraph{Property annotations:} Each object that is a leaf-level synset of the WordNet hierarchy is associated with the set of object properties in \benchmark, each of which is fully simulatable in \simulator (full list of properties in Table~\ref{tab:supp_props}). The properties define which predicates can apply to the object - e.g. an object having the \texttt{cookable} property means it can be \texttt{cooked} and \texttt{not cooked}. This association is done task-agnostically because objects/states that are irrelevant or undesirable for a specific activity will still provide important learning signals. All properties that apply to many of the 2,964 leaf synsets and therefore require a large-scale annotation are done by either GPT-3~\cite{gpt3} or crowdworkers. GPT-3 is used for all properties (total of eight) for which the Hamming distance and false-positive rate for a sample compared to a human-annotated ground-truth are both less than 10\%. Human annotation is used for five more properties. The remaining properties are either sensitive to the simulator implementation and are therefore annotated manually, or can be inferred from the other annotations and are determined analytically. 

This annotation is done only for leaf-level synsets, then properties of higher-level synsets are inferred from the leaf-level annotations. This prevents definition unsolveability. In more detail: as shown in Fig.~\ref{fig:activity_pipeline}, the hierarchical structure is applied to the object space. This allows activity definition annotators to refer to them instead of leaf-level synsets; for example, a \texttt{PuttingAwayGroceries} definition can have ``five \texttt{edible\_fruit.n.01}s'' rather than ``two \texttt{apple.n.01}s and three \texttt{banana.n.01}s'', allowing more variation in activity instantiation because far more object models are valid. However, 3-D object models are generally only attached to leaf- or near-leaf-level synsets, meaning that a definition may have a higher-level synset with various predicates attached to it, and at simulation time the object model (associated with one of the synset's descendants) must have those states simulated. A problem might occur when e.g. \texttt{container.n.01} is annotated as \texttt{fillable} and the definition calls for a \texttt{container.n.01} to be \texttt{filled} with a \texttt{liquid}, but the actual model that gets sampled into the simulator to satisfy \texttt{container.n.01} is a cloth bag that doesn't support being \texttt{filled} with a \texttt{liquid}. So to avoid this unsolveability, the properties of any non-leaf synset are exactly the \textit{intersection of all its descendants' properties}.

{\scriptsize
\centering
\begin{longtable}{|l|p{.20\textwidth}|p{.25\textwidth}|p{.28\textwidth}|}
        \hline
        \textbf{Object property} & \textbf{Annotation method} & \textbf{Prompt to annotator} & \textbf{Example objects}
        \\\hline
        \texttt{assembleable} & 
            manual &
            N/A & 
             \texttt{desk.n.01},  \texttt{table.n.02}
        \\\hline
        \texttt{breakable} & 
            human &
            Mark if the object can be broken into smaller pieces by a human dropping it on the floor without a tool.  & 
              \texttt{wine\_bottle.n.01}, \texttt{room\_light.n.01} 
        \\\hline
        \texttt{cloth} & 
            manual &
            N/A & 
             \texttt{hammock.n.02}, \texttt{canvas.n.01} 
        \\\hline
        \texttt{coldSource} & 
            GPT-3 &
            Is [object] a source of cold? & 
             \texttt{refrigerator.n.01}, \texttt{ice.n.01}
        \\\hline
        \texttt{cookable} & 
            GPT-3 &
            Can a [object] be cooked? & 
             \texttt{biscuit.n.01},
            \texttt{pizza.n.01}
        \\\hline
        \texttt{deformable} & 
             prog. (all \texttt{softBody}, \texttt{cloth}, \texttt{rope}) &
            N/A & 
              \texttt{tortilla.n.01},  \texttt{clay.n.01} 
        \\\hline
        \texttt{diceable} & 
             prog. (all \texttt{sliceable}'s derivative synsets &
            N/A & 
              \texttt{half\_\_apricot.n.01},  \texttt{half\_\_brisket.n.01} 
        \\\hline
        \texttt{drapeable} & 
             prog. (all \texttt{cloth}, \texttt{rope} &
            N/A & 
              \texttt{dress.n.01},  \texttt{bath\_towel.n.01} 
        \\\hline
        \texttt{fillable} & 
             manual &
            N/A & 
              \texttt{stockpot.n.01},  \texttt{bucket.n.01} 
        \\\hline
        \texttt{fireSource} & 
            GPT-3 &
            Is a [object] designed to create fire? & 
             \texttt{lighter.n.01}, \texttt{sparkler.n.01} 
        \\\hline
        \texttt{flammable} & 
            human &
            Mark if the object can catch fire (i.e. burn with a flame). &
             \texttt{candle.n.01}, \texttt{mail.n.01}
        \\\hline
        \texttt{foldable} & 
             prog. (all \texttt{cloth}, \texttt{softBody}) &
            N/A & 
             \texttt{tortilla.n.01}, \texttt{jean\_jacket.n.01} 
        \\\hline
        \texttt{freezable} & 
             prog. (all \texttt{heatable}) &
            N/A & 
             \texttt{olive\_oil.n.01}, \mbox{\texttt{ginger\_beer.n.01}}
        \\\hline
        \texttt{heatable} & 
            prog. (all rigid bodies) &
            N/A & 
             \texttt{oil.n.01}, \texttt{fish\_knife.n.01} 
        \\\hline
        \texttt{heatSource} & 
            GPT-3 &
            Is a [object] a source of heat? & 
             \texttt{oven.n.01}, \texttt{toaster.n.01} 
        \\\hline
        \texttt{liquid} & 
            GPT-3 &
            Is a [object] a liquid? & 
             \texttt{gasoline.n.01}, \mbox{\texttt{liquid\_soap.n.01}} 
        \\\hline
        \texttt{macroPhysicalSubstance} & 
            manual &
            N/A & 
             \texttt{grated\_cheese.n.01}, \texttt{blueberry.n.02} 
        \\\hline
        \texttt{meltable} & 
            manual &
            N/A & 
             \texttt{cheese.n.01}, \texttt{chocolate.n.02} 
        \\\hline
        \texttt{microPhysicalSubstance} & 
            manual &
            N/A & 
             \texttt{brown\_sugar.n.01}, \texttt{cinnamon.n.03} 
        \\\hline
        \texttt{mixingTool} & 
            manual &
            N/A & 
             \texttt{putty\_knife.n.01}, \texttt{teaspoon.n.02} 
        \\\hline
        \texttt{needsOrientation} & 
            manual &
            N/A & 
             \texttt{cumin\_\_shaker.n.01}, \texttt{salt\_\_shaker.n.01} 
        \\\hline
        \texttt{openable} & 
            human &
            Mark if the object is designed to be opened. &
             \texttt{mixer.n.04}, \texttt{keg.n.01} 
        \\\hline
        \texttt{particleApplier} & 
            manual &
            N/A &
             \texttt{pepper\_mill.n.01}, \texttt{detergent\_\_atomizer.n.01} 
        \\\hline
        \texttt{particleRemover} & 
            GPT-3 &
            Can a [object] absorb liquid? &
             \texttt{scrub\_brush.n.01}, \texttt{broom.n.01} 
        \\\hline
        \texttt{particleSource} & 
            manual (on top of \texttt{particleApplier} &
            N/A &
             \texttt{sink.n.01}
        \\\hline
        \texttt{particleSink} & 
            manual (on top of \texttt{particleRemover} &
            N/A &
             \texttt{sink.n.01}
        \\\hline
        \texttt{rigidBody} & 
            manual &
            N/A & 
             \texttt{tomato.n.01}, \texttt{desk.n.01} 
        \\\hline
        \texttt{rope} & 
            manual &
            N/A & 
             \texttt{ribbon.n.01}, \texttt{fairy\_light.n.01} 
        \\\hline
        \texttt{sliceable} & 
            GPT-3 &
            Can a [object] be sliced easily by a human with a knife? &
             \texttt{sweet\_corn.n.01}, \mbox{\texttt{sandwich.n.01}} 
        \\\hline
        \texttt{slicingTool} & 
            GPT-3 &
            Can a [object] slice an apple? & 
             \texttt{blade.n.09}, \texttt{razor.n.01}
        \\\hline
        \texttt{softBody} & 
            manual &
            N/A & 
             \texttt{dough.n.01}, \texttt{pillow.n.01} 
        \\\hline
        \texttt{substance} & 
             prog. (all \texttt{liquid}, \texttt{vis.Subst.}, \texttt{phys.Subst.}) &
            N/A & 
             \texttt{water.n.06}, \texttt{milk.n.01} 
        \\\hline
        \texttt{toggleable} & 
            human &
            The object can be switched between a finite number of \textit{discrete} states and is designed to do so. &
            \texttt{hot\_tub.n.01}, \texttt{light\_bulb.n.01} 
        \\\hline
        \texttt{unfoldable} & 
            prog. (all \texttt{foldable}) &
            N/A & 
             \texttt{tissue.n.02}, \texttt{foil.n.01} 
        \\\hline
        \texttt{visualSubstance} & 
            manual &
            N/A & 
             \texttt{coriander.n.02}, \mbox{\texttt{cocoa\_powder.n.01}}
        \\\hline
        \texttt{waterCook} & 
            manual (special case of \texttt{cookable} &
            N/A & 
             \texttt{chickpea.n.01}, \mbox{\texttt{white\_rice.n.01}}
        \\\hline
    \caption{Properties annotated for \benchmark for each object category and included in the \dataset knowledge base}
    \label{tab:supp_props}
\end{longtable}
}

\paragraph{Property parameter annotations:} \bddl separates objects and object states (i.e. terms and predicates), but simulating realistic ``cooking'' or ``filling'' requires information for object-object property tuples, not just objects or object properties alone. For example, in the real world, an \texttt{apple.n.01} and a \texttt{chicken.n.01} do not become cooked at the same temperature, so simply knowing they are both \texttt{cookable} is insufficient. \dataset therefore includes manual annotation of several property parameters that are (object, object property) specific, such as \texttt{cookTemperature} for all \texttt{cookable} objects. 

Examples of object-property pairs and parameters:
\vspace{-2mm}
\begin{itemize}[leftmargin=2em]
    \item Temperature required for \texttt{cookable} object to go from \texttt{(cooked object) = False} to \texttt{(cooked object) = True}
    \begin{itemize}
        \item \texttt{crab.n.05}: must reach \textbf{63\textdegree C}
        \item \texttt{squash.n.02}: must reach \textbf{58\textdegree C}
         \item \texttt{meatball.n.01}: must reach \textbf{63\textdegree C}
        \item \texttt{chicken\_leg.n.01}: must reach \textbf{74\textdegree C}
    \end{itemize}
    \item Temperature generated by \texttt{heatSource} object. For \texttt{toggleable heatSource}s, this may require \texttt{toggledOn(object) = True}
    \begin{itemize}
        \item \texttt{toaster\_oven.n.01}: generates \textbf{204\textdegree C}
        \item \texttt{ember.n.01}: generates \textbf{1093\textdegree C}
        \item \texttt{hand\_blower.n.01}: generates \textbf{45\textdegree C}
        \item \texttt{coffee\_maker.n.01}: generates \textbf{93\textdegree C}
    \end{itemize}
\end{itemize}

\paragraph{Transition rules:} There are many activities that involve complex chemical and physical processes that are beyond the capability of the state-of-the-art simulation technology. For example, blending different fruits into a smoothie or sanding a rusted surface are extremely difficult to simulate, but the agent actions to complete these tasks are still within reach, e.g. placing the fruits inside a blender. Therefore, in ordet to support these type of activities, we create a set of transition rules that will be used by \simulator to bypass the underlying physics but still produce visually realistic physical transitions, e.g. smoothie particles being generated inside the blender after the blender is turned on.


Examples of transition rules: 
\vspace{-2mm}
\begin{itemize}[leftmargin=2em]
    \item Composition and decomposition of objects 
    \begin{itemize}
        \item Transition rule used in \texttt{make a strawberry slushie}
        \begin{itemize}
            \item \textbf{Inputs:} \texttt{strawberry.n.01}, \texttt{ice.n.01}, \texttt{lemon\_juice.n.01}, \texttt{agave.n.01}
            \item \textbf{Transition machine:} \texttt{blender.n.01}
            \item \textbf{Outputs:} \texttt{smoothie.n.01}
        \end{itemize}
        \item Transition rule used in \texttt{make gazpacho}
        \begin{itemize}
            \item \textbf{Inputs:} \texttt{basil.n.03}, \texttt{salt.n.02}, \texttt{black\_pepper.n.02}, \texttt{tomato\_juice.n.01}, \texttt{cucumber.n.02}, \texttt{water.n.06}, \texttt{lemon\_juice.n.01}
            \item \textbf{Transition machine:} \texttt{saucepan.n.01}
            \item \textbf{Outputs:} \texttt{gazpacho.n.01}
        \end{itemize}
    \end{itemize}
    \item Realistic cleaning rules
    \begin{itemize}
    \item Transition rule used in \texttt{CleanTheExteriorOfYourGarage}
    \begin{itemize}
        \item \textbf{Substance covering object:} \texttt{paint.n.01} or \texttt{spray\_paint.n.01}
        \item \textbf{Objects needed to remove:} \texttt{particleRemover} saturated with (\texttt{solvent.n.01} or \texttt{acetone.n.01})
    \end{itemize}
    \item Transition rule used in \texttt{CleanYourRustyGardenTools}
    \begin{itemize}
        \item \textbf{Substance covering object:} \texttt{rust.n.01} or \texttt{patina.n.01} or \texttt{incision.n.01} (simulated as particles but exposed to annotators as unary \texttt{scratched} predicate) or \texttt{tarnish.n.01} (simulated as particles but exposed to annotators as unary \texttt{tarnished} predicate)
        \item \textbf{Objects needed to remove:} \texttt{emery\_paper.n.01} or \texttt{whetstone.n.01}
    \end{itemize}
    \end{itemize}
\end{itemize}

\paragraph{Activity definitions in \bddl:} Finally, the object spaces and the relevant properties and predicates they enable are offered to lay annotators to generate \bddl definitions. Annotators build activity definitions using an annotation interface that includes a visual version of \bddl~\cite{srivastava2021behavior} (details on new features in \bddl are in  Sec.~\ref{sec:bddl}). The annotation interface enforces the requirements needed to make definitions logically solvable and well-formed. We collect one definition per activity for a total of 1,000 \bddl activity definitions.

Listings~\ref{lst:bddl3},~\ref{lst:bddl4}, and~\ref{lst:bddl5} show examples of \benchmark activity definitions, while Listing~\ref{lst:bddl1} and~\ref{lst:bddl2} show examples of \oldbenchmark definitions. We see that while the numbers of objects and literals are similar, speaking to the fact that both benchmarks have reached similar scale and detail for the activities they have, \benchmark has far larger and more detailed activity distribution. The \texttt{BakingSugarCookies} activity involves transition rules that turn ingredients in \texttt{:init} to scones in \texttt{:goal}, a process that will require the listed mixer and oven in \simulator. By contrast, cooking in \oldbenchmark was limited to single objects transitioning from \texttt{not cooked} to \texttt{cooked}; other benchmarks are similar or lack cooking entirely. \texttt{CleanYourLaundryRoom} involves specific cleansing agents to clean mold, whereas \oldbenchmark cleaning tasks only had two types of ``dirtiness'' (\texttt{stained} and \texttt{dusty}) and the only rule was to use water in the case of \texttt{stained}. \texttt{CleanTheBottomOfAnIron} also shows rust-specific cleaning (requiring emery paper).

\subsection{Quality assessment of \benchmark annotations}
\label{sec:annot_qa}

\begin{table}[h]
    \centering
     \resizebox{\textwidth}{!}{  
    \begin{tabular}{|c|c|c|c|c|}
        \hline
        & Article Collection & Object Extraction & Human Properties & GPT-3 / Machine Properties \\ 
         \hline
        Accuracy (Approval Rate)  & 0.974 & 0.968 & 0.990 & 0.988 \\ 
        \hline
        F1-score & 0.984 & 0.990 & 0.912 & 0.930 \\ 
        \hline
        False Discovery Rate & 0.026 & 0.032 & 0.031 & 0.029 \\ 
        \hline
        False Positive Rate & N/A & N/A & 0.002 & 0.003 \\ 
        \hline
    \end{tabular}}
    \caption{Quality check results for object and object property annotations: for each stage in the annotation pipeline, report human verification results. The F1-score is $\frac{2}{\texttt{precision}^{-1} + \texttt{recall}^{-1}}$. The False Discovery Rate is FDR = FP/(TP+FP) and the False Positive Rate FPR = FP/(TN+FP). The high accuracy and F1-scores and low FDRs and FPRs show that our annotation results are of high quality. }
    \label{tab:activity_qc}
\end{table}

\begin{table}[h]
    \centering
     \resizebox{10cm}{!}{  
    \begin{tabular}{|c|c|c|c|c|}
    \hline
    & \multicolumn{4}{c|}{Activity Definition} \\
        \hline
        & Question 1 & Question 2 & Question 3 & Question 4\\ 
         \hline
        Average Rating  & 4.875 & 4.942 & 4.967 & 4.975 \\ 
        \hline
        Standard Deviation & 0.331 & 0.234 & 0.364 & 0.156 \\ 
    \hline
    \end{tabular}}
    \caption{Quality check results for activity definitions. Q1: Are the objects listed relevant to the definition of the task?, Q2: Do the initial placements/locations of objects make sense?, Q3: Are the actions to perform (“goals”) relevant to the definition of the activity?, Q4: Is this definition reasonable?\\ 
    Each activity definition was evaluated on a scale of 1-5 for each of the questions, and the average score and standard deviation are presented. The increased granularity still reflects the high quality of our results.}
    \label{tab:bddl_def_qc}
\end{table}

Our quality control investigation shows us that all labeling done by crowdworkers and GPT-3 is of the highest quality. Five crowdworkers with an extensive background in data labeling for large machine learning projects, coding, or data verification affirmed the results from earlier crowdworkers. The accuracies for all the labeling tasks were all above 96\%, the F1 scores were above 91\% and the false positive and false discovery rates were between 2-3\% as shown in Table~\ref{tab:activity_qc}. We noticed that the Synset Verification is a bit lower in accuracy than the other labeling tasks. This may be due to subtleties in acceptable synset definitions: for example, a ''reasonably narrow hypernym'' of a word not found in WordNet~\cite{wordnet} is acceptable, such as a "dispenser" for a "soap dispenser", but there may be gray areas regarding what is considered "reasonably narrow" (e.g. would a "hand tool" be reasonable as a replacement for a "soap dispenser"?)

The activity definition process was evaluated with more granularity using a Likert scale and an array of questions. We found that the response values had consistently high averages (very close to the maximum score of 5) and low standard deviations as shown in Table~\ref{tab:activity_qc}. We also found no significant discrepancy for activity definition scores across topics (e.g. tasks related to "cleaning"), showing that the \bddl definitions were rated uniformly across the span of activity categories.

\section{New \bddl features}
\label{sec:bddl}

\benchmark uses an expanded version of the~\bddl used in \oldbenchmark~\cite{srivastava2021behavior}, in order to support the new set of diverse activities. There are three new features here: 1) representation of substances, 2) three-valued predicates, and 3) composition and decomposition of objects. 

\paragraph{Representation of substances:} \benchmark introduces \textit{substances}, objects that are arbitrarily subdivideable and do not obey clear instance boundaries such as \texttt{water.n.06} or \texttt{flour.n.01}. The lack of instance boundaries is difficult with traditional PDDL/\bddl: for example, if there are two bottles of orange juice where the juice inside one is called \texttt{orange\_juice.n.01\_1} and the juice inside the other is called \texttt{orange\_juice.n.01\_2}, any mixing of the particles makes it near-impossible for the agent to satisfy a \texttt{:goal} condition pertaining to an instance. Even with quantification, a condition like \texttt{exists (orange\_juice.n.01) (filled orange\_juice.n.01 glass.n.01\_3)} is still unfair: if the agent mixes particles from the two different \texttt{orange\_juice} instances such that together they \texttt{filled glass.n.01\_3} but neither instance alone has enough particles in \texttt{glass.n.01\_3} to fill it, the \texttt{:goal} will not be met.

We simply enforce that there is up to one instance of any \texttt{substance} in a definition. The annotator can still control quantity by spawning it in as many containers as desired. Unlike labeling every particle separately, this maintains a compact representation. This does not allow for some of the substances to be different from some (e.g. some of the orange juice is cold and some is hot), but we consider this an acceptable limitation.

Furthermore, particle-based objects can be computationally expensive to simulate. When an annotator uses \texttt{orange\_juice.n.01} only as a container of orange juice and not actually the particles (e.g. in a \texttt{PuttingAwayGroceries} activity), this becomes a waste of computational resources. Therefore, we introduce a custom container synset for every substance that has the same properties and WordNet hierarchy structure as \texttt{bottle.n.01} and instruct annotators to use it if all they want is the container.

\paragraph{Three-valued predicates:} A common issue in \oldbenchmark activities is that success on predicates involving naturally continuous-valued quantities is sudden and arbitrary: cracking a window a little bit suddenly changes it from \texttt{not open} to \texttt{open}, and the activity from undone to done – even though the window is not a typical human conception of ``open’’. PDDL 2.1~\cite{fox2003pddl21} has numerical fluents and derived predicates, which offer continuous states. However, this granularity may not be crucial to the high-level activities in \benchmark, and the concept is also difficult to communicate to lay annotators.

We therefore treat certain predicates as three-valued by having two Boolean predicates where the negations are the same, such as \texttt{filled} and \texttt{empty} (rather than just \texttt{not filled}). The annotators still see one Boolean predicate, in this case \texttt{filled}, and negate it to say the opposite, but we assume that when they negate, they mean something decisively empty. This is a strong assumption but generally safe as people tend not to add expressions that seem like common sense. Wherever we see the predicate negated in the definition, we switch it out with the other predicate. This applies to \texttt{filled}/\texttt{empty}, \texttt{open}/\texttt{closed}, and \texttt{folded}/\texttt{unfolded}.

To ensure that the definition is logically the same after the switch, the underlying~\bddl implementation converts the definition using De Morgan’s Law such that all negations are only applied to atomic formulae before the switch occurs.


\paragraph{Composition and decomposition of objects:} In standard PDDL/~\bddl, the objects in \texttt{:objects} are assumed to persist throughout the activity. There is no concept of creating a new object that did not appear in the \texttt{:init} or explicitly destroying an object. 

To enable our transition rules that involve turning some objects into others, we require a representation that will provide the desired information unambiguously. The \texttt{:objects} section cannot be inferred simply from a \texttt{:goal} that has new objects in it, because \texttt{:goal} is not exhaustive. We therefore introduce the \texttt{future} predicate, used in \texttt{:init} on all objects that do not appear in the scene when the agent enters, but must be present for the \texttt{:goal} to be satisfied. All objects in \texttt{future} predicates appear in \texttt{:objects} and they cannot appear in any other literals in \texttt{:init}. This approach takes inspiration from~\cite{wichlacz2019ghost}, which updates a reference to an object (e.g. a wall) as it keeps changing form as more sub-objects (e.g. blocks) are added to it.

\section{Scene and Object Models}
\label{as:modeling}
In this section, we provide more details about the object and scene models presented in \dataset, including the selection, modeling, and annotation processes.

The survey outlined in Sec.~\ref{sec:survey} provided us with a list of 1000 activities that humans prefer robots to perform. However, these activities are not restricted to a unique type of scene (e.g., houses). We chose scene types necessary to cover the \benchmark activities, including eight scene types: houses (15), houses with gardens (8), hotels (3), offices (5), grocery stores (4), generic halls (4), restaurants (6), and schools (5) (see Table~\ref{tab:model_scene}). These scenes cover activities that require a specific type (e.g., shopping, cooking, restocking) and activities that are generic and could happen in multiple scene types (e.g., cleaning). We also annotated how many activities could be performed on each type to guide us on how many instances of each scene type should we include in \dataset. The main type of scene is still households: we improved 15 household scenes from \oldbenchmark and annotated it further with light sources, new textures, etc. We then acquired additional instances of each scene type from online marketplaces such as TurboSquid~\cite{turbosquid}. 

Several of the available scene models in the marketplaces did not contain all the necessary rooms to perform the natural activities in \benchmark, e.g., restaurants did not include the kitchen, offices did not include the restrooms, or houses did not include the gardens.
We collected separate models for those and contracted professional 3D designers to connect them.

\begin{figure}[th!]
 \centering
 \includegraphics[width=0.9\textwidth]{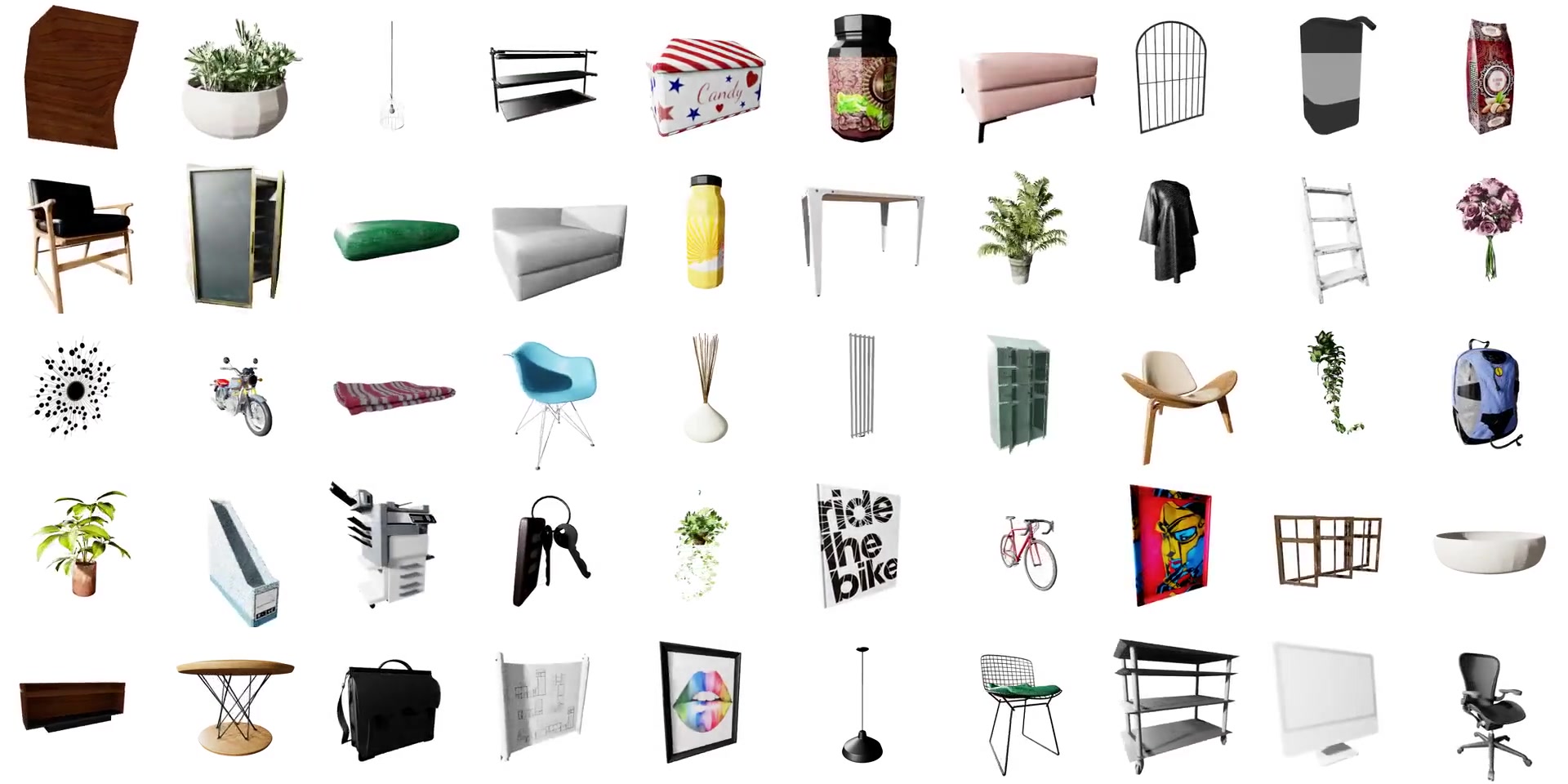}
 \caption{\revision{A collage of objects included in the \dataset. These objects highlight key features of \simulator such as transparency, articulation, heat simulation, and lighting.}}
 \label{fig:model_object}
\end{figure}

Scene models were acquired with the necessary object models. However, they were not enough to cover the objects required by the activities in \benchmark. 
We acquired additional models to support the activities, as provided by the activity annotation process described in Sec.~\ref{as:act_ann}, totaling 9,000+ object models from 1,900+ categories. The diversity of the object categories included in the dataset can be observed in Fig.~\ref{fig:model_object}.

As provided by the 3D vendors, the scene and object models cannot be used directly for the simulation of the 1000 activities in \benchmark in \simulator due to 1) lack of category annotation, 2) lack (or incorrect) of light sources in scenes, 3) incorrect part segmentation, 4) lack of articulation, 5) missing interactive elements like buttons, 6) lack of a unified canonical frame orientation for sampling, and 7) poor physical properties for realistic simulation. We manually cleaned and annotated all scenes and objects to correct these elements.

We will publicly release all the scenes and models to be used by other researchers. The models will be encrypted and only be used within \simulator in order to comply with the rights of the model authors and the vendors' agreement. The documentation of the annotation process as well as source code for the pipeline will similarly be released on our website, allowing users to easily import their own objects and scenes into \simulator for use in \benchmark activities.

\begin{figure}[th!]
    \centering
    \includegraphics[width=\textwidth]{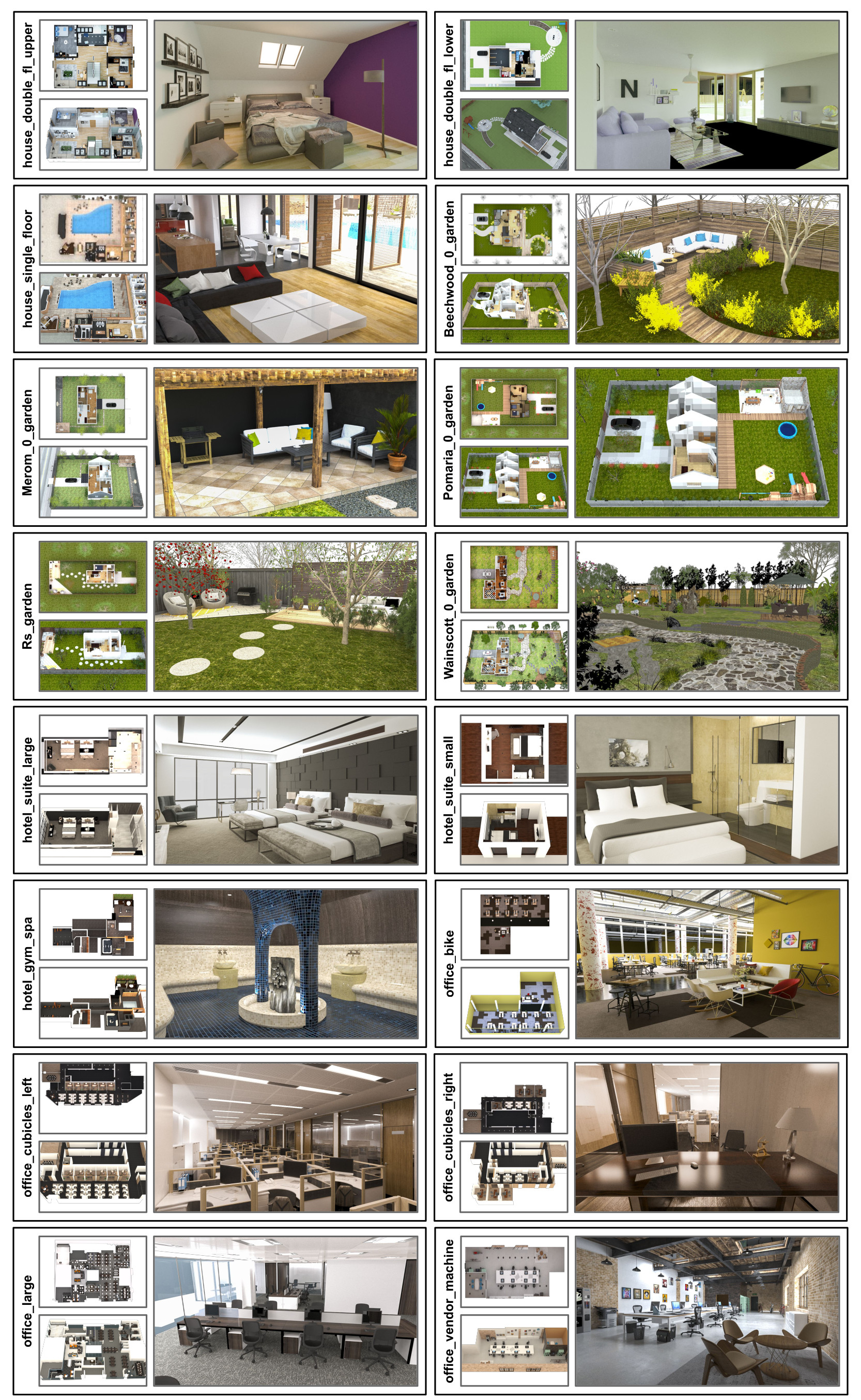}
    \caption{\revision{Part 1 of a scene collage showing views of all 50 scenes.}}
\end{figure}

\begin{figure}[th!]
    \centering
    \includegraphics[width=\textwidth]{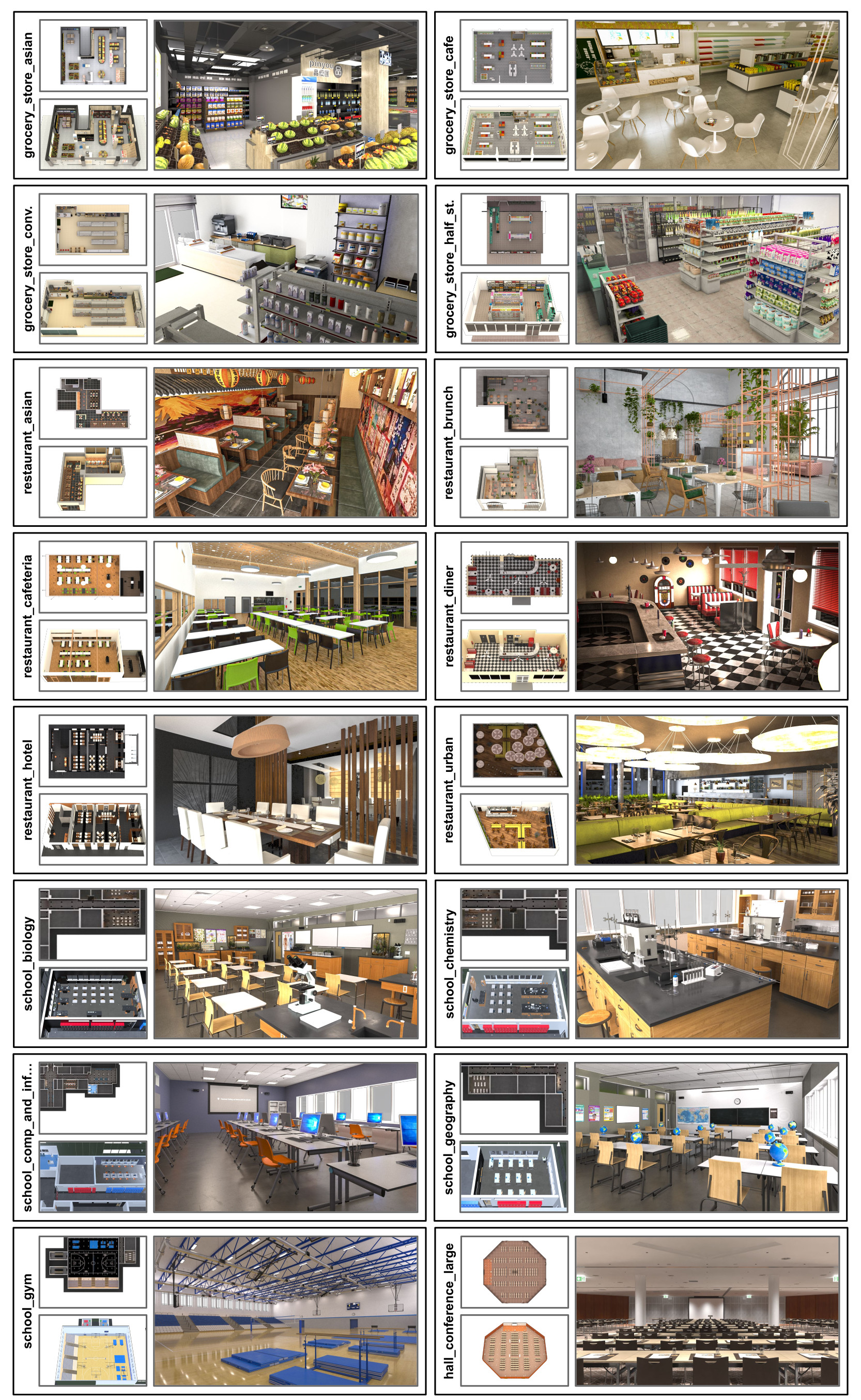}
    \caption{\revision{Part 2 of a scene collage showing views of all 50 scenes.}}
\end{figure}

\begin{figure}[th!]
    \centering
    \includegraphics[width=\textwidth]{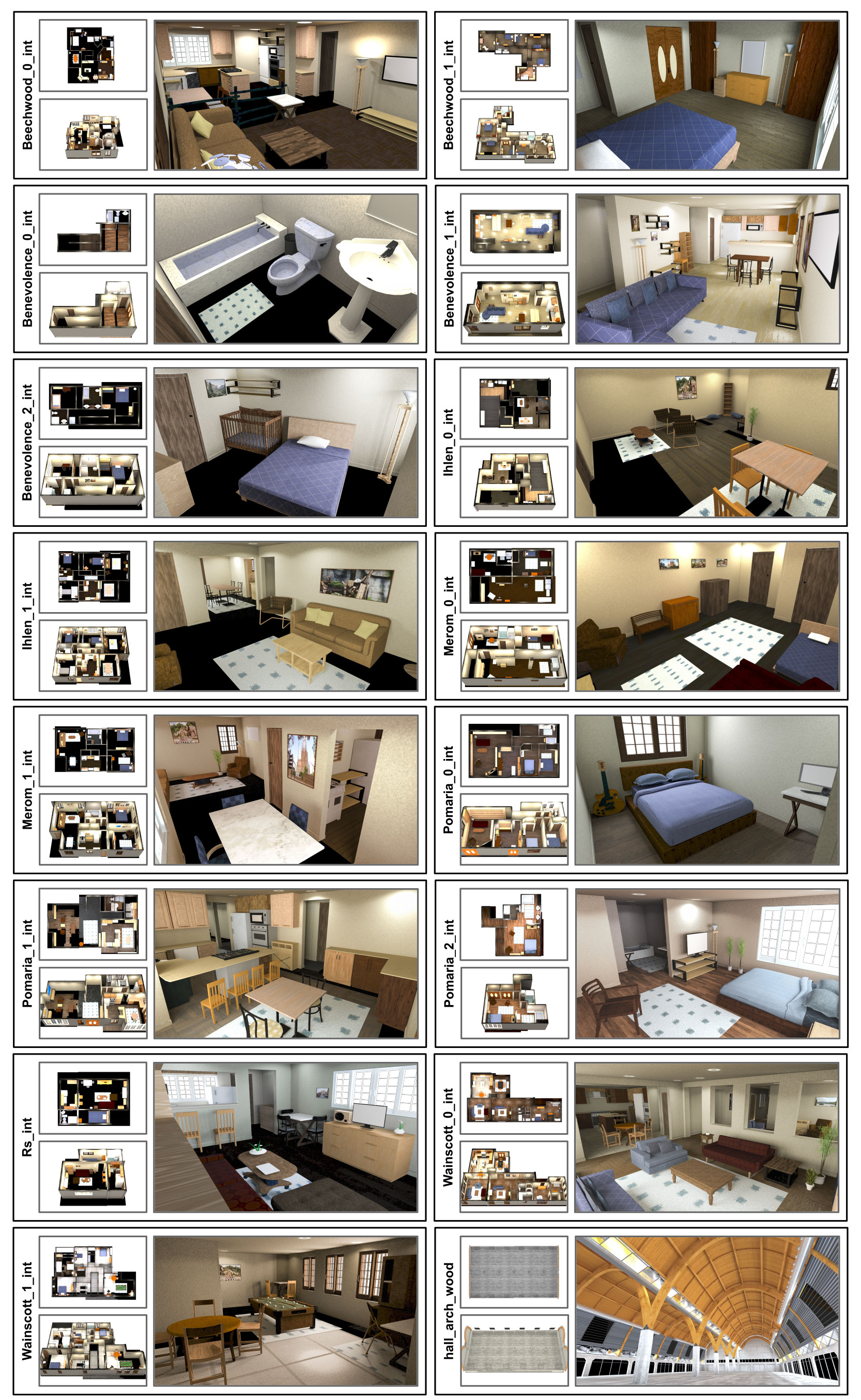}
    \caption{\revision{Part 3 of a scene collage showing views of all 50 scenes.}}
\end{figure}

\begin{figure}[th!]
    \centering
    \includegraphics[width=\textwidth]{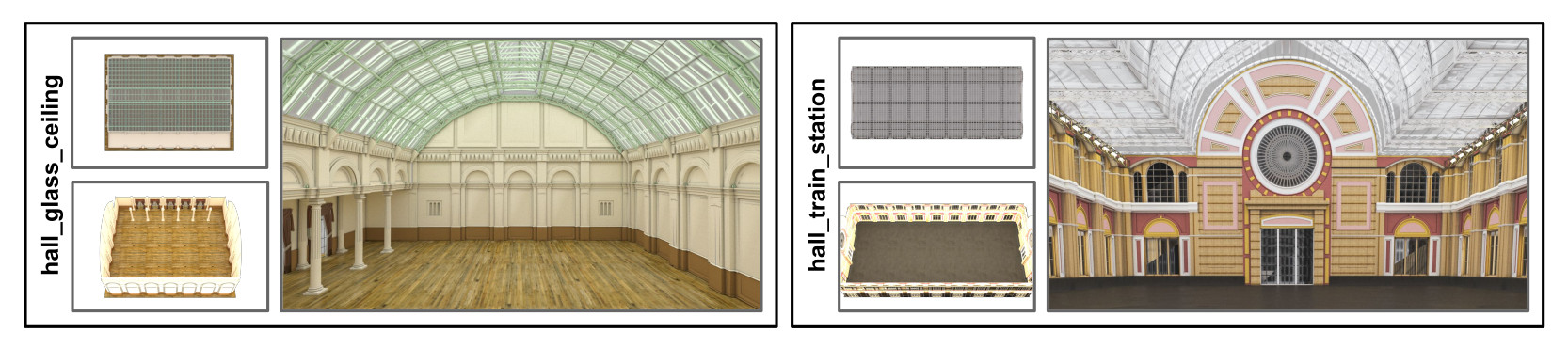}
    \caption{\revision{Part 4 of a scene collage showing views of all 50 scenes.}}
\end{figure}

\vspace{1.5cm}

{\scriptsize
\begin{longtable}{|m{.03\textwidth}|m{.16\textwidth}|m{.02\textwidth}|m{.02\textwidth}|m{.02\textwidth}|m{.27\textwidth}|m{.27\textwidth}|}
\hline
Scene Type
& Scene Name                            & Obj. Cnt.    & Syn. Cnt.    & Rm. Cnt.   & Room Types                                                                                                                                       & Example Activities                                                                                                                                                                                                                 \\ \hline
\multirow{15}{*}{\rotatebox[origin=c]{90}{\textbf{BEHAVIOR-100 Houses}}}
& Beechwood\_0\_int                     & 136          & 32           & 8          & \texttt{bathroom}, \texttt{corridor}, \texttt{dining\_room}, \texttt{entryway}, \texttt{kitchen}, \texttt{living\_room}, \texttt{private\_office}, \texttt{utility\_room}                                                & \texttt{clean a hot water dispenser}, \texttt{freeze lasagna}, \texttt{clean batting gloves}      \\ \cline{2-7} 
& Beechwood\_1\_int                     & 129          & 21           & 9          & \texttt{bathroom}, \texttt{bedroom}, \texttt{childs\_room}, \texttt{closet}, \texttt{corridor}, \texttt{playroom}, \texttt{television\_room }                                                                   & \texttt{clean your kitty litter box}, \texttt{store baby clothes}, \texttt{cleaning pet bed}      \\ \cline{2-7} 
& Benevolence\_0\_int                   & 21           & 10           & 4          & \texttt{bathroom}, \texttt{corridor}, \texttt{empty\_room}, \texttt{entryway}                                                                                                        & \texttt{laying clothes out}, \texttt{pack a pencil case}, \texttt{sweeping outside entrance}      \\ \cline{2-7} 
& Benevolence\_1\_int                   & 74           & 22           & 5          & \texttt{corridor}, \texttt{dining\_room}, \texttt{kitchen}, \texttt{living\_room}, \texttt{storage\_room}                                                                                     & \texttt{hanging blinds}, \texttt{sorting volunteer materials}, \texttt{wash baby bottles}      \\ \cline{2-7} 
& Benevolence\_2\_int                   & 63           & 23           & 5          & \texttt{bathroom}, \texttt{bedroom}, \texttt{corridor}                                                                                                                      & \texttt{clean walls}, \texttt{washing fabrics}, \texttt{folding clean laundry}      \\ \cline{2-7} 
& Ihlen\_0\_int                         & 68           & 21           & 7          & \texttt{bathroom}, \texttt{corridor}, \texttt{dining\_room}, \texttt{garage}, \texttt{living\_room}, \texttt{storage\_room}                                                                            & \texttt{de-clutter your garage}, \texttt{sorting household items}, \texttt{set up a home office in your garage}      \\ \cline{2-7} 
& Ihlen\_1\_int                         & 147          & 27           & 9          & \texttt{bathroom}, \texttt{bedroom}, \texttt{corridor}, \texttt{dining\_room}, \texttt{kitchen}, \texttt{living\_room}, \texttt{staircase}                                                                      & \texttt{clean jewels}, \texttt{clean green beans}, \texttt{hanging up curtains}      \\ \cline{2-7} 
& Merom\_0\_int                         & 82           & 27           & 6          & \texttt{bathroom}, \texttt{childs\_room}, \texttt{living\_room}, \texttt{playroom}, \texttt{storage\_room}, \texttt{utility\_room}                                                                     & \texttt{changing dog's water}, \texttt{wash a bra}, \texttt{wash a wool coat}      \\ \cline{2-7} 
& Merom\_1\_int                         & 123          & 28           & 8          & \texttt{bathroom}, \texttt{bedroom}, \texttt{childs\_room}, \texttt{corridor}, \texttt{dining\_room}, \texttt{kitchen}, \texttt{living\_room}, \texttt{staircase}                                                        & \texttt{store an uncooked turkey}, \texttt{drying table}, \texttt{clean flip flops}      \\ \cline{2-7} 
& Pomaria\_0\_int                       & 71           & 22           & 6          & \texttt{bathroom}, \texttt{bedroom}, \texttt{corridor}, \texttt{private\_office}, \texttt{television\_room }                                                                                  & \texttt{prepare sea salt soak}, \texttt{clean sheets}, \texttt{dispose of medication}      \\ \cline{2-7} 
& Pomaria\_1\_int                       & 89           & 24           & 6          & \texttt{bathroom}, \texttt{corridor}, \texttt{kitchen}, \texttt{living\_room}, \texttt{pantry\_room}, \texttt{utility\_room}                                                                           & \texttt{store brownies}, \texttt{clean a book}, \texttt{baking sugar cookies}      \\ \cline{2-7} 
& Pomaria\_2\_int                       & 42           & 18           & 3          & \texttt{bathroom}, \texttt{bedroom}, \texttt{corridor}                                                                                                                      & \texttt{replacing screens}, \texttt{clean baby toys}, \texttt{putting up posters}      \\ \cline{2-7} 
& Rs\_int                               & 72           & 32           & 5          & \texttt{bathroom}, \texttt{bedroom}, \texttt{entryway}, \texttt{kitchen}, \texttt{living\_room}                                                                                               & \texttt{dispose of a pizza box}, \texttt{putting food in fridge}, \texttt{putting out condiments}      \\ \cline{2-7} 
& Wainscott\_0\_int                     & 187          & 35           & 9          & \texttt{bathroom}, \texttt{bedroom}, \texttt{dining\_room}, \texttt{kitchen}, \texttt{living\_room}                                                                                           & \texttt{make microwave popcorn}, \texttt{make frozen lemonade}, \texttt{make cake mix}      \\ \cline{2-7} 
& Wainscott\_1\_int                     & 150          & 26           & 10         & \texttt{bathroom}, \texttt{bedroom}, \texttt{corridor}, exercise\_room, \texttt{playroom}, \texttt{private\_office}, \texttt{utility\_room}                                                            & \texttt{decorating for religious ceremony}, \texttt{stash snacks in your room}, \texttt{disinfect laundry}      \\ \hline
\multirow{5}{*}{\rotatebox[origin=c]{90}{{\textbf{BEHAVIOR-100 Houses+Garden}}}}
& Beechwood\_0\_garden                  & 335          & 45           & 9          & \texttt{bathroom}, \texttt{corridor}, \texttt{dining\_room}, \texttt{entryway}, \texttt{garden}, \texttt{kitchen}, \texttt{living\_room}, \texttt{private\_office}, \texttt{utility\_room}                                        & \texttt{tidy your garden}, \texttt{opening doors}, \texttt{wash goalkeeper gloves}      \\ \cline{2-7} 
& Merom\_0\_garden                      & 197          & 51           & 8          & \texttt{bathroom}, \texttt{childs\_room}, \texttt{garden}, \texttt{living\_room}, \texttt{playroom}, \texttt{sauna}, \texttt{storage\_room}, \texttt{utility\_room}                                                      & \texttt{clean your kitty litter box}, \texttt{clean a glass windshield}, \texttt{hang icicle lights}      \\ \cline{2-7} 
& Pomaria\_0\_garden                    & 260          & 49           & 7          & \texttt{bathroom}, \texttt{bedroom}, \texttt{corridor}, \texttt{garden}, \texttt{private\_office}, \texttt{television\_room }                                                                          & \texttt{putting up Christmas lights outside}, \texttt{prepare a hanging basket}, \texttt{clean a patio}      \\ \cline{2-7} 
& Rs\_garden                            & 185          & 47           & 6          & \texttt{bathroom}, \texttt{bedroom}, \texttt{entryway}, \texttt{garden}, \texttt{kitchen}, \texttt{living\_room}                                                                                       & \texttt{cleaning driveway}, \texttt{clean a longboard}, \texttt{roast nuts}      \\ \cline{2-7} 
& Wainscott\_0\_garden                  & 712          & 61           & 10         & \texttt{bathroom}, \texttt{bedroom}, \texttt{dining\_room}, \texttt{garden}, \texttt{kitchen}, \texttt{living\_room}                                                                                   & \texttt{clean an espresso machine}, \texttt{clearing food from table into fridge}, \texttt{painting porch}      \\ \hline
\multirow{3}{*}{\rotatebox[origin=c]{90}{\textbf{New Houses}}}
& house\_single\_floor                  & 1375         & 137          & 23         & \texttt{bathroom}, \texttt{bedroom}, \texttt{childs\_room}, \texttt{closet}, \texttt{corridor}, \texttt{dining\_room}, \texttt{empty\_room}, \texttt{entryway}, \texttt{garden}, \texttt{kitchen}, \texttt{living\_room}, \texttt{sauna}, \texttt{utility\_room}      & \texttt{turning out all lights before sleep}, \texttt{iron curtains}, \texttt{packing hobby equipment}      \\ \cline{2-7} 
& {\tiny house\_double\_floor\_lower}   & 304          & 79           & 6          & \texttt{bathroom}, \texttt{corridor}, \texttt{garage}, \texttt{garden}, \texttt{kitchen}, \texttt{living\_room}                                                                                        & \texttt{wash towels}, \texttt{clean a fish}, \texttt{clean brass}      \\ \cline{2-7} 
& {\tiny house\_double\_floor\_upper}   & 325          & 78           & 5          & \texttt{bathroom}, \texttt{bedroom}, \texttt{television\_room }                                                                                                             & \texttt{clean a saxophone}, \texttt{taking down curtains}, \texttt{clean walls}      \\ \hline
\multirow{4}{*}{\rotatebox[origin=c]{90}{\textbf{Grocery Stores}}}
& grocery\_store\_asian                 & 3402         & 79           & 2          & \texttt{bathroom}, \texttt{grocery\_store}                                                                                                                         & \texttt{buy alcohol}, \texttt{buy boxes for packing}, \texttt{buy a microwave oven}      \\ \cline{2-7} 
& grocery\_store\_cafe                  & 6994         & 71           & 4          & \texttt{bar} , \texttt{bathroom}, \texttt{dining\_room}, \texttt{grocery\_store}                                                                                                      & \texttt{defrost meat}, \texttt{clean reusable shopping bags}, \texttt{buy a good avocado}      \\ \cline{2-7} 
& {\tiny grocery\_store\_convenience}   & 1889         & 82           & 2          & \texttt{bathroom}, \texttt{grocery\_store}                                                                                                                         & \texttt{washing windows}, \texttt{picking up prescriptions}, \texttt{buy food for vegetarians}      \\ \cline{2-7} 
& {\tiny grocery\_store\_half\_stocked} & 3804         & 51           & 2          & \texttt{bathroom}, \texttt{grocery\_store}                                                                                                                         & \texttt{buying gardening supplies}, \texttt{buy basic garden tools}, \texttt{buy boxes for packing}      \\ \hline
\multirow{4}{*}{\rotatebox[origin=c]{90}{\textbf{Halls}}}
& hall\_arch\_wood                      & 78           & 11           & 2          & \texttt{bathroom}, \texttt{empty\_room}                                                                                                                         &    \texttt{clean walls}  \\ \cline{2-7} 
& hall\_train\_station                  & 141          & 15           & 2          & \texttt{bathroom}, \texttt{empty\_room}                                                                                                                            &  \texttt{clean cement}    \\ \cline{2-7} 
& hall\_glass\_ceiling                  & 116          & 20           & 2          & \texttt{bathroom}, \texttt{empty\_room}                                                                                                                            &  \texttt{preparing food for a fundraiser}   \\ \cline{2-7} 
& hall\_conference\_large               & 3372         & 26           & 2          & \texttt{bathroom}, \texttt{conference\_hall}                                                                                                                       &  \texttt{distributing event T-shirts}    \\ \hline
\multirow{3}{*}{\rotatebox[origin=c]{90}{\textbf{Hotels}}}
& hotel\_gym\_spa                       & 322          & 43           & 9          & \texttt{bathroom}, \texttt{corridor}, \texttt{gym} , \texttt{hammam} , \texttt{locker\_room} , \texttt{sauna}, \texttt{spa}                                                                                         & \texttt{turning on the hot tub}, \texttt{adding chemicals to hot tub}, \texttt{clean a sauna}      \\ \cline{2-7}
& hotel\_suite\_large                   & 218          & 51           & 2          & \texttt{bathroom}, \texttt{bedroom}                                                                                                                                & \texttt{clean vinyl shutters}, \texttt{cleaning computer}, \texttt{clean white marble}      \\ \cline{2-7} 
& hotel\_suite\_small                   & 48           & 28           & 2          & \texttt{bathroom}, \texttt{bedroom}                                                                                                                                & \texttt{clean cork mats}, \texttt{putting out clean towels}, \texttt{clean a shower}      \\ \hline
\multirow{5}{*}{\rotatebox[origin=c]{90}{\textbf{Offices}}}
& office\_bike                          & 479          & 49           & 3          & \texttt{bathroom}, \texttt{break\_room}, \texttt{shared\_office}                                                                                                            & \texttt{set up a webcam}, \texttt{set up two computer monitors}, \texttt{clean an office chair}      \\ \cline{2-7} 
& office\_cubicles\_left                & 787          & 44           & 12         & \texttt{bathroom}, \texttt{conference\_hall}, \texttt{copy\_room}, \texttt{corridor}, \texttt{lobby}, \texttt{meeting\_room}, \texttt{private\_office}, \texttt{shared\_office}                                          & \texttt{clean up your desk}, \texttt{clean a LED screen}, \texttt{laying out snacks at work}      \\ \cline{2-7} 
& office\_cubicles\_right               & 406          & 37           & 11         & \texttt{bathroom}, \texttt{conference\_hall}, \texttt{copy\_room}, \texttt{corridor}, \texttt{lobby}, \texttt{meeting\_room}, \texttt{private\_office}, \texttt{shared\_office}                                          & \texttt{making photocopies}, \texttt{clean a LED screen}, \texttt{clean a keyboard}      \\ \cline{2-7} 
& office\_large                         & 1151         & 46           & 24         & \texttt{bathroom}, \texttt{break\_room}, \texttt{conference\_hall}, \texttt{copy\_room}, \texttt{corridor}, \texttt{lobby}, \texttt{meeting\_room}, \texttt{phone\_room}, \texttt{private\_office}, \texttt{shared\_office}                & \texttt{emptying trash cans}, \texttt{putting meal in fridge at work}, \texttt{brewing coffee}      \\ \cline{2-7} 
& office\_vendor\_machine               & 225          & 45           & 4          & \texttt{bathroom}, \texttt{break\_room}, \texttt{meeting\_room}, \texttt{shared\_office}                                                                                             & \texttt{clean a computer monitor}, \texttt{make the workplace exciting}, \texttt{dispose of glass}      \\ \hline
\multirow{6}{*}{\rotatebox[origin=c]{90}{\textbf{Restaurants}}}
& restaurant\_asian                     & 1221         & 56           & 3          & \texttt{bathroom}, \texttt{dining\_room}, \texttt{kitchen}                                                                                                                  & \texttt{grill vegetables}, \texttt{cook tofu}, \texttt{clean clams}      \\ \cline{2-7} 
& restaurant\_brunch                    & 1096         & 70           & 4          & \texttt{bar} , \texttt{bathroom}, \texttt{dining\_room}, \texttt{kitchen}                                                                                                             & \texttt{stock a bar}, \texttt{cook squid}, \texttt{washing vegetables}      \\ \cline{2-7} 
& restaurant\_cafeteria                 & 292          & 45           & 3          & \texttt{bathroom}, \texttt{dining\_room}, \texttt{kitchen}                                                                                                                  & \texttt{clean a popcorn machine}, \texttt{store coffee beans or ground coffee}, \texttt{roast meat}      \\ \cline{2-7} 
& restaurant\_diner                     & 174          & 39           & 3          & \texttt{bar} , \texttt{bathroom}, \texttt{dining\_room}                                                                                                                      & \texttt{sweeping floors}, \texttt{installing smoke detectors}, \texttt{clean a lobster}      \\ \cline{2-7} 
& restaurant\_hotel                     & 1080         & 81           & 4          & \texttt{bathroom}, \texttt{dining\_room}, \texttt{kitchen}, \texttt{lobby}                                                                                                           & \texttt{clean eggs}, \texttt{setting table for coffee}, \texttt{clean a pizza stone}      \\ \cline{2-7}
& restaurant\_urban                     & 1368         & 90           & 4          & \texttt{bar} , \texttt{bathroom}, \texttt{dining\_room}, \texttt{kitchen}                                                                                                             & \texttt{cook pumpkin seeds}, \texttt{putting roast in oven}, \texttt{thaw frozen fish}      \\ \hline
\multirow{5}{*}{\rotatebox[origin=c]{90}{\textbf{Schools}}}
& school\_biology                       & 717          & 53           & 4          & \texttt{bathroom}, \texttt{biology\_lab} , \texttt{corridor}                                                                                                                 & \texttt{clean an eraser}, \texttt{turning sprinkler off}, \texttt{clean a glass pipe}      \\ \cline{2-7} 
& school\_chemistry                     & 890          & 69           & 4          & \texttt{bathroom}, \texttt{chemistry\_lab} , \texttt{corridor}                                                                                                               & \texttt{clean clear plastic}, \texttt{clean an eraser}, \texttt{clean a whiteboard}      \\ \cline{2-7} 
& {\tiny school\_comp\_lab\_infirmary}  & 828          & 61           & 6          & \texttt{bathroom}, \texttt{computer\_lab}, \texttt{corridor}, \texttt{infirmary}                                                                                                     & \texttt{clean a computer monitor}, \texttt{prepare an emergency school kit}, \texttt{clean a keyboard}      \\ \cline{2-7} 
& school\_geography                     & 556          & 42           & 5          & \texttt{bathroom}, \texttt{classroom}, \texttt{corridor}                                                                                                                    & \texttt{clean gold}, \texttt{clean a glass pipe}, \texttt{clean an eraser}      \\ \cline{2-7} 
& school\_gym                           & 853          & 36           & 7          & \texttt{bathroom}, \texttt{corridor}, \texttt{gym} , \texttt{locker\_room}                                                                                                             & \texttt{clean a dirty baseball}, \texttt{dispose of glass}, \texttt{wash towels}      \\ \hline

\caption{Statistics and information for each of 50 scenes in B1K: scene type, name, number of objects, unique synsets and rooms, room types, and example activities }
\label{tab:model_scene}
\end{longtable}
}

\section{Simulation}

\subsection{Extended Object States and Logical Predicates in \simulator}
\label{as:object_states}
\simulator extends the infrastructure of extended object states and logical predicates from iGibson 2.0~\cite{li2021ig2} to accommodate the diversity and realism of \benchmark.

\subsubsection{Extended object states associated with object category properties}
For computational efficiency purposes, not all extended object states need to be maintained and updated for every object. For example, \texttt{cookable} objects like apples need to keep track of their \texttt{Temperature}, but probably tables don't need to, at least for the purpose of simulating \benchmark activities. Hence, given the object category property annotation from \dataset (see Table~\ref{tab:supp_props}), \simulator selectively keeps track of a subset of the extended states for objects of each category. The mapping from the object category property to the required object states can be found in Table~\ref{table:obj_cat_prop}.
\begin{table}[!htbp]
\begin{center}
{%
\scriptsize
\renewcommand{\arraystretch}{1.2}
\begin{tabular}{ | m{3.5cm} | m{3.7cm} | } 
\hline
\textbf{Object category property} & \textbf{Required extended object states}\\ 
\hline
\texttt{cookable} & \texttt{MaxTemperature}, \texttt{Temperature}\\ 
\hline
\texttt{overcookable} & \texttt{MaxTemperature}, \texttt{Temperature}\\ 
\hline
\texttt{freezable} & \texttt{Temperature}\\ 
\hline
\texttt{flammable} & \texttt{Temperature}\\ 
\hline
\texttt{heatable} & \texttt{Temperature}\\ 
\hline
\texttt{metable} & \texttt{Temperature}\\ 
\hline
\texttt{soakable} & \texttt{SoakedLevel}\\ 
\hline
\texttt{toggleable} & \texttt{ToggledState}\\ 
\hline
\texttt{sliceable} & \texttt{SlicedState}\\ 
\hline
\texttt{breakable} & \texttt{BrokenState}\\ 
\hline
\texttt{heatSource} & \texttt{ToggledState}\\ 
\hline
\texttt{fireSource} & \texttt{ToggledState}\\ 
\hline
\texttt{coldSource} & \texttt{ToggledState}\\ 
\hline
\texttt{waterSource} & \texttt{ToggledState}\\ 
\hline
\end{tabular}
}
\vspace{3pt}
\caption{Extended object states associated with object category properties. Adapted from~\cite{li2021ig2}.}
\label{table:obj_cat_prop}
\end{center}
\end{table}

\subsubsection{Object model properties}
We also need to perform physical and semantic annotation for each object model so that they can be realistically simulated and support the evolution of extended object states. Some of the physical properties can be programmatically generated from the 3D assets, such as \texttt{Shape}, \texttt{KinematicStructure} and  \texttt{StableOrientations}, while others need to be annotated (included in \dataset), such as \texttt{Weight}. \texttt{Type} of an object model, which is also derived from the object category property annotation from \dataset (see Table~\ref{tab:supp_props}), determines how the object will be simulated in \simulator. For example, cloths and fluids will be simulated using underlying particle systems. Also, some of the object category property requires additional semantic annotation for each model of that category, e.g. for \texttt{waterSource} objects, \texttt{WaterSourceLink} is required to indicate the exact location to generate water. An exhaustive list of all the object model properties can be found in Table~\ref{table:obj_model_prob}.
\begin{table}[!htbp]
\begin{center}
{%
\scriptsize
\renewcommand{\arraystretch}{1.2}
\begin{tabular}{ | m{2.5cm} | m{2.7cm} | m{7.5cm} | } 
\hline
\textbf{Object model property} & \textbf{Relevant object category property} & \textbf{Description} \\ 
\hline
\texttt{Shape} &  & Model of the 3D shape of each link of the object\\ 
\hline
\texttt{Weight} &  & Weight of the object \\ 
\hline
\texttt{CenterOfMass} &  & Mean position of the matter in the object\\ 
\hline
\texttt{MomentOfInertia} &  & Resistance of the object to change its angular velocity\\ 
\hline
\texttt{KinematicStructure} &  & Structure of links and joints connecting them in the form of URDF (non-articulated objects are composed of one link)\\ 
\hline
\texttt{StableOrientations} &  & A list of stable orientations assuming the object is placed on a flat surface, computed using a 3D geometry library\\ 
\hline
\texttt{HeatSourceLink} & \texttt{heatSource} & Virtual (non-colliding) fixed link that generates heat\\
\hline
\texttt{FireSourceLink} & \texttt{fireSource} & Virtual (non-colliding) fixed link that generates fire\\
\hline
\texttt{CleaningToolLink} & \texttt{cleaningTool} & Fixed link that needs to contact dirt particles for the tool to clean them\\ 
\hline
\texttt{WaterSourceLink} &  \texttt{waterSource} & Virtual (non-colliding) fixed link that generates water\\
\hline
\texttt{WaterSinkLink} &  \texttt{waterSource} & Virtual (non-colliding) fixed link that absorbs water\\ 
\hline
\texttt{TogglingLink} &  \texttt{toggleable} & Virtual (non-colliding) fixed link that changes the toggled state of the object when contacted\\ 
\hline
\texttt{SlicingLink} & \texttt{slicingTool} & Fixed link that changes the sliced state of another object if it contacts it with enough force\\ 
\hline
\texttt{RelevantJoints} & \texttt{openable} & List of joints that are relevant to indicate whether an object is open\\ 
\hline
\texttt{AttachmentKeypoints} & \texttt{assembleable} & List of keypoint locations that will be connected with those of other objects when they are close enough\\ 
\hline
\texttt{ClothKeypoints} & \texttt{foldable}, \texttt{unfoldable} & List of keypoint locations are used to determine if the object is folded (if they are close enough) or unfolded (if they are far enough)\\ 
\hline
\texttt{ContainerVolume} & \texttt{fillable} & Virtual (non-colliding) fixed link that represents the inner volume of a fillable container.\\ 
\hline
\texttt{Type} & \texttt{cloth}, \texttt{deformable}, \texttt{liquid}, \texttt{rope}, \texttt{physicalSubstance}, \texttt{visualSubstance}  & Type of the object, e.g. rigid body, fluid, physical/visual substance, cloth, deformable, which determines how it will be simulated in \simulator \\ 
\hline
\end{tabular}
}
\vspace{3pt}
\caption{Permanent object model properties. Adapted from~\cite{li2021ig2}.}
\label{table:obj_model_prob}
\end{center}
\end{table}

\subsubsection{Object states}
While the object properties mentioned above stay the same during simulation (i.e. an apple is always \texttt{cookable} and a sink always have the same \texttt{WaterSourceLink} defined in its local frame), the object states could change. The underlying physics engine handles the kinematic state changes, such as \texttt{Pose}, \texttt{AABB}, \texttt{JointStates}, \texttt{ParticlePositions} for fluids and cloths, etc. On top of these, \simulator handles the non-kinematic state changes that are essential for logical predicates (described in the next subsection). For example, \simulator update the \texttt{Temperature} of objects at every simulation step by checking if they are near/inside any \texttt{heatSource} or \texttt{coldSource}. An exhaustive list of all the object states can be found in Table~\ref{table:obj_states}.
\begin{table}[!htbp]
\begin{center}
{\scriptsize%
\renewcommand{\arraystretch}{1.2}
\begin{tabular}{ | m{4cm} | m{9cm} | } 
\hline
\textbf{Object State} & \textbf{Description and Update Rules} \\ 
\hline
\texttt{Pose} & 6 DoF pose (position and orientation) of the object in world reference frame, updated by the underlying physics engine.\\ 
\hline
\texttt{AABB} & Axis-aligned bounding box (coordinates of two opposite corners) of the object in the world reference frame, updated by the underlying physics engine.\\ 
\hline
\texttt{JointStates} & State of all internal DoFs of the (articulated) object for the structure defined by \texttt{KinematicStructure}, updated by the underlying physics engine.\\ 
\hline
\texttt{ParticlePositions} & Positions of all the underlying particles for cloth and fluid, updated by the underlying physics engine.\\ 
\hline
\texttt{InContactObjs} & List of all objects in physical contact with the object, updated by the underlying physics engine.\\ 
\hline
\texttt{ConnectedObjs} & List of all objects that are connected to the object, either via a fixed joint for rigid bodies or via an attachment for cloth and deformables.\\ 
\hline
\texttt{InSamePositiveVerticalAxisObjs} & List of all objects in the positive vertical axis drawn from the object's center of mass, updated by shooting a ray upwards in the positive z-axis and gather the objects hit by the ray.\\ 
\hline
\texttt{InSameNegativeVerticalAxisObjs} & List of all objects in the negative vertical axis drawn from the object's center of mass, updated by shooting a ray downwards in the negative z-axis and gather the objects hit by the ray.\\ 
\hline
\texttt{InSameHorizontalPlaneObjs} & List of all objects in the horizontal plane drawn from the object's center of mass, updated by shooting a number of ray in the x-y plane and gather the objects hit by the rays.\\ 
\hline
\texttt{Temperature}, $T$ & Object's current temperature in $^\circ$C, updated by detecting if the object is affected by any heat source or heat sink.\\ 
\hline
\texttt{MaxTemperature}, $T_\textit{max}$ & Maximum temperature of the object reached historically during this simulation run, updated by keeping track of all the \texttt{Temperature} in the history.\\ 
\hline
\texttt{SoakedLevel}, $w$ & Amount of liquid absorbed by the object corresponding to the number of liquid particles contacted, updated by detecting if the object is in contact with any liquid particle. This is maintained for every type of \texttt{liquid} separately. \\ 
\hline
\texttt{CoveredLevel}, $c$ & Amount of \texttt{visualSubstance} that covers the object, updated by detecting if the particles of the \texttt{visualSubstance} are in contact with anything that can potentially remove them from the object, e.g. \texttt{cleaningTool}. This is maintained for every type of \texttt{visualSubstance} separately. \\ 
\hline
\texttt{ToggledState}, $\textit{TS}$ & Binary state indicating if the object is currently on or off, updated by detecting if the agent is in contact with the \texttt{TogglingLink}.\\ 
\hline
\texttt{SlicedState}, $\textit{SS}$ & Binary state indicating whether the object has been sliced (irreversible), updated by detecting if the object is in contact with any \texttt{SlicingTool} that exerts a force above a certain threshold $F_\mathit{sliced}$. We assume as default force threshold of $F_\mathit{sliced} = \SI{10}{\newton}$, a value that can be configured per object category and model.\\ 
\hline
\texttt{BrokenState}, $\textit{BS}$ & Binary state indicating if the object is broken, updated by detecting if the object has a contact force with any other object above a certain threshold $F_\mathit{broken}$, a value that can be configured per object category and model.\\ 
\hline
\end{tabular}
}
\vspace{3pt}
\caption{Object states maintained by \simulator. Adapted from~\cite{li2021ig2}.}
\label{table:obj_states}
\end{center}
\end{table}

\subsubsection{Logical predicates as checking functions}
For each logical predicate that is relevant for \benchmark activities, we define a checking function that maps a given physical state (kinematic and non-kinematic) into a binary logical state that \bddl operates on. For example, \texttt{OnTopOf} is based on the pose of two objects and their contact information whereas \texttt{Cooked} is based on the \texttt{Temperature}. The details of the checking functions for all the logical predicates can be found in Table~\ref{table:obj_pred_checking}.

\begin{table}[!htbp]
\begin{center}
{\scriptsize%
\renewcommand{\arraystretch}{1.2}
\begin{tabular}{ | m{0.20\textwidth} | m{0.7\textwidth} | } 
\hline
\textbf{Predicate} & \textbf{Description} \\ 
\hline
\texttt{InsideOf($o_1$,$o_2$)} & Object $o_1$ is inside of object $o_2$ if we can find two orthogonal axes crossing at $o_1$ center of mass that intersect $o_2$ collision mesh in both directions.\\ 
\hline
\texttt{OnTopOf($o_1$,$o_2$)} & Object $o_1$ is on top of object $o_2$ if $o_2 \in \texttt{InSameNegativeVerticalAxisObjs}(o_1) \land o_2 \not\in \texttt{InSamePositiveVerticalAxisObjs}(o_1) \land \texttt{InContactWith}(o_1, o_2)$, where $\texttt{InSamePositive/NegativeVerticalAxisObjs}(o_1)$ is the list of objects in the same positive/negative vertical axis as $o_1$ and $\texttt{InContactWith}(o_1, o_2)$ is whether the two objects are in physical contact.\\ 
\hline
\texttt{NextTo($o_1$,$o_2$)} & Object $o_1$ is next to object $o_2$ if $o_2 \in \texttt{InSameHorizontalPlaneObjs}(o_1) \land l2(o_1, o_2) < t_\mathit{NextTo}$, where $\texttt{InSameHorizontalPlaneObjs}(o_1)$ is a list of objects in the same horizontal plane as $o_1$, $l2$ is the L2 distance between the closest points of the two objects, and $t_\mathit{NextTo}$ is a distance threshold that is proportional to the average size of the two objects. \\
\hline
\texttt{InContactWith($o_1$,$o_2$)} & Object $o_1$ is in contact with $o_2$ if their surfaces are in contact in at least one point, i.e., $o_2 \in \texttt{InContactObjs}(o_1)$. \\ 
\hline
\texttt{ConnectedWith($o_1$,$o_2$)} & Object $o_1$ is connected with $o_2$ if $o_2 \in \texttt{ConnectedObjs}(o_1)$. \\ 
\hline
\texttt{Under($o_1$,$o_2$)} & Object $o_1$ is under object $o_2$ if $o_2 \in \texttt{InSamePositiveVerticalAxisObjs}(o_1)$ $\land o_2 \not\in \texttt{InSameNegativeVerticalAxisObjs}(o_1)$.\\
\hline
\texttt{OnFloor($o_1$,$o_2$)} & Object $o_1$ is on the room floor $o_2$ if $\texttt{InContactWith}(o_1, o_2)$ and $o_2$ is of \texttt{Room} type. \\ 
\hline
\texttt{Open(o)} & Any joints (internal articulated degrees of freedom) of object $o$ are open. Only joints that are relevant to consider an object \texttt{Open} are used in the predicate computation, e.g. the door of a microwave but not the buttons and controls. To select the relevant joints, object models of categories that can be \texttt{Open} undergo an additional annotation that produces a \texttt{RelevantJoints} list. A joint is considered open if its joint state $q$ is 5\% over the lower limit, i.e. $q > 0.05(q_\mathit{UpperLimit} - q_\mathit{LowerLimit}) + q_\mathit{LowerLimit}$. \\
\hline
\texttt{Cooked(o)} & The temperature of object $o$ was over the cooked threshold, $T_\mathit{cooked}$, and under the burnt threshold, $T_\mathit{burnt}$, at least once in the history of the simulation episode, i.e., $T_\mathit{cooked} \leq T_o^\mathit{max} < T_\mathit{burnt}$. We annotate the cooked temperature $T_\mathit{cooked}$ for each object category that can be \texttt{Cooked}.\\
\hline
\texttt{Burnt(o)} & The temperature of object $o$ was over the burnt threshold, $T_\mathit{burnt}$, at least once in the history of the simulation episode, i.e., $T_o^\mathit{max} \geq T_\mathit{burnt}$. We annotate the burnt temperature $T_\mathit{burnt}$ for each object category that can be \texttt{Burnt}.\\
\hline
\texttt{OnFire(o)} & The temperature of object $o$ is above the on-fire threshold, $T_\mathit{onfire}$, i.e., $T_o \leq T_\mathit{onfire}$. We assume as default on-fire temperature $T_\mathit{onfire}=300^\circ C$, a value that can be adapted per object category and model.\\
\hline
\texttt{Frozen(o)} & The temperature of object $o$ is under the freezing threshold, $T_\mathit{frozen}$, i.e., $T_o \leq T_\mathit{frozen}$. We assume as default freezing temperature $T_\mathit{frozen}=0^\circ C$, a value that can be adapted per object category and model.\\
\hline
\texttt{Heated(o)} & The temperature of object $o$ is above the heated threshold, $T_\mathit{heated}$, i.e., $T_o \leq T_\mathit{heated}$. We assume as default heated temperature $T_\mathit{heated}=75^\circ C$, a value that can be adapted per object category and model.\\
\hline
\texttt{Boiled(l)} & The temperature of liquid $l$ is above the boiling point, $T_\mathit{boiled}$, i.e., $T_o \leq T_\mathit{boiled}$. We assume as default boiling point $T_\mathit{boiling}=100^\circ C$, a value that can be adapted per object category and model.\\
\hline
\texttt{Soaked(o, l)} & The soaked level $w$ of the liquid $l$ for the object $o$ is over a threshold, $w_\mathit{soaked}$, i.e., $w \geq w_\mathit{soaked}$. The default value for the threshold is $w_\mathit{soaked}=50$, (the object is soaked if it absorbs more than $50$ liquid particles), a value that can be adapted per object category and model and per liquid type.\\
\hline

\texttt{Filled(o, l)} & Object $o$ is filled by liquid $l$ if the number of particles of $l$ that is inside the \texttt{ContainerVolume} of $o$ is above a certain threshold percentage of the total volume. The default value for the threshold is $w_\mathit{filled}=0.5$. \\
\hline
\texttt{Covered(o, s)} & For \texttt{visualSubstance} $s$, check if the covered level $c$ of $s$ for the object $o$ is over a threshold, $c_\mathit{covered}$, i.e., $c \geq c_\mathit{covered}$; for \texttt{physicalSubstance} $s$, check if the number of particles of $s$ that are in contact with the object $o$ is over the same threshold. The default value for the threshold is $c_\mathit{covered}=50$ (50 substance particles), a value that can be adapted per object category and model, and per substance.\\
\hline
\texttt{ToggledOn(o)} & Object $o$ is toggled on or off. It is a direct query of the object's extended state $\mathit{TS}$, the toggled state.\\ 
\hline
\texttt{Sliced(o)} & Object $o$ is sliced or not. It is a direct access of the object's extended state $\mathit{SS}$, the sliced state.\\
\hline
\texttt{Broken(o)} & Object $o$ is broken or not. It is a direct access of the object's extended state $\mathit{BS}$, the broken state.\\
\hline
\texttt{Folded(o)} & Object $o$ is folded if its corresponding \texttt{ClothKeypoints} are sufficiently close to each other.\\
\hline
\texttt{Unfolded(o)} & Object $o$ is unfolded if its corresponding \texttt{ClothKeypoints} are sufficiently far from each other.\\
\hline
\texttt{Assembled(o)} & Object $o$ is assembled if all of its parts have been correctly connected: every pair of parts $p_i$ and $p_j$ are connected (or not) with each other (i.e. $\texttt{IsConnected}(p_i, p_j)$) in a specific way defined by each object model.\\
\hline
\texttt{Hung($o_1$,$o_2$)} & Object $o_1$ is hung onto object $o_2$ if they are connected, $\texttt{IsConnected}(o_1, o_2)$.\\
\hline
\texttt{Blended($o_1$ ... $o_n$)} & Objects $o_1$ to $o_n$ are blended if they are in contact with each other, i.e. $\texttt{InContactWith}(o_i, o_j)$ for all pairs.\\
\hline
\texttt{InFoVOfAgent(o)} & Object $o$ is in the field of view of the agent, i.e., at least one pixel of the image acquired by the agent's onboard sensors corresponds to the surface of $o$.\\
\hline
\texttt{InHandOfAgent(o)} & Object $o$ is grasped by the agent's hands (i.e. assistive grasping is activated on that object).\\
\hline
\texttt{InReachOfAgent(o)} & Object $o$ is within $d_\mathit{reach}=2$ meters away from the agent.\\
\hline
\texttt{InSameRoomAsAgent(o)} & Object $o$ is located in the same room as the agent.\\
\hline
\end{tabular}
}
\vspace{3pt}
\caption{\textbf{Logical Predicates:} Description of the checking functions. Adapted from~\cite{li2021ig2}.}
\label{table:obj_pred_checking}
\end{center}
\end{table}

\subsubsection{Logical predicates as sampling functions}
For each logical predicate that is relevant for \benchmark activities, we also define a generative function that samples a valid physical state given a binary logical state. This functionality is essential for infinite activity initialization. For example, if the initial conditions of the activity require a plate to be \texttt{OnTopOf} a dining table, there are an infinite number of exact poses of the plate that can satisfy this condition. Our generative functions will find a valid solution and physically put the plate on top of the table. Other examples include setting the \texttt{Temperature} of an object to make it \texttt{Frozen} or the joint configuration of an object to make it \texttt{Open}. The details of the sampling functions for all the logical predicates can be found in Table~\ref{table:obj_pred_checking}.

\begin{table}[!t]
\begin{center}
{\scriptsize%
\renewcommand{\arraystretch}{1.2}
\begin{tabular}{ | m{0.20\textwidth} | m{0.7\textwidth} | } 
\hline
\textbf{Predicate} & \textbf{Sampling Mechanism} \\ 
\hline
\texttt{InsideOf($o_1$,$o_2$)} & Only \texttt{InsideOf($o_1$,$o_2$) = True} can be sampled. $o_1$ is randomly sampled within $o_2$ using a ray-casting mechanism adopted from~\cite{srivastava2021behavior}. $o_1$ is guaranteed to be supported fully by a surface and free of collisions with any other object except $o_2$.\\ 
\hline
\texttt{OnTopOf($o_1$,$o_2$)} & Only \texttt{OnTopOf($o_1$,$o_2$) = True} can be sampled. $o_1$ is randomly sampled on top of $o_2$ using a ray-casting mechanism adopted from~\cite{srivastava2021behavior}. $o_1$ is guaranteed to be supported fully by a surface and free of collisions with any other object except $o_2$.\\ 
\hline
\texttt{ConnectedWith($o_1$,$o_2$)} & Create a rigid joint between the two objects if they are both rigid bodies, or an attachment between them otherwise.\\ 
\hline
\texttt{Under($o_1$,$o_2$)} & Only \texttt{Under($o_1$,$o_2$) = True} can be sampled. $o_1$ is randomly sampled on top of the floor region beneath $o_2$ using a ray-casting mechanism adopted from~\cite{srivastava2021behavior}. $o_1$ is guaranteed to be supported fully by a surface and free of collisions with any other object except the floor.\\
\hline
\texttt{OnFloor($o_1$,$o_2$)} & Only \texttt{OnFloor($o_1$,$o_2$) = True} can be sampled. $o_1$ is randomly sampled on top of $o_2$, which is the floor of a certain room, using the scene's room segmentation mask. $o_1$ is guaranteed to be supported fully by a surface and free of collisions with any other object except $o_2$.\\ 
\hline
\texttt{Open(o)} & To sample an object $o$ with the predicate \texttt{Open(o) = True}, a subset of the object's relevant joints (using the \texttt{RelevantJoints} model property) are selected, and each selected joint is moved to a uniformly random position between the openness threshold and the joint's upper limit. To sample an object $o$ with the predicate \texttt{Open(o) = False}, all of the object's relevant joints (using the \texttt{RelevantJoints} model property) are moved to a uniformly random position between the joint's lower limit and the openness threshold.\\
\hline
\texttt{Cooked(o)} & To sample an object $o$ with the predicate \texttt{Cooked(o) = True}, the object's \texttt{MaxTemperature} is updated to $\max(T_o^\mathit{max}, T_\mathit{cooked})$. Similarly, to sample an object $o$ with the predicate \texttt{Cooked(o) = False}, the object's \texttt{MaxTemperature} is updated to $\min(T_o^\mathit{max}, T_\mathit{cooked} - 1)$. \\
\hline
\texttt{Burnt(o)} & To sample an object $o$ with the predicate \texttt{Burnt(o) = True}, the object's \texttt{MaxTemperature} is updated to $\max(T_o^\mathit{max}, T_\mathit{burnt})$. Similarly, to sample an object $o$ with the predicate \texttt{Cooked(o) = False}, the object's \texttt{MaxTemperature} is updated to $\min(T_o^\mathit{max}, T_\mathit{burnt} - 1)$.\\
\hline
\texttt{OnFire(o)} & To sample an object $o$ with the predicate \texttt{OnFire(o) = True}, the object's \texttt{Temperature} is updated to a uniformly random temperature between $T_\mathit{onfire} + 10$ and $T_\mathit{onfire} + 50$. To sample an object $o$ with the predicate \texttt{OnFire(o) = False}, the object's \texttt{Temperature} is updated to $T_\mathit{onfire} - 1$. \\
\hline
\texttt{Frozen(o)} & To sample an object $o$ with the predicate \texttt{Frozen(o) = True}, the object's \texttt{Temperature} is updated to a uniformly random temperature between $T_\mathit{frozen} - 10$ and $T_\mathit{frozen} - 50$. To sample an object $o$ with the predicate \texttt{Frozen(o) = False}, the object's \texttt{Temperature} is updated to $T_\mathit{frozen} + 1$. \\
\hline
\texttt{Heated(o)} & To sample an object $o$ with the predicate \texttt{Heated(o) = True}, the object's \texttt{Temperature} is updated to a uniformly random temperature between $T_\mathit{heated} + 10$ and $T_\mathit{heated} + 50$. To sample an object $o$ with the predicate \texttt{Heated(o) = False}, the object's \texttt{Temperature} is updated to $T_\mathit{heated} - 1$. \\
\hline
\texttt{Boiled(l)} & To sample a liquid type $l$ with the predicate \texttt{Boiled(l) = True}, the \texttt{Temperature} of all particles of $l$ is updated to a uniformly random temperature between $T_\mathit{boiled} + 10$ and $T_\mathit{boiled} + 50$. To sample a liquid type $l$ with the predicate \texttt{Boiled(o) = False}, the \texttt{Temperature} of all particles of $l$ is updated to $T_\mathit{boiled} - 1$. \\
\hline
\texttt{Soaked(o, l)} & To sample an object $o$ and a liquid type $l$ with the predicate \texttt{Soaked(o, l) = True}, the object's \texttt{SoakedLevel} $w$ for $l$ is updated to match the \texttt{Soaked} threshold of $w_\mathit{soaked}$. To sample an object $o$ and a liquid type $l$ with the predicate \texttt{Soaked(o, l) = False}, the object's \texttt{SoakedLevel} $w$ is updated to 0.\\
\hline
\texttt{Filled(o, l)} & To sample an object $o$ and a liquid type $l$ with the predicate \texttt{Filled(o, l) = True}, an appropriate number of particles of $l$ are sampled inside the \texttt{ContainerVolume} of $o$ so that they fill up enough of its volume ($\geq w_\mathit{filled}=0.5$). To sample an object $o$ and a liquid type $l$ with the predicate \texttt{Filled(o, l) = False}, all particles of $l$ that are inside the \texttt{ContainerVolume} of $o$ (if any) are removed.\\
\hline
\texttt{Covered(o, s)} & To sample an object $o$ and a substance $s$ with \texttt{Covered(o, s) = True}, a fixed number of particles of $s$ are randomly placed on the surface of $o$ using a ray-casting mechanism adopted from~\cite{srivastava2021behavior}. To sample an object $o$  and a \texttt{visualSubstance} $s$ with \texttt{Covered(o, s) = False}, all particles of $s$ is removed from $o$ and the corresponding \texttt{CoveredLevel} is set to 0.\\
\hline
\texttt{ToggledOn(o)} & The \texttt{ToggledState} of the object is updated to match the required predicate value. \\ 
\hline
\texttt{Sliced(o)} & The \texttt{SlicedState} of the object is updated to match the required predicate value. Also, the whole object are replaced with the two halves, that will be placed at the same location and inherit the extended states from the whole object (e.g. \texttt{Temperature}).\\
\hline
\texttt{Broken(o)} & The \texttt{BrokenState} of the object is updated to match the required predicate value. Also, the whole object is broken down into pieces, that will be placed at the same location.\\
\hline
\end{tabular}
}
\vspace{3pt}
\caption{\textbf{Logical Predicates:} Description of the sampling functions. Adapted from~\cite{li2021ig2}.}
\label{table:obj_pred_sampling}
\end{center}
\end{table}


\subsection{AMT Visual Realism Study}
\label{amt_study_appx}

We conducted an Amazon Mechanical Turk (AMT) study to evaluate \simulator's relative visual realism compared to multiple other simulation environments. We selected 50 representative $1280 \times 720$ images from \simulator, AI2-Thor, ThreeDWorld, Habitat 2.0, and iGibson 2.0, and shuffled them randomly into 50 groups of 5 images, where each group contained a unique image from each simulation environment. For each group, participants were asked to rank the images in terms of visual realism, assigning the most realistic images of the group $1$, and subsequent images $2, ..., 5$. Image shuffling was randomized between participants. When presenting our results, we invert the aggregated mean and standard deviation across 60 participants, such that a score of a $5$ would represent the most visually realistic images.

Our criteria for selecting the images are the following: (a) we only included photos taken from \textit{fully interactive} scenes for a fair comparison, and (b) rendering must come from within the simulation environment, without customized tuning (i.e. new users can expect this visual quality off-the-shelf with minimal adjustment).

\begin{figure}[t]
\centering
\captionsetup[subfigure]{aboveskip=1pt,belowskip=1pt}
\begin{subfigure}[t]{0.304\textwidth}{%
\includegraphics[width=\textwidth]{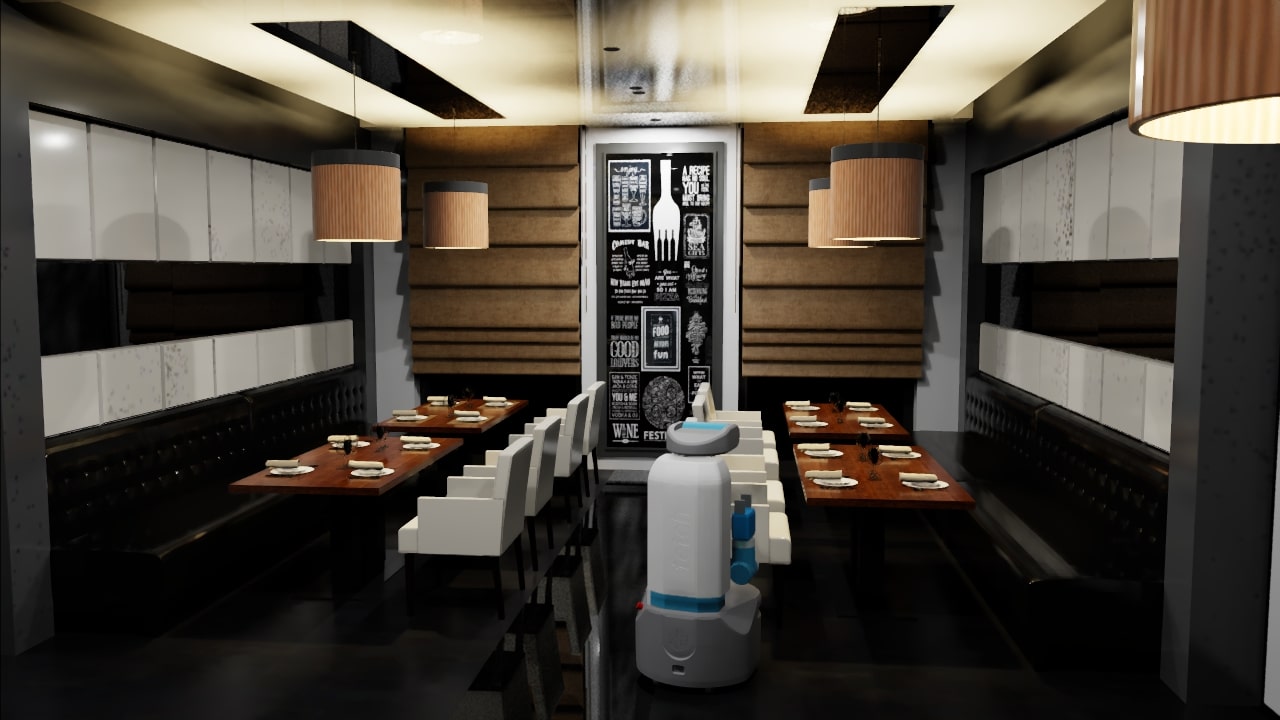}}%
\caption*{\scriptsize3rd Person View}%
\end{subfigure}%
\hfill%
\begin{subfigure}[t]{0.17\textwidth}{\includegraphics[width=\textwidth]{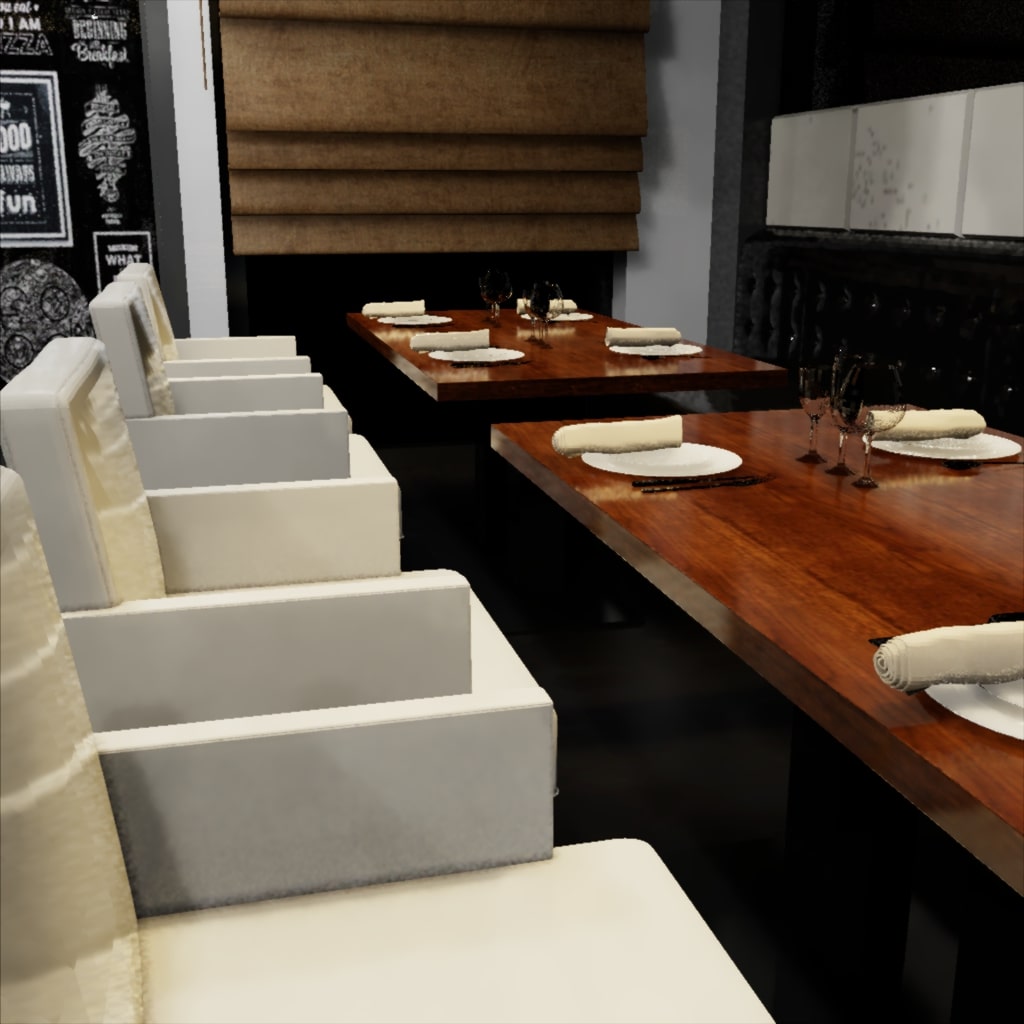}}%
\caption*{\scriptsize RGB}%
\end{subfigure}%
\hfill%
\begin{subfigure}[t]{0.17\textwidth}{\includegraphics[width=\textwidth]{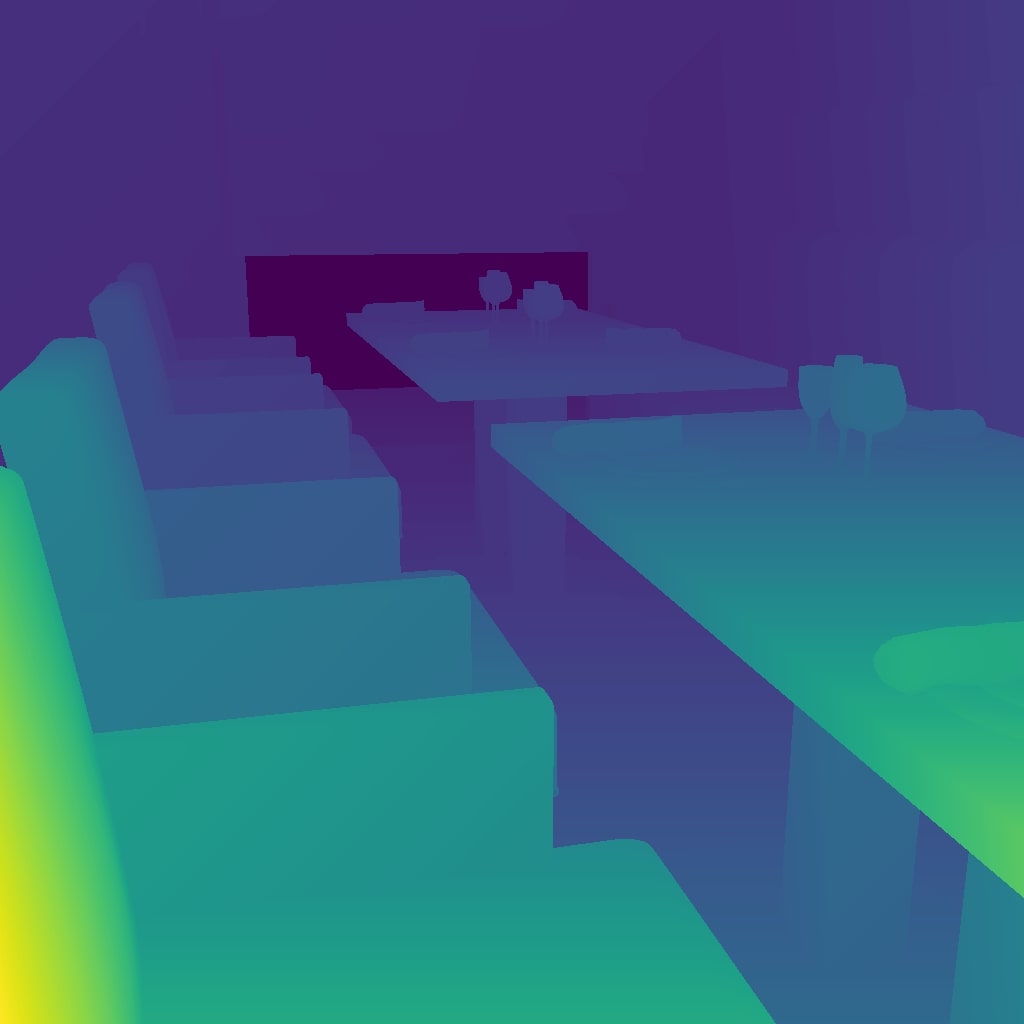}}%
\caption*{\scriptsize Depth}%
\end{subfigure}%
\hfill%
\begin{subfigure}[t]{0.17\textwidth}{\includegraphics[width=\textwidth]{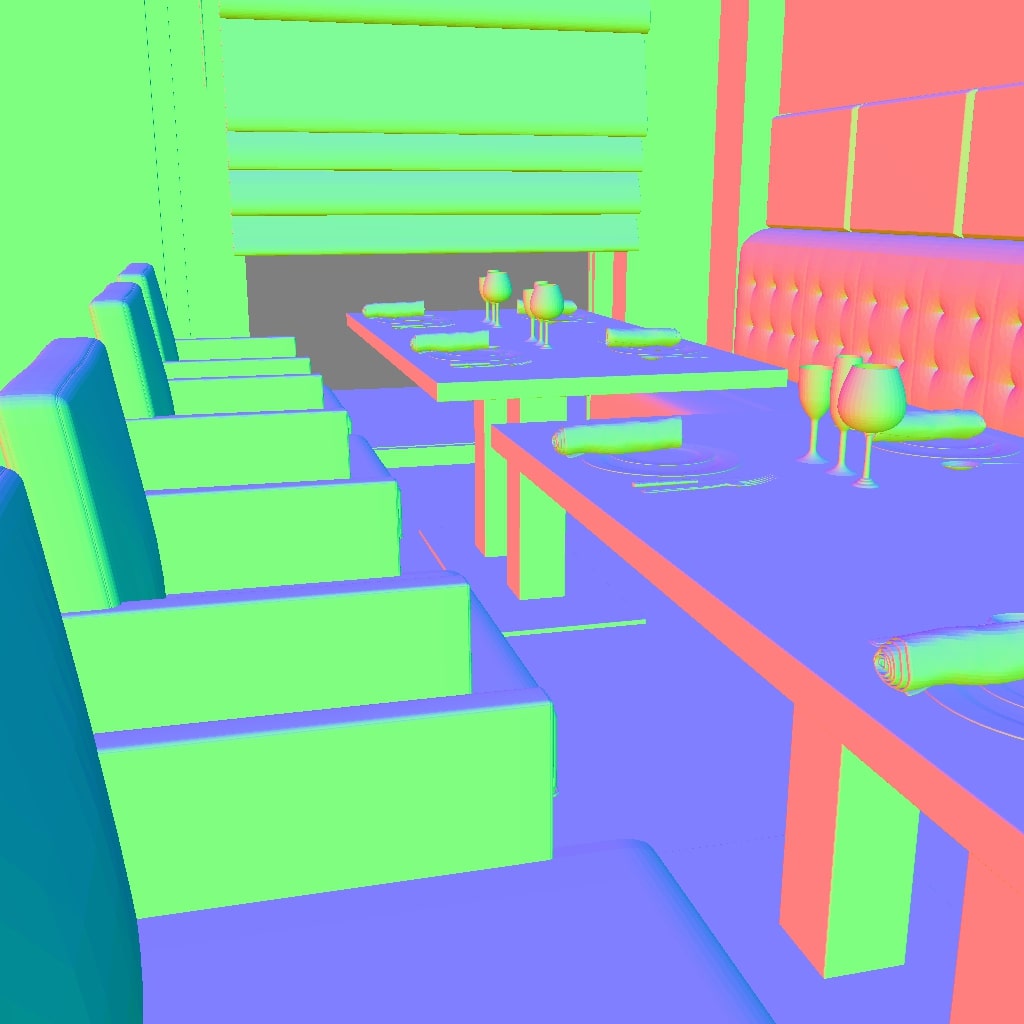}}%
\caption*{\scriptsize Normals}%
\end{subfigure}%
\hfill%
\begin{subfigure}[t]{0.17\textwidth}{\includegraphics[width=\textwidth]{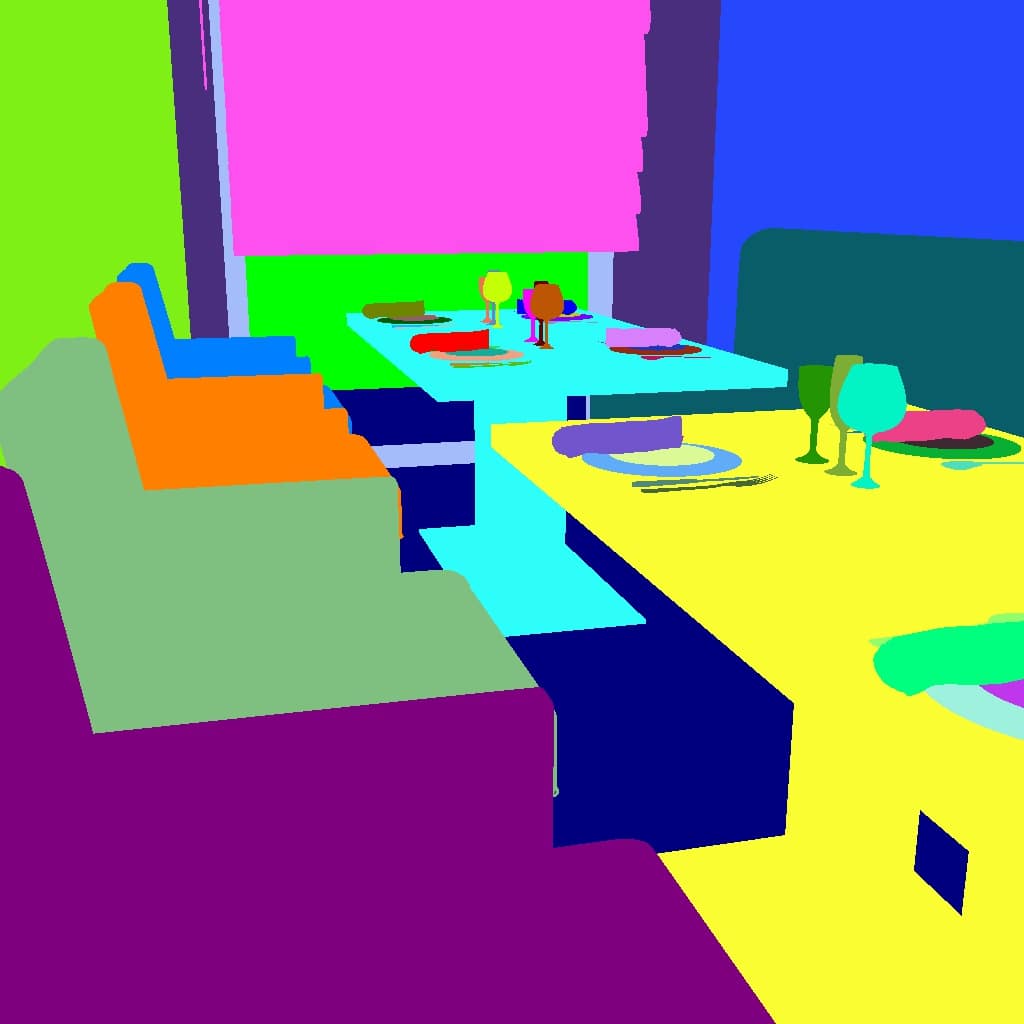}}%
\caption*{\scriptsize Inst. Segm.}%
\end{subfigure}%
\captionsetup[figure]{font=small,skip=-2pt}%
\caption{\textbf{Visual modalities provided by \simulator.} The robot observes the scene (left, 3rd person view) and obtains visual observations from its onboard sensors including RGB images, depth, normals, and instance segmentation. Rendered RGB images are highly realistic, and, combined with the diverse set of visual observations, can be used to train visual policies.}
\label{fig:sim_modalities}
\end{figure}

\subsection{Visual Modalities}
\simulator provides diverse perceptual modules to capture realistic sensor modalities from an agent's perspective, including RGB, Depth, Semantic Segmentation, Normal, and Optical Flow images, in addition to non-visual modalities such as proprioception and LiDAR scans (Fig.~\ref{fig:sim_modalities}). \simulator also provides varying abstraction levels for agent action spaces, including low-level control, assistive manipulation, and primitive skill execution. Overall, these modules are intended to be useful to the broad embodied AI research community, with the goal of accelerating breakthroughs on \benchmark.

\subsection{Performance Benchmark}

\revision{
To evaluate the performance of \simulator under different conditions, we performed rigorous speed test in two representative scenes (\texttt{Rs\_int} and \texttt{house\_single\_floor}) with different number of objects on a single-GPU, single-process setup. We adopt the ``idle'' setup from \citet{li2021ig2} and \citet{szot2021habitat}: a robot is placed in the scene (except the last row "- Robot"), and stays still with zero velocity action. At each time step, the simulator runs the physics simulation, extended object state update, and transition machine update loop, and renders a $128\times128$ RGB image. We use action time step of $t_a = \frac{1}{60}\text{s}$ and physics time step of $t_s = \frac{1}{60}\text{s}$. Our benchmark runs on a Ubuntu machine with Intel(R) Core(TM) i7-10700K CPU @ 5.10GHz and one Nvidia GeForce RTX 3080 GPU in a single process setting. The results are summarized in Table~\ref{tab:benchmark}.
}

\revision{
\simulator runs at a comparable speed to previous works like iGibson 2.0~\cite{li2021ig2} under similar settings, but with much higher rendering quality thanks to ray-tracing. It maintains a reasonable speed even for a large scene with over 600 objects. \simulator is also highly configurable and provides flexible interface for users to balance between simulation fidelity and speed given their use cases and research interests. If the user isn't interested in fluid or cloth in their tasks, they can turn off these features to harvest performance speedup. Similarly, if the user is only interested in kinematics-only rearrangement tasks, or non-robotics embodied AI applications (e.g. a virtual camera), they can turn off object state update, or remove the robot, respectively. We are actively working on improving aspects that can provide simulation speedups, e.g. in areas like object sleeping, mesh simplification, more efficient object state update. Furthermore, since real-time ray-tracing and fluid/cloth simulation is still an active area of research, we expect the upcoming hardware and software advances from Nvidia will lead to significant performance improvements in these areas.
}

\begin{table}[t]
    \centering
    \begin{tabular}{l|cc}
    \toprule
    \multicolumn{1}{c|}{\multirow{2}{*}{Eval. Conditions}} & \multicolumn{2}{c}{Scene} \\
    \multicolumn{1}{c|}{} & \texttt{Rs\_int} (81 objs) & \texttt{house\_single\_floor} (621 objs) \\ \midrule
    Full Feature Set & $24$ & $11$ \\
    - Fluid and Cloth & $58$ & $26$ \\
    - Object State Update & $77$ & $55$ \\ 
    - Robot & $90$ & $60$ \\ \bottomrule
    \end{tabular}
    \caption{\revision{Benchmarking \simulator performance: simulation steps per second (SPS, higher better) in two representative scenes with 81 and 621 objects, under different evaluation conditions.}}
    \label{tab:benchmark}
\end{table}


\section{Baselines Details}
\label{sec:baseline_appendix}

In this section, we include additional information about the baselines evaluated and analyzed in the main paper: the visuomotor control baseline and two variants of the baselines using action primitives, with and without the history of observations.

\begin{figure}[t]
    \centering
    \includegraphics[width=0.95\textwidth]{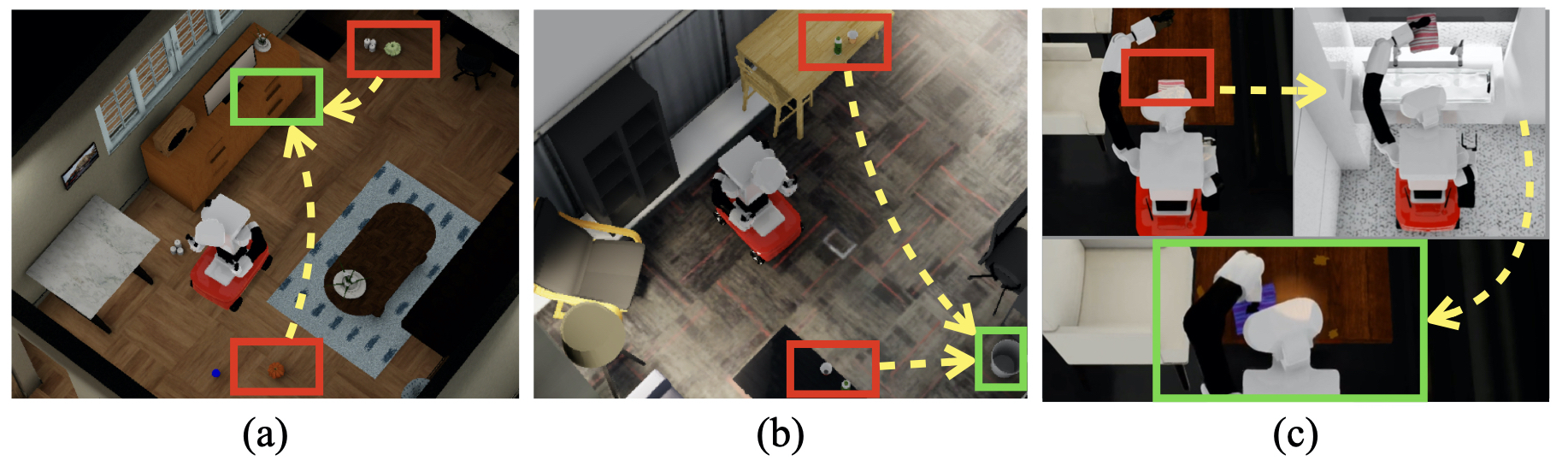}
    \caption{\textbf{Activities used in our evaluation.} From left to right: (a). \texttt{StoreDecoration}, (b). \texttt{CollectTrash} and (c). \texttt{CleanTable}. The rectangles indicate relevant locations for the activities, e.g., locations with objects to grasp (\textit{red rectangles}) or to use/place them (\textit{green rectangles}). Even though these activities are some of the most simple in \benchmark, they are still very long-horizon, requiring hundreds of low-level commands or the correct concatenation of multiple action primitives.}
    \label{fig:baseline_task}
\end{figure}

\subsection{Network Architecture}
\begin{figure}
    \centering
    \begin{tabular}{cc}        
        \includegraphics[width=0.45\linewidth]{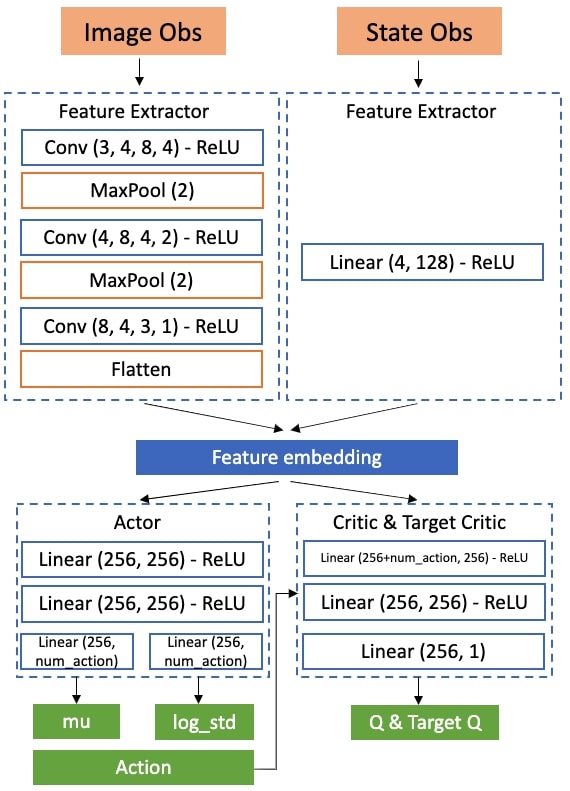} &
        \includegraphics[width=0.45\linewidth]{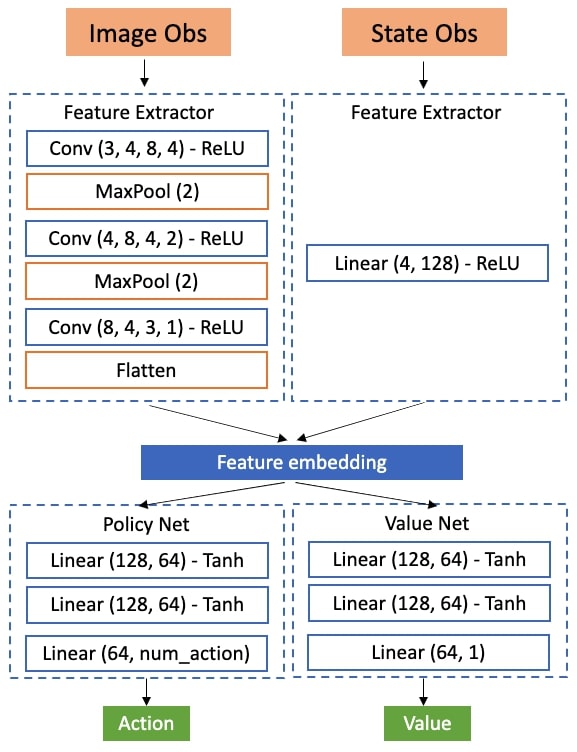} \\
        a) SAC Architecture & b) PPO Architecture \\
    \end{tabular}
    \caption{
    Policy architecture of SAC (\rlvmc) and PPO (\rlprim and \rlprimhist). The policy maps egocentric visual observations and proprioceptive information into an action that controls the robot base, arm, and gripper. PPO outputs a discrete high-level action primitive while SAC outputs a continuous low-level joint control directly.
    }
    \label{fig:appendix_ppo_vmc_arch}
\end{figure}

For \rlprim and \rlprimhist that use PPO as the underlying RL algorithm, the architecture consists of a visual feature extractor that takes the egocentric visual observation as input and a state encoder neural network which takes whether the robot is grasping an object or not as input. The image input is normalized using a per-channel moving average. These two encodings are passed into an MLP module, which is then processed by a value head to predict the value for the given observations and an action head to produce a discrete action that corresponds to an action primitive executed on an object. The size of input images is $128\times 128\times 3$. The feature extractor is a sequential architecture of Conv-ReLU-MaxPooling-Flatten. MLP converts the feature into a $128$-dimensional vector. The details of \rlprim's network architecture is illustrated in Fig.~\ref{fig:appendix_ppo_vmc_arch}~(b). For \rlvmc that uses SAC as the underlying RL algorithm, we use the same feature extractor as \rlprim. \rlvmc consists of an actor network, a critic network, and a target critic network. All of them are MLPs with ReLU activation functions. Fig.~\ref{fig:appendix_ppo_vmc_arch}~(a) illustrates the details of \rlvmc's network architecture. \rlvmc has the continuous action space and outputs low-level control actions directly.

\subsection{Task settings}
Fig.~\ref{fig:baseline_task} depicts the three activities we consider in our experiments: 
\vspace{-2mm}
\begin{itemize}[leftmargin=2em]
    \item \texttt{StoreDecoration}: a tidying activity where the agent must pick up and store Halloween decorations into a cabinet operating articulated objects. Action space: \texttt{navigate}, \texttt{pick}, \texttt{place}, \texttt{push}. The goal is to store two pumpkins into a drawer (see Fig.~\ref{fig:baseline_task} (a)).
    \item \texttt{CollectTrash}: A collecting activity that requires the agent to gather empty bottles and cups and throw them into a trash bin. Action space: \texttt{navigate}, \texttt{pick}, \texttt{place}. The goal is to throw two bottles and two cups into a trash bin (see Fig.~\ref{fig:baseline_task} (b)).
    \item \texttt{CleanTable}: A cleaning activity that involves challenging cloth manipulation and fluids for table cleaning. Action space: \texttt{navigate}, \texttt{pick}, \texttt{dip}, \texttt{wipe}. The goal is to cleaning a table with a soaked cloth (see Fig.~\ref{fig:baseline_task} (c)). 
\end{itemize}
Each activity takes place in a different B1K scene, \texttt{Rs\_int}, \texttt{mockup\_apt}, and \texttt{restaurant\_hotel}. 


\subsection{Training details}

\paragraph{Learning objectives.} 
For \rlprim and \rlprimhist, we use an on-policy reinforcement learning algorithm Proximal Policy Optimization (PPO). $\theta$ is the policy parameter, $\hat{E}_t$ denotes the empirical expectation over timesteps. $r_{t}$ is the ratio of the probability under the new and old policies, respectively. $\hat{A}_t$ is the estimated advantage at time $t$. $\epsilon$ is the clipping hyperparameter. The objective function of PPO is shown in Equation~\ref{equ:ppo}:
\begin{equation}
    C^{CLIP}(\theta) = \hat{E}_t[\min r_t(\theta)\hat{A}_t, clip(r_t(\theta), 1-\epsilon, 1+\epsilon)\hat{A}_t].
    \label{equ:ppo}
\end{equation}
For \rlvmc, we use Soft Actor Critic (SAC) algorithm. The objective function is shown in Equation~\ref{equ:sac}:
\begin{equation}
    \pi^* = \arg\max_\pi E_{\tau\sim\pi}\left[\sum_{t=0}^\infty \gamma^t \left(R(s_t, a_t, s_{t+1})+\alpha H (\pi(\cdot|s_t)\right)\right],
    \label{equ:sac}
\end{equation}
where $\alpha$ is the coefficient that determines the weight of two terms and $H$ denotes the entropy.

\paragraph{Reward function.} 
We use the success signal provided by the BDDL activity definition as the reward function for training the policy in all methods. For instance, in the \texttt{StoreDecorations} task, the BDDL task goal definition is \texttt{Forall(decoration)\{Inside(decoration, cabinet)\}}. The agent receive a positive signal if and only if one of the \texttt{decoration} is physically placed inside the \texttt{drawer} of the \texttt{cabinet}. This is a challenging sparse reward setting since the agent needs to plan multiple steps to reach the final goal without any intermediate subgoal rewards (e.g. \texttt{push} open the \texttt{drawer} must take place before \texttt{pick} and \texttt{place} the \texttt{decoration}, but no reward signal will be given to the \texttt{push} open action).

\paragraph{Hyperparameters.} 
The hyperparameters for SAC and PPO in the baselines are summarized in Table~\ref{tab:vmc_hyper_param} and Table~\ref{tab:ppo_hyper_param}. Both SAC and PPO are trained for 30,000 time steps. We train with seeds $0, 1, 2$ and evaluate with seed $0$.

\paragraph{Computation.}
\revision{For each run of our experiments, we use either a single Nvidia GeForce RTX 2080 Ti or a single Nvidia RTX A-6000 GPU, together with Intel Xeon CPU, and 40GB of RAM. During training, the GPU memory usage is around 9GB. Depending on which task, the total training iterations range between 10k to 25k, and the total wall-clock training time range between 3.75 to 7.5 hours.}

\begin{table}[t]
\centering
\begin{minipage}[t]{.41\linewidth}%
    \begin{tabular}{c|c}
    \toprule
        Learning Rate & $0.0003$ \\
        Buffer Size & $300$ \\
        Batch Size &  $64$ \\
        Discount ($\gamma$) & $0.99$ \\
        Soft Update Coefficient & $0.005$ \\
    \bottomrule
    \end{tabular}
    \caption{Hyperparameters of SAC for the \rlvmc baseline}
    \label{tab:vmc_hyper_param}
\end{minipage}
\hfill
\begin{minipage}[t]{.41\linewidth}%
    \begin{tabular}{c|c}
    \toprule
        Learning Rate & $0.0003$ \\
        Buffer Size & $300$ \\
        Batch Size &  $64$ \\
        Discount ($\gamma$) & $0.99$ \\
        GAE Parameter $\gamma_{gae}$ & $0.99$ \\
        Clipping Parameter $\epsilon$ & $0.2$ \\
        Entropy Coeff $c_1$ & $0.0$ \\
        VF Coeff $c_2$ & $0.5$ \\
    \bottomrule
    \end{tabular}
    \caption{Hyperparameters of PPO for the \rlprim and \rlprimhist baselines}
    \label{tab:ppo_hyper_param}
\end{minipage}
\end{table}

\subsection{Action Primitives} 
\benchmark activities are long-horizon which require hundreds if not thousands of environment steps (low-level robot control signals) to be completed.  A way to overcome this challenge (e.g., reinforcement learning) is to modify the original action space with a set of \textit{action primitives}, i.e., time-extended actions that correspond to multiple low-level commands and that achieve some expected outcome, such as grasping an object or navigating to a location. For our baselines, we defined six of these primitives, and implemented them using a sampling-based motion planner.

All the primitives are compositional: collision-free paths of the entire robot to navigate to a location, collision-free paths of the robot's arm to reach a \texttt{6D\_pose} with the end-effector, or a predefined arm joint configuration, trajectories where the end-effector follows a \texttt{line} in Cartesian space, possibly colliding/interacting with the environment, and sequences of actions to \texttt{open}/\texttt{close} the robot's gripper while holding the position.
All manipulation primitives start with a trajectory to move the arm from a tucked configuration (folded arm) to an untucked configuration to enable subsequent arm interaction, and end with the inverse motion, from untucked to tucked configuration, to allow subsequent collision-free navigation.
The action primitives take as parameter an object to apply the action on; we assume access to the 3D position of the objects to obtain the necessary parameters to query the motion planner. This procedure is replicated on the real robot, where we combine detections from YOLO~\cite{redmon2018yolov3} with information from the depth map of the RGB-D images to obtain the parameters (see Sec.~\ref{as:s2r}). For the navigation actions, we assume a set of known relevant locations to navigate to (next to the rectangles in Fig.~\ref{fig:baseline_task}) 

\begin{figure}[h!]
\centering
\begin{subfigure}[t]{0.8\textwidth}{\centering
\includegraphics[trim=0 200 0 200,clip,width=0.33\linewidth]{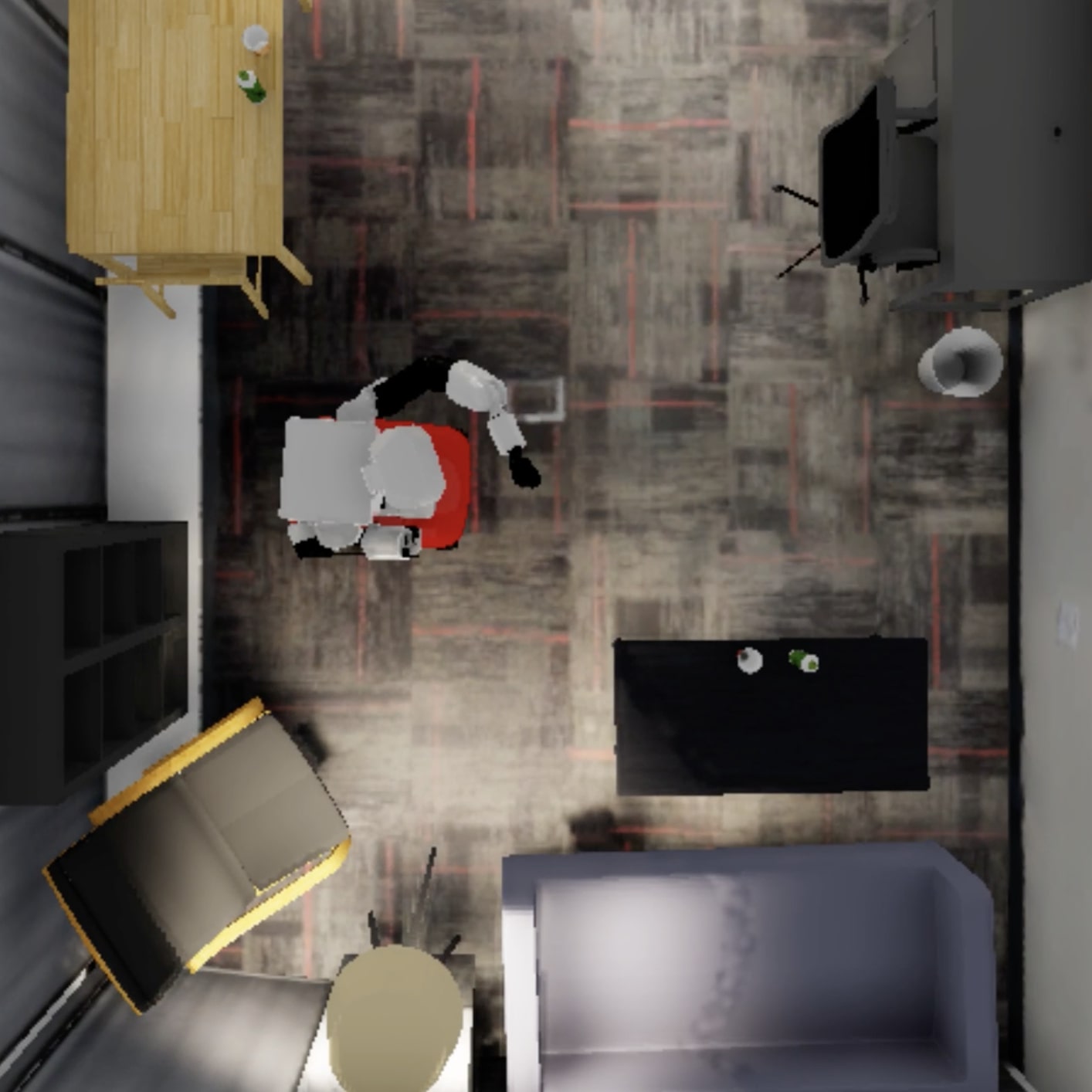}%
\hfill%
\includegraphics[trim=0 200 0 200,clip,width=0.33\linewidth]{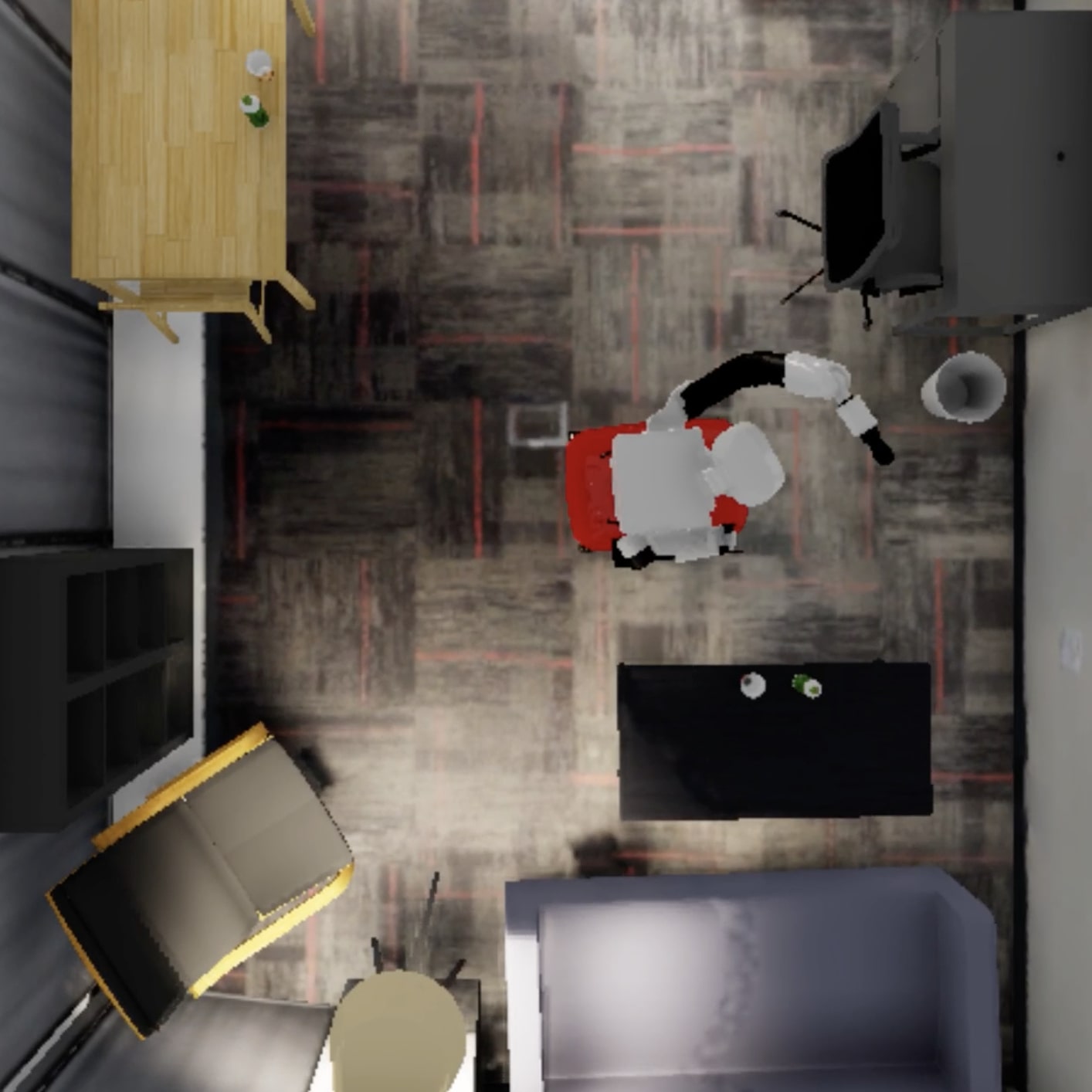}%
\hfill%
\includegraphics[trim=0 200 0 200,clip,width=0.33\linewidth]{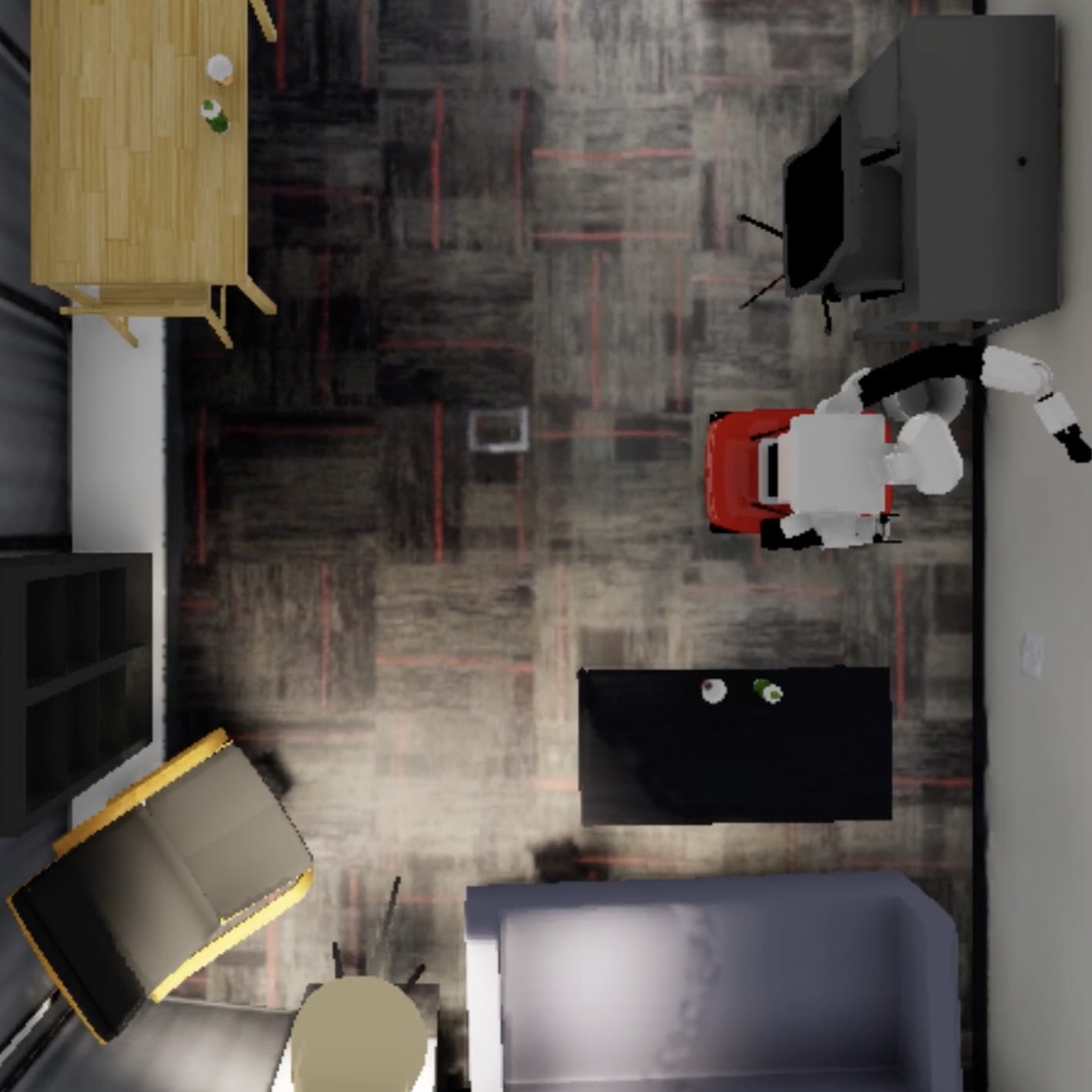}%
}
\vspace{-5pt}
\caption{\texttt{navigate}}
\vspace{10pt}
\end{subfigure}
\begin{subfigure}[t]{0.8\textwidth}{\centering
\includegraphics[trim=0 200 0 200,clip,width=0.33\linewidth]{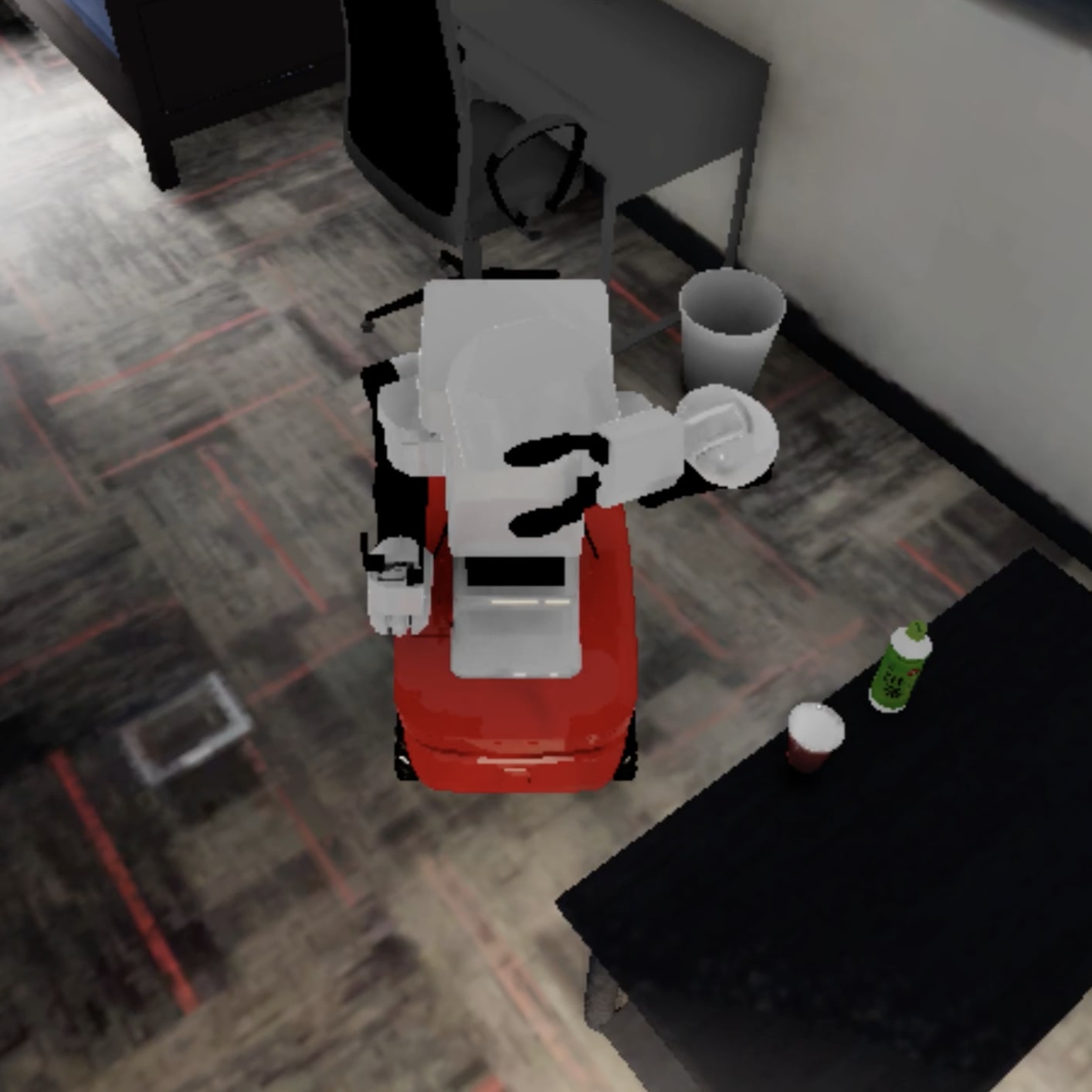}%
\hfill%
\includegraphics[trim=0 200 0 200,clip,width=0.33\linewidth]{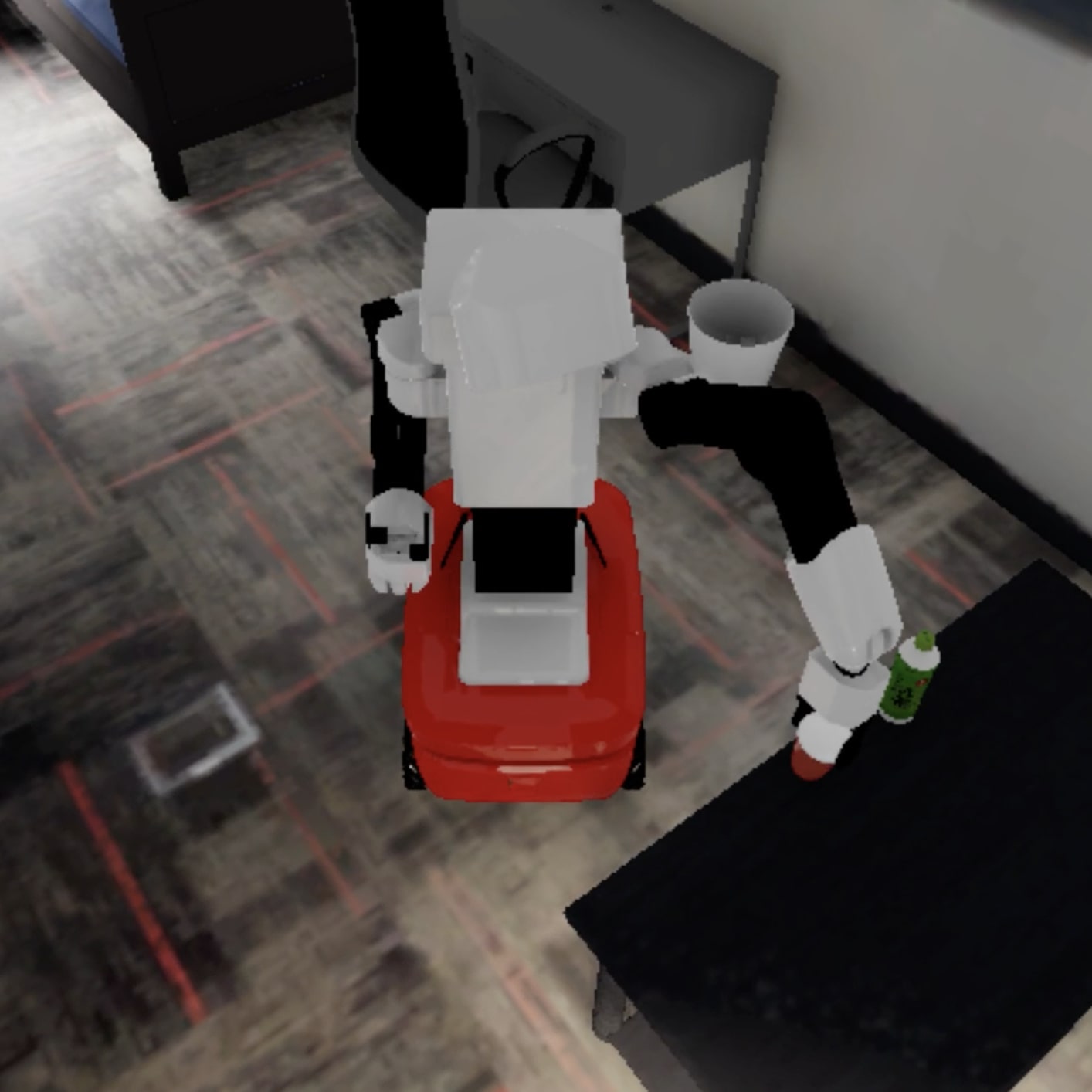}%
\hfill%
\includegraphics[trim=0 200 0 200,clip,width=0.33\linewidth]{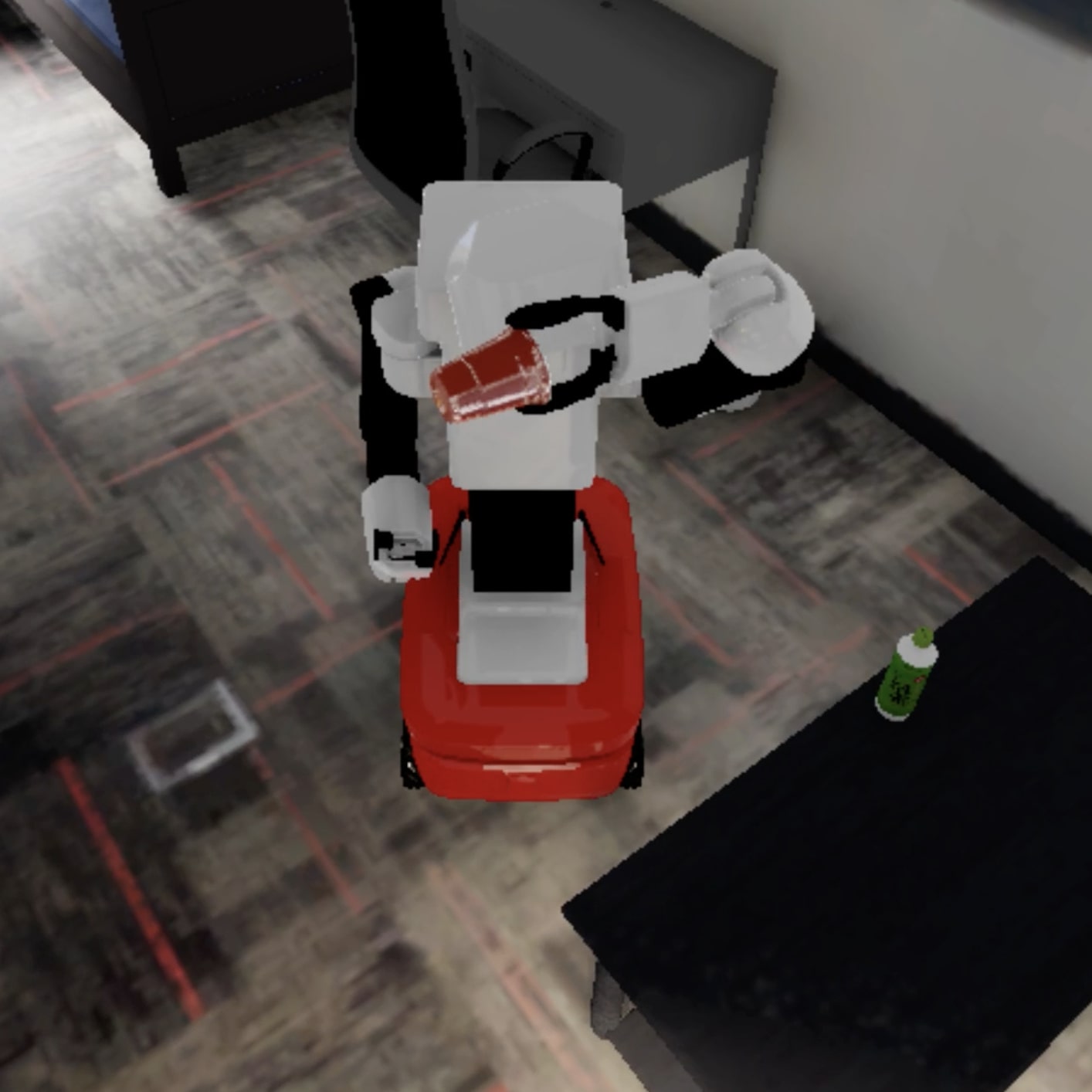}
}%
\vspace{-5pt}%
\caption{\texttt{pick}}%
\vspace{10pt}%
\end{subfigure}
\begin{subfigure}[t]{0.8\textwidth}{\centering
\includegraphics[trim=0 0 0 250,clip, width=0.33\linewidth]{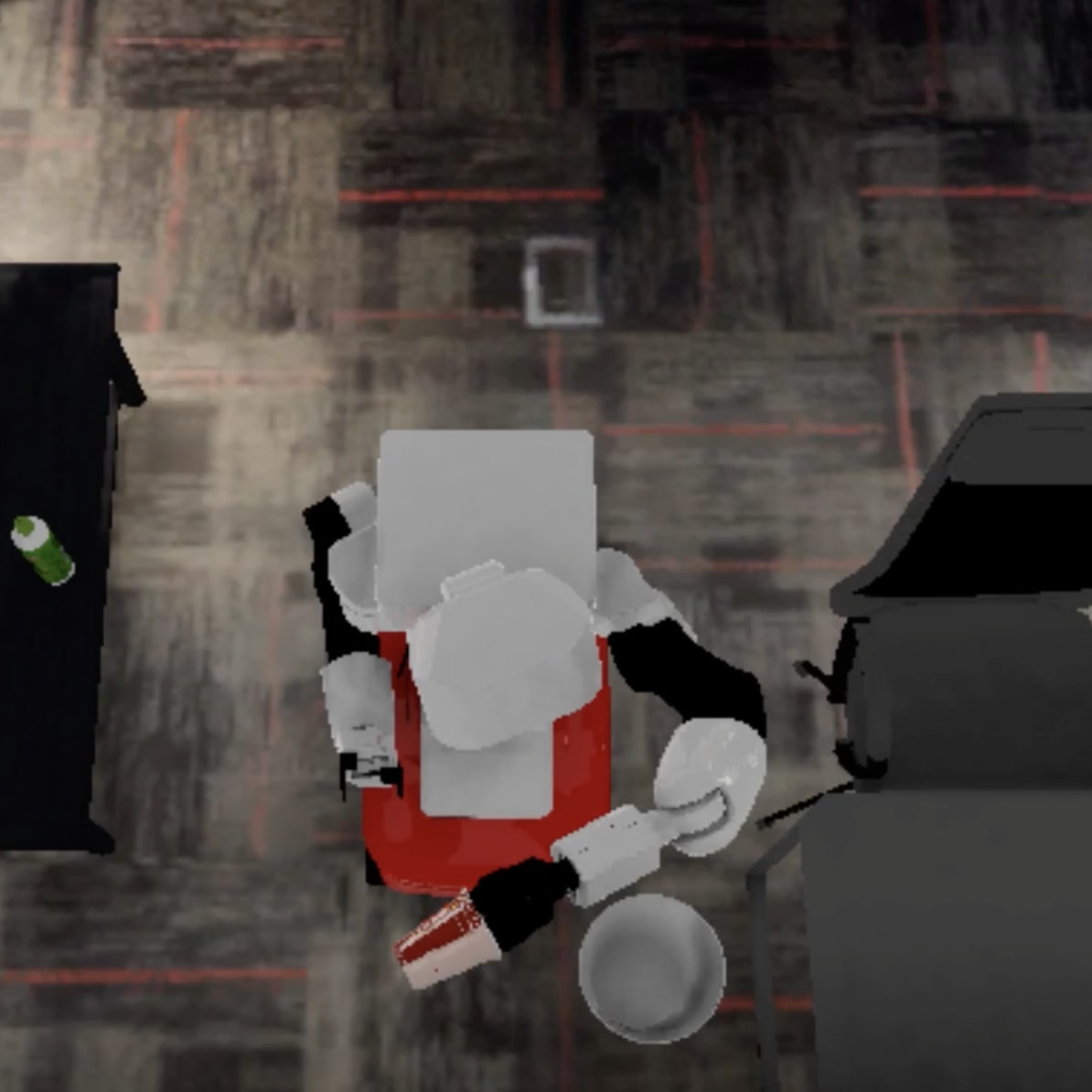}%
\hfill%
\includegraphics[trim=0 0 0 250,clip,width=0.33\linewidth]{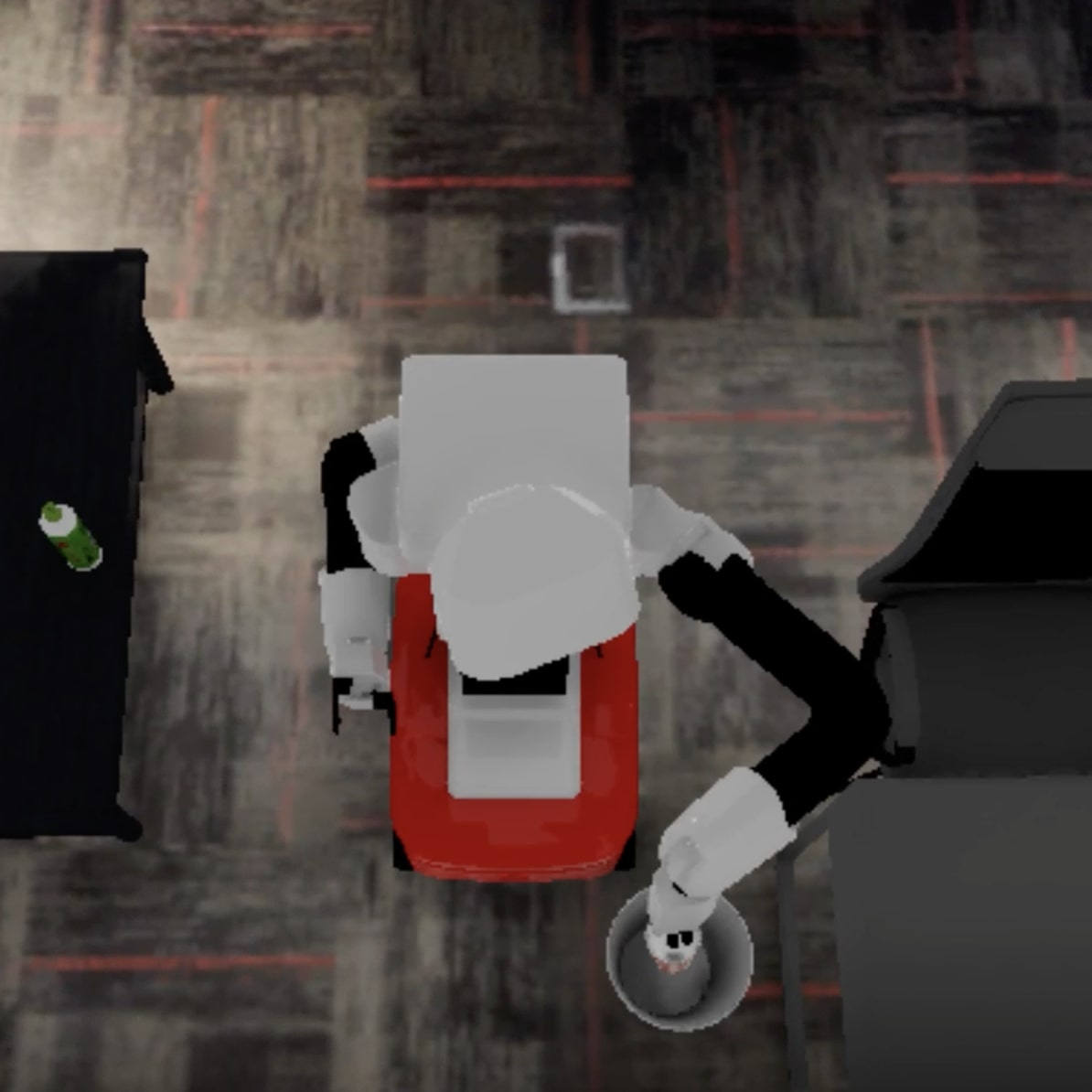}%
\hfill%
\includegraphics[trim=0 0 0 250,clip,width=0.33\linewidth]{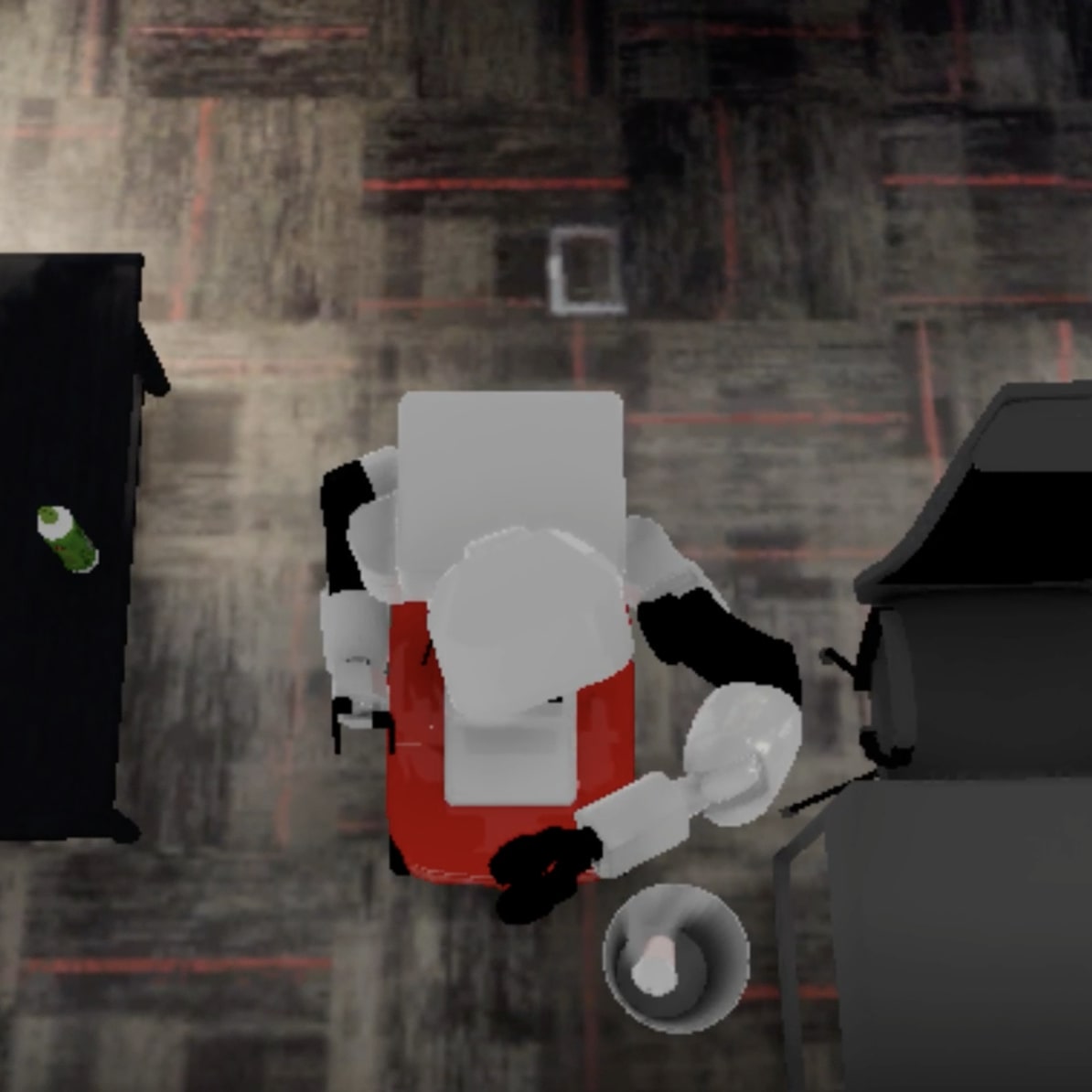}
}%
\vspace{-5pt}%
\caption{\texttt{place}}%
\vspace{10pt}%
\end{subfigure}
\begin{subfigure}[t]{0.8\textwidth}{\centering
\includegraphics[trim=0 220 0 20,clip,width=0.33\linewidth]{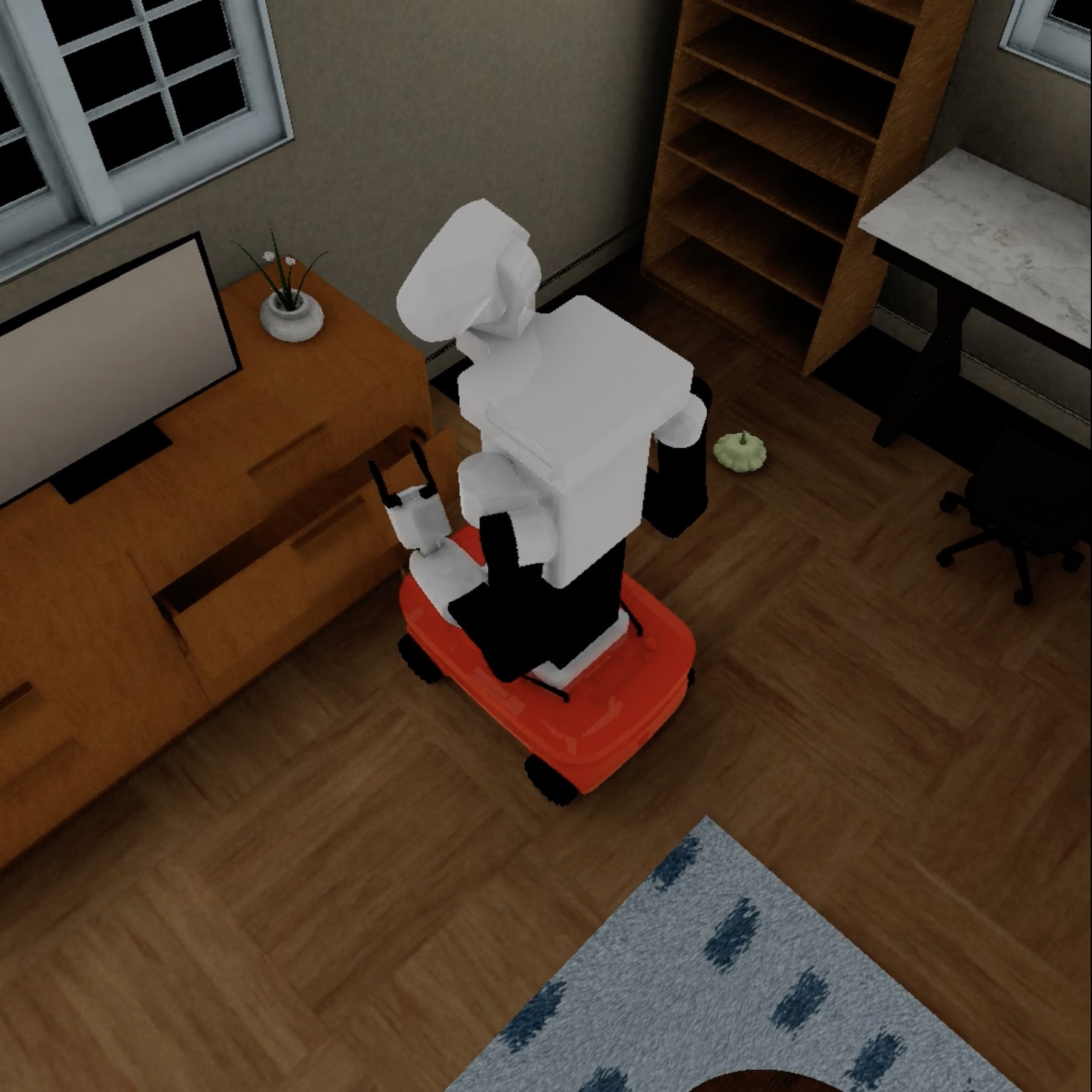}%
\hfill%
\includegraphics[trim=0 220 0 20,clip,width=0.33\linewidth]{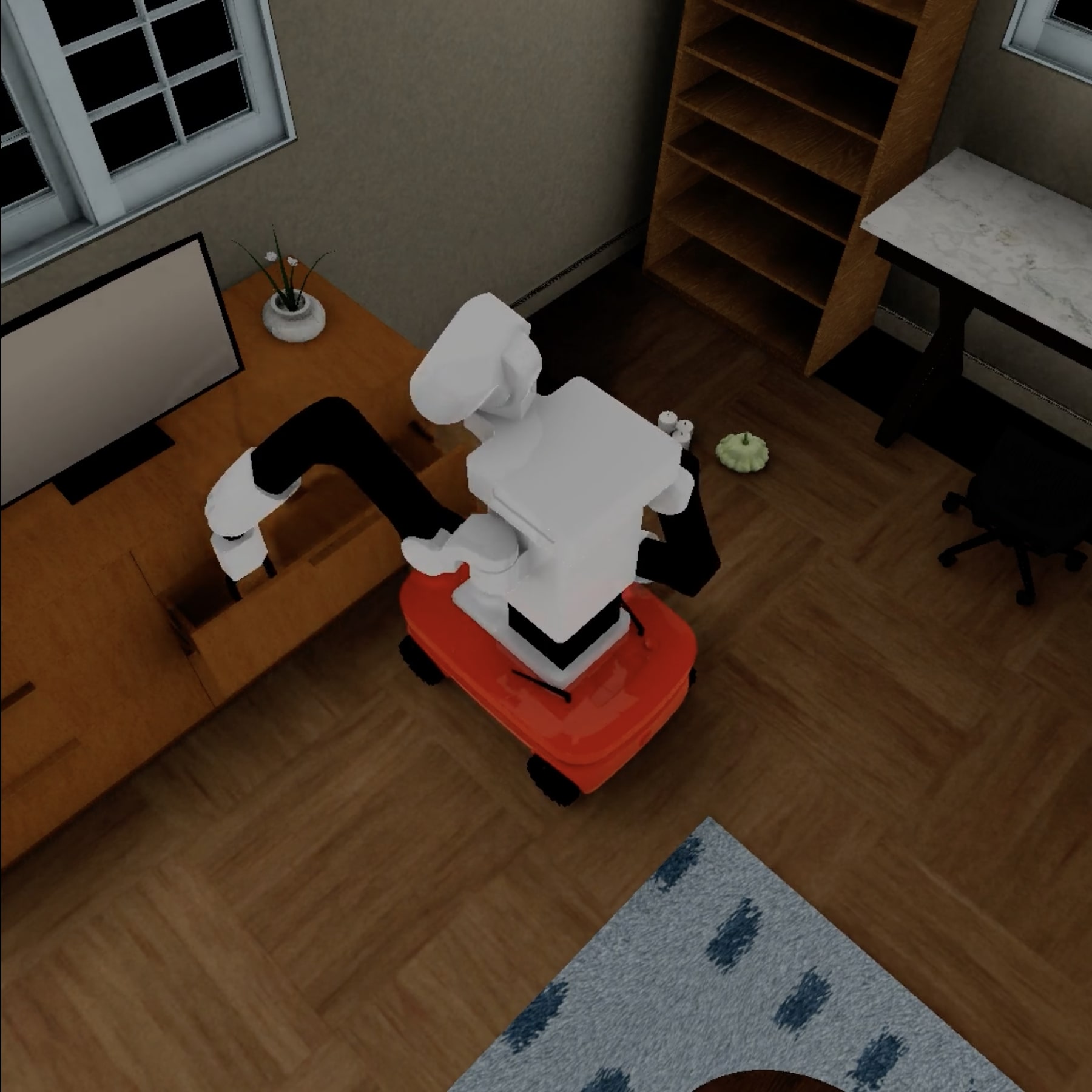}%
\hfill%
\includegraphics[trim=0 220 0 20,clip,width=0.33\linewidth]{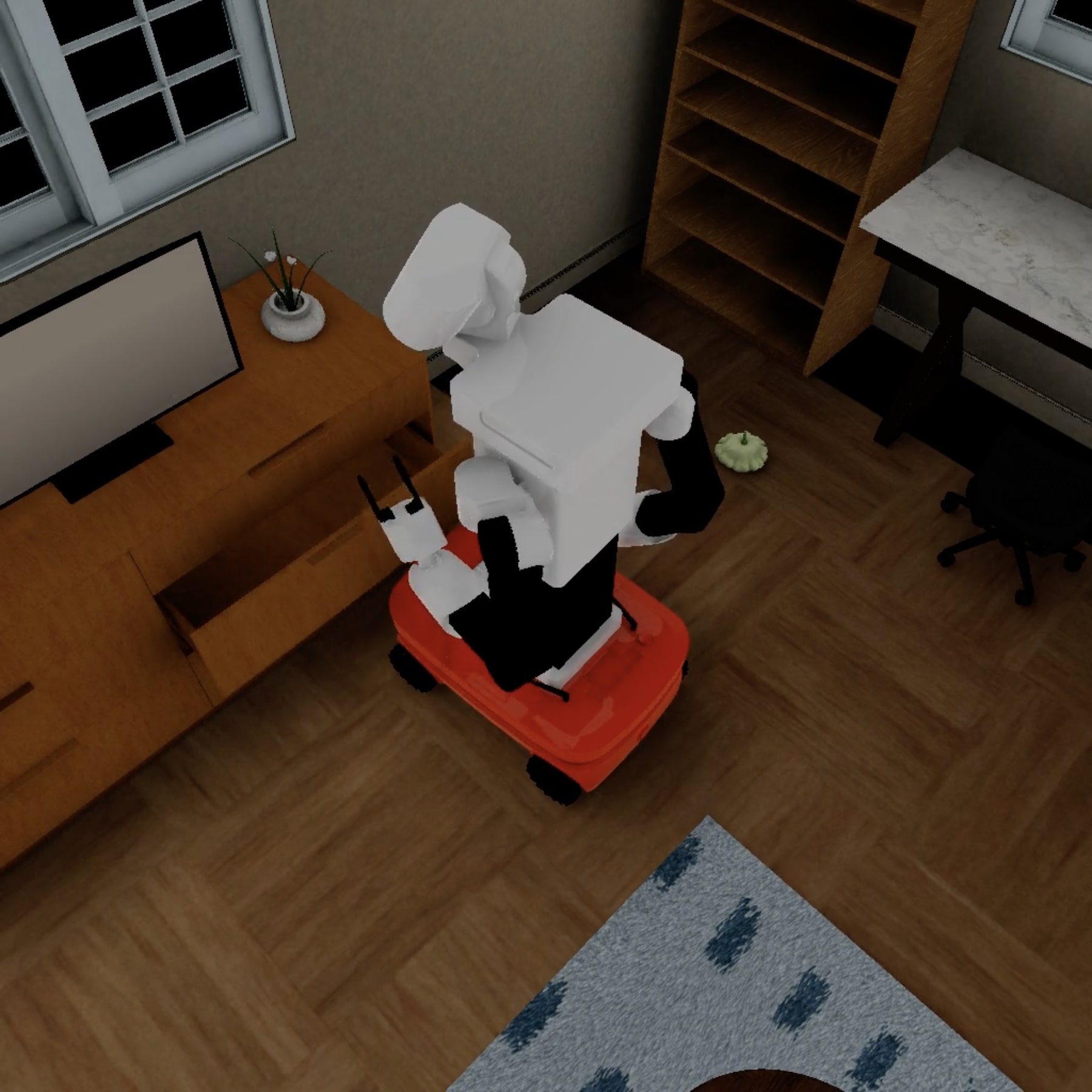}
}%
\vspace{-5pt}
\caption{\texttt{push} (open)}
\vspace{10pt}
\end{subfigure}
\begin{subfigure}[t]{0.8\textwidth}{\centering
\includegraphics[trim=0 200 0 200,clip,width=0.33\linewidth]{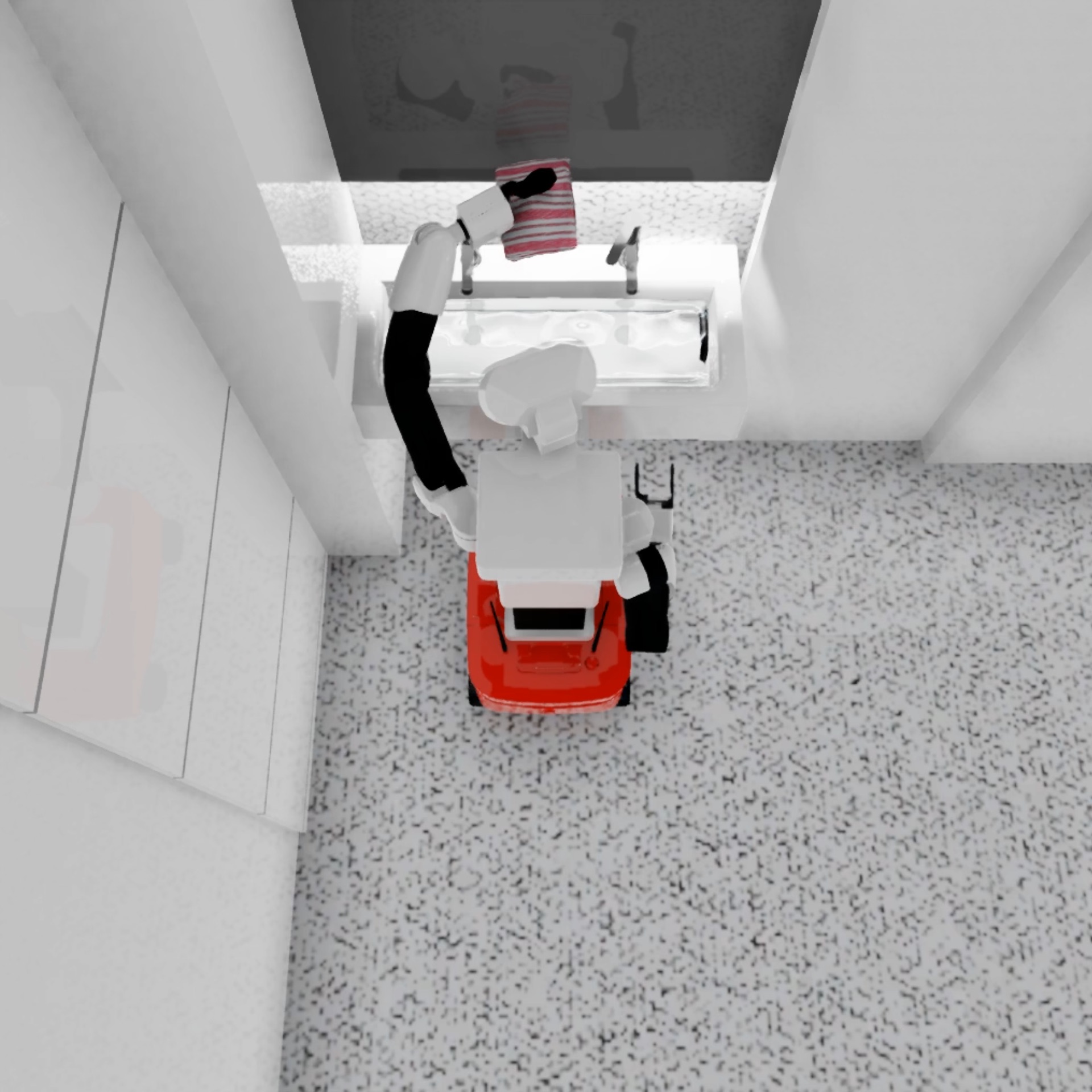}%
\hfill%
\includegraphics[trim=0 200 0 200,clip,width=0.33\linewidth]{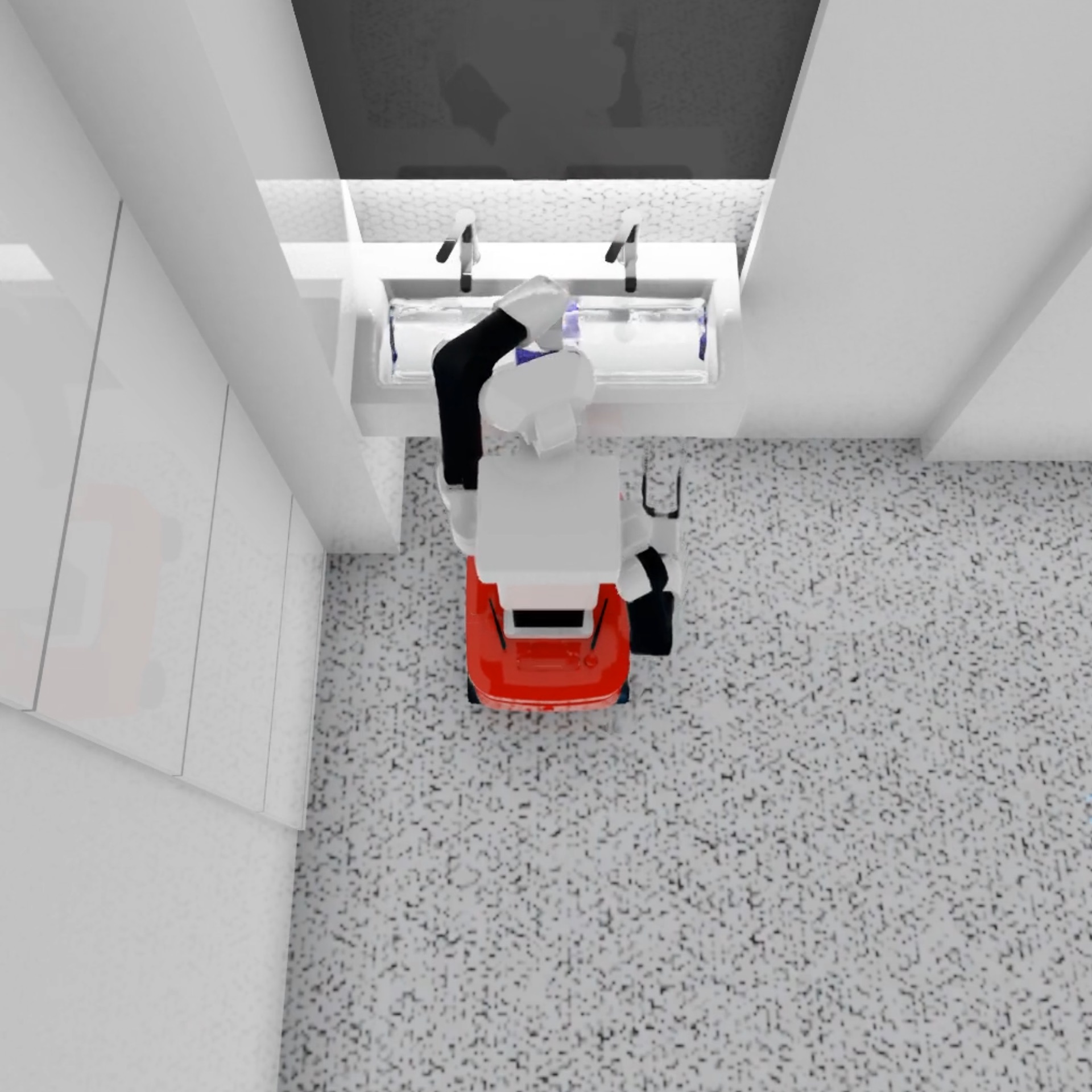}%
\hfill%
\includegraphics[trim=0 200 0 200,clip,width=0.33\linewidth]{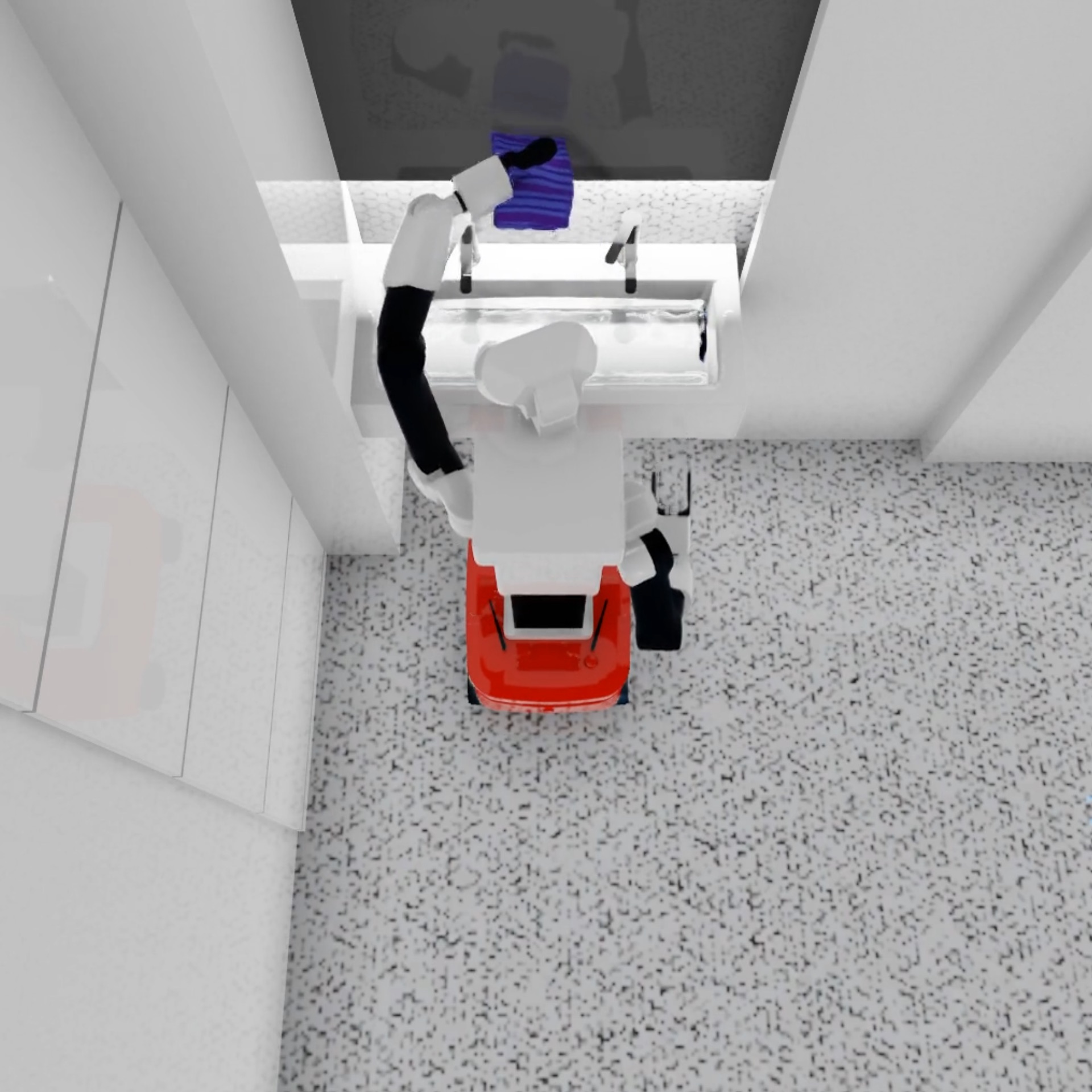}
}%
\vspace{-5pt}%
\caption{\texttt{dip}}%
\vspace{10pt}%
\end{subfigure}
\begin{subfigure}[t]{0.8\textwidth}{\centering%
\includegraphics[trim=0 200 0 200,clip,width=0.33\linewidth]{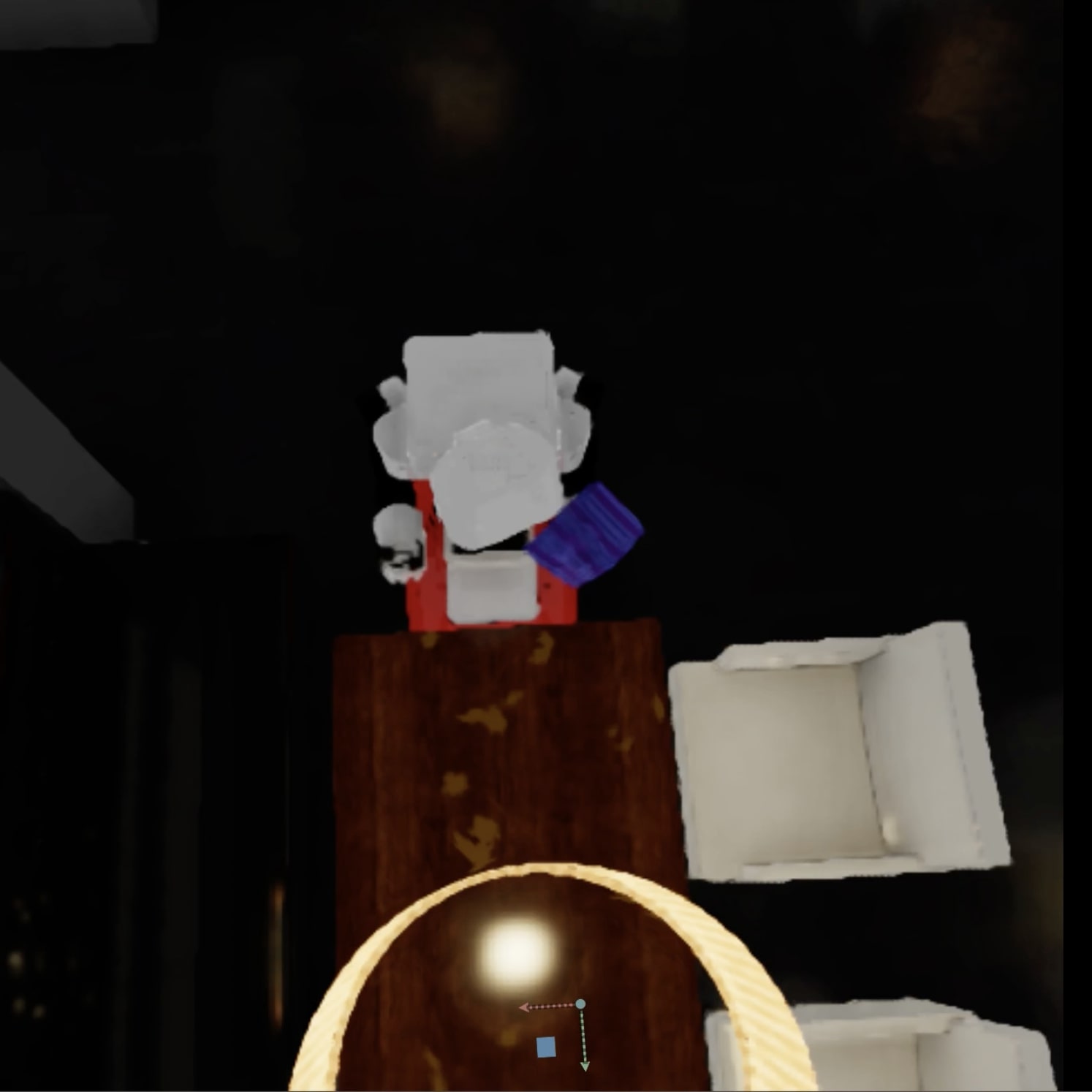}%
\hfill%
\includegraphics[trim=0 200 0 200,clip,width=0.33\linewidth]{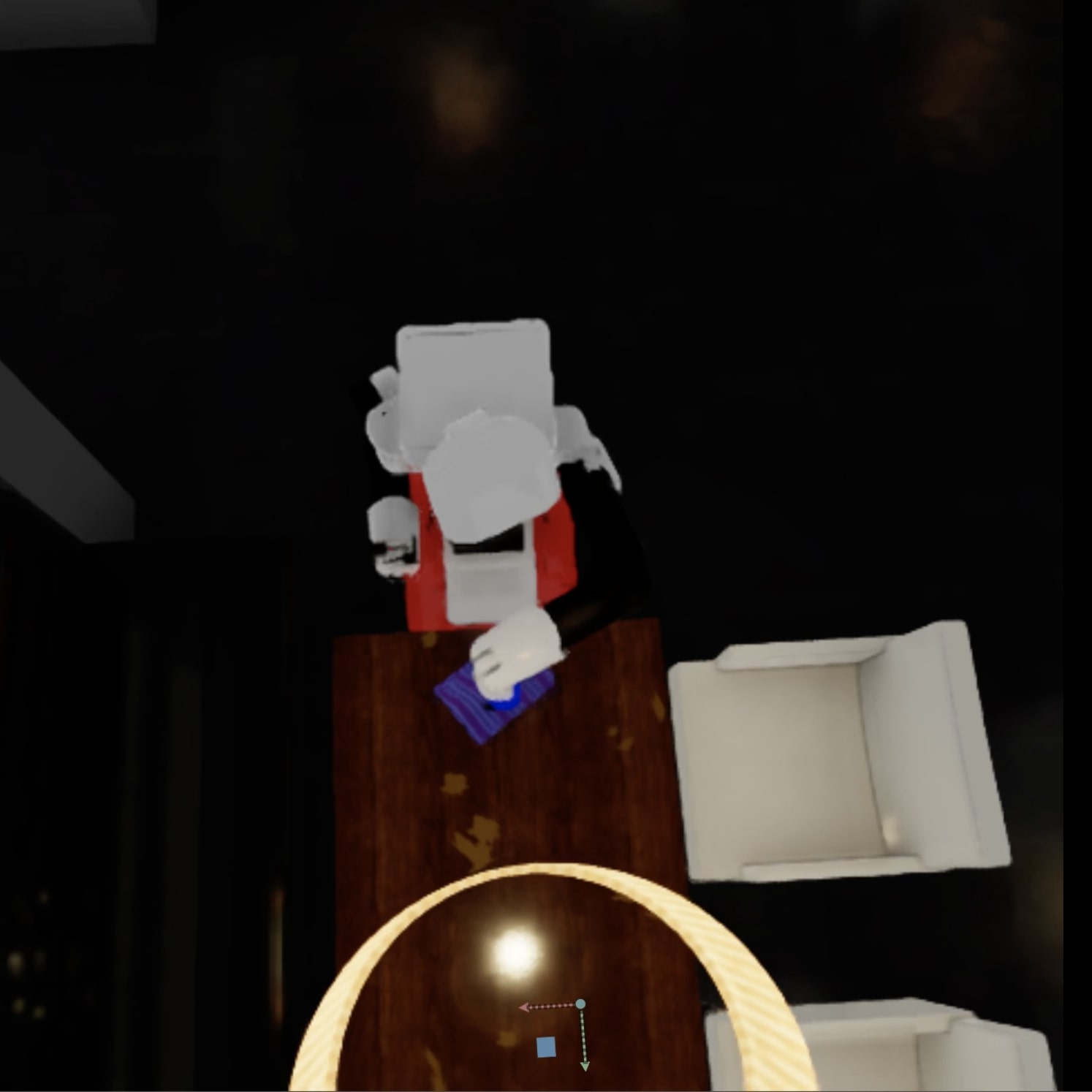}%
\hfill%
\includegraphics[trim=0 200 0 200,clip,width=0.33\linewidth]{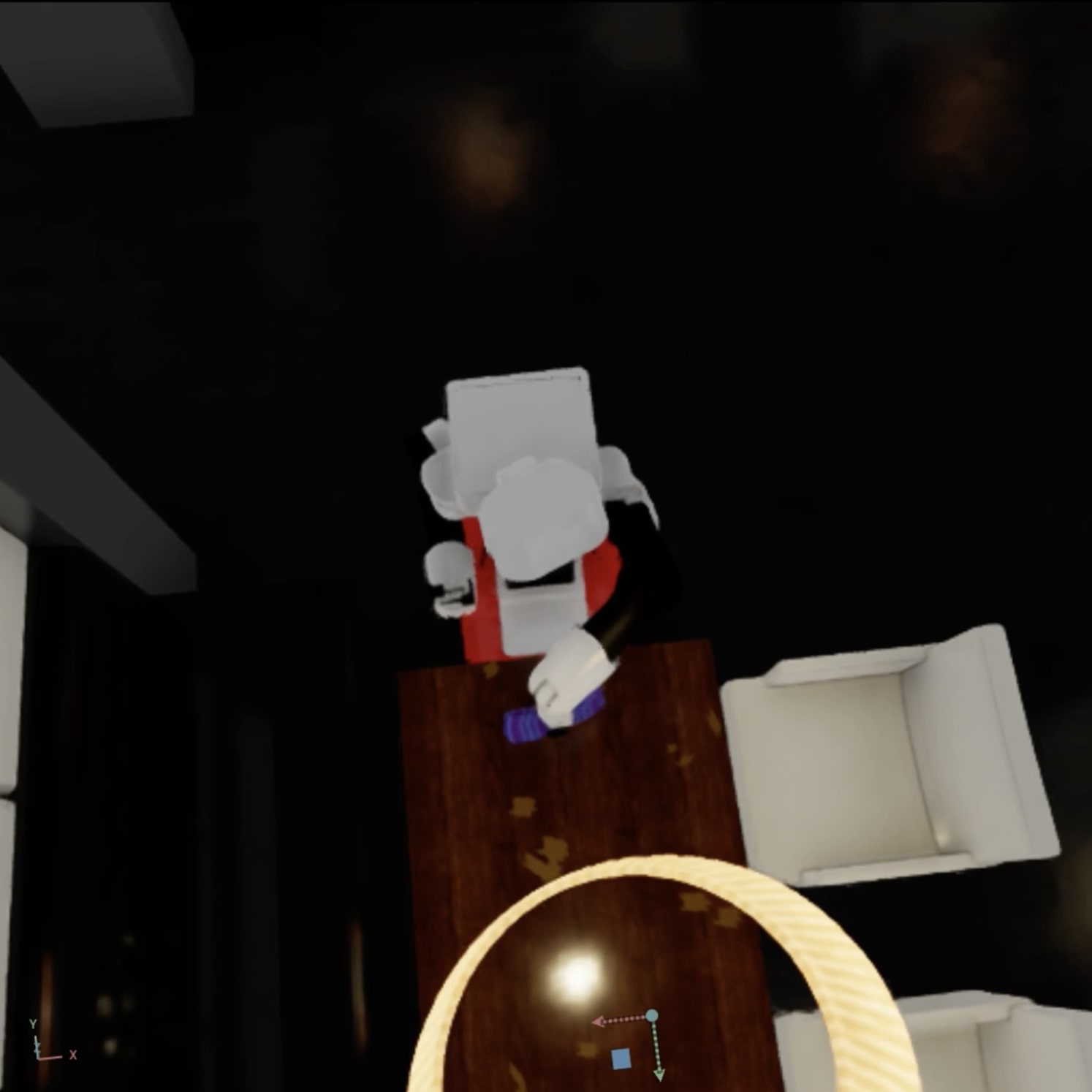}
}%
\vspace{-5pt}%
\caption{\texttt{wipe}}%
\vspace{10pt}%
\end{subfigure}
\caption{Overview of all the high-level action primitives with the pre-conditions (left), post-conditions (right), and intermediate states (middle) during execution.
}
\label{fig:appendix_high_level_planner}
\end{figure}

The primitives we used in our experiments can be seen in Fig.~\ref{fig:appendix_high_level_planner} and include:
\vspace{-2mm}
\begin{itemize}[leftmargin=2em]
\item[a)] \texttt{navigate}: Collision-free trajectory of the entire robot to a location.
\item[b)] \texttt{pick}: Composition of 1) a trajectory to a pre-grasp \texttt{6D\_pose} above the object to pick, 2) a \texttt{line} trajectory in Cartesian space down towards the object, interrupted when there is contact, 3) a \texttt{closing} action, 4) a first retracting trajectory following a \texttt{line} upwards, and 5) a second retracting trajectory to reach the untucked joint configuration. The primitive only executes if the robot is not currently picking another object.
\item[c)] \texttt{place}: Composition of 1) a trajectory to a pre-place \texttt{6D\_pose} above the object to place the grasped object on, 2) an \texttt{opening} action, 3) a second retracting trajectory to reach the untucked joint configuration. The primitive only executes if the robot is currently holding an object.
\item[d)] \texttt{push}: Composition of 1) a trajectory to a pre-push \texttt{6D\_pose} above the object to push-open, 2) a \texttt{line} trajectory in Cartesian space down towards the object, interrupted if there is contact, 3) a \texttt{line} trajectory in Cartesian space to push-open the object, e.g., towards the robot, 4) a first retracting trajectory following a \texttt{line} upwards, and 5) a second retracting trajectory to reach the untucked joint configuration. 
\item[e)] \texttt{dip}: Composition of 1) a trajectory to a pre-dip \texttt{6D\_pose} above the object to dip into, 2) a \texttt{line} trajectory in Cartesian space down towards the object to dip, 3) a \texttt{line} trajectory in Cartesian space upwards, and 4) a retracting trajectory to reach the initial joint configuration. The primitive only executes if the robot is currently holding an object.
\item[f)] \texttt{wipe}: Composition of 1) a trajectory to a pre-wipe \texttt{6D\_pose} above the object to wipe, 2) a \texttt{line} trajectory in Cartesian space down towards the object to be wiped, interrupted when there is contact, 3) a \texttt{line} trajectory in Cartesian space horizontally to wipe left-right, 4) a \texttt{line} trajectory in Cartesian space horizontally to wipe towards the robot, 4) a first retracting trajectory following a \texttt{line} upwards, and 5) a second retracting trajectory to reach the untucked joint configuration. The primitive only executes if the robot is currently holding an object.
\end{itemize}

\subsection{Additional Metrics: success score Q}
The success score Q \cite{srivastava2021behavior} for each task is reported in Table~\ref{tab:baseline_q_score}. We observe the same trend in Q as in the success rate. This is another indication of the effectiveness of our approach.
\begin{table}[h!]
    \centering
    \resizebox{\textwidth}{!}{\begin{tabular}{l|cc|ccc}
    \toprule
    \multicolumn{1}{c|}{\multirow{2}{*}{Method}} & \multicolumn{2}{c|}{Policy Features} & \multicolumn{3}{c}{Success score Q} \\
    \multicolumn{1}{c|}{} & Primitives & History & \texttt{StoreDecoration} & \texttt{CollectTrash} & \texttt{CleanTable} \\ \midrule
    \rlvmc & \textcolor{red}\faTimes & \textcolor{red}\faTimes & $0.0 \pm 0.0$ & $0.0 \pm 0.0$ & $0.0 \pm 0.0$ \\
    \rlprim & \textcolor{indiagreen}\faCheck & \textcolor{red}\faTimes & $0.50 \pm 0.05$ & $0.49 \pm 0.04$ & $0.77 \pm 0.08$ \\
    \rlprimhist & \textcolor{indiagreen}\faCheck & \textcolor{indiagreen}\faCheck & $0.59 \pm 0.03$ & $0.68 \pm 0.02$  & $0.88 \pm 0.02$ \\ \bottomrule
    \end{tabular}
    }
    \caption{Performance of three baseline methods, in terms of the success score Q.}
    \label{tab:baseline_q_score}
\end{table}

\section{Details of the Real-World Setup}
\label{as:s2r}

\paragraph{Scene.} Our real-world experiments take place in a mockup studio-apartment in our lab that includes a bedroom, living room, and dining area. 
This real-world scene was modeled with a digital counterpart in simulation for \benchmark: the \texttt{mockup\_apt} scene. Fig.~\ref{fig:sim2real_setup} depicts both real and digital twin side-by-side. The digital model has been created by first, scanning the real-world apartment using a phone and commercial software~\cite{scaniverse}, then, replacing and projecting the texture of walls, floors, and ceilings, and finally, replacing the existing objects with 3D models from our \dataset. This process should minimize the effect caused by differences between the real world and the 3D model. The simulated and the real robot use the same 2D map of the scene for localization. In this way, 2D locations in the real world correspond to the same 2D locations in simulation. 

\paragraph{Robot platform.} We use in our experiments a Tiago++ model from PAL Robotics, with an omnidirectional base, two 7-degrees-of-freedom arms with parallel-yaw grippers, a 1-degree-of-freedom prismatic torso, two SICK LiDAR sensors (back and front of the base), and an ASUS Xtion RGB-D camera mounted on the robot's head, which can be controlled in yaw and pitch.
All sensors and actuators are connected through the Robot Operating System, ROS~\cite{quigley2009ros}. The code runs on a laptop with an Nvidia GTX 1070 that sends the commands to the onboard robot computer to be executed.

\paragraph{Action primitives on the real robot.} We implemented a version of the action primitives \texttt{navigate}, \texttt{pick} and \texttt{place} on the real robot similar to their implementation in \simulator. For navigation, we use a 2D sampling-based motion planner on the 2D map we use for localization and that has been created from the 3D model of the apartment. For manipulation in the real world, we need to obtain the parameters necessary to plan the manipulation primitives (\texttt{pick} and \texttt{place}) from the sensor signals. To do that, we first obtain detections from a YOLO v3~\cite{redmon2018yolov3} object detector on the RGB images from the robot's camera. If the robot attempts to execute an action primitive on an object that has not been detected, we return failure (\textit{Object Detection Failure} in our analysis). If the object to manipulate has been detected, we query the pixels in the depth map corresponding to the 21$\times$21 window at the center of the detection bounding box, and use this information to compute the centroid of the corresponding 3D points. This location will be used as an object location for grasping or placing. This information is enough to create a sequence of paths using a sampling-based motion planner (RRT~\cite{kuffner2000rrt}) similar to the ones used in simulation. The motion planner uses a voxel representation of the scene obtained from the depth sensor.

\paragraph{Experimental setup.} We randomized three parameters in our real robot experiments: the initial location of the robot (chosen from the three possible activity locations), the positions of the objects on the table, and the orientations of the objects (upright or lying down). The experiment runs cover uniformly these parameters. Additionally, for vision-based policies, we evaluated two lighting conditions: full lighting (including the ceiling) and only lamps.
Failures are defined as follows:
Motion planning failures occurred when the robot was unable to plan an end effector trajectory to complete the action primitive. This includes failures caused by navigation noise, where after navigating to the task location, the robot was too far away from the target object to execute the pick or place action. 
Grasping failures include errors that occurred during the execution of the grasp such as pushing the object off the table while attempting a grasp or losing the object because it slips out, as well as the robot dropping the object after grasping it.
Placing failures include the robot dropping the object in the wrong location. 
Object detection failures primarily resulted from our object detector, YOLO v3~\cite{redmon2018yolov3}, failing to detect the object to interact with; a second, less common, object detection failure corresponds to the depth camera returning invalid measurements for the detected object. 
Finally, we consider policy failures when the policy selects the same invalid action three times in a row, e.g., requesting to navigate to the bin when it is already in front of the bin. The main policy error corresponds to the policy repeatedly choosing the same action primitive. We observed empirically that, even though the policy selection presents some small randomness when the policy requests the same invalid action three times in a row, most subsequent calls will be also the same invalid action and we decide to terminate early.

\subsection{Further Characterization of the Sim-Real Gap}

\begin{figure}[h!]
    \centering
    \includegraphics[width=0.31\linewidth]{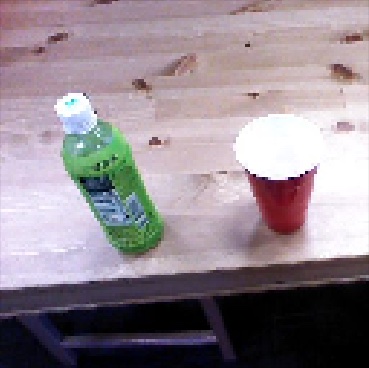}
    \hfill
    \includegraphics[width=0.31\linewidth]{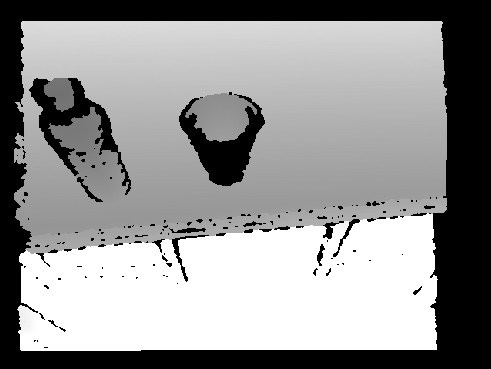}
    \hfill
    \includegraphics[width=0.31\linewidth]{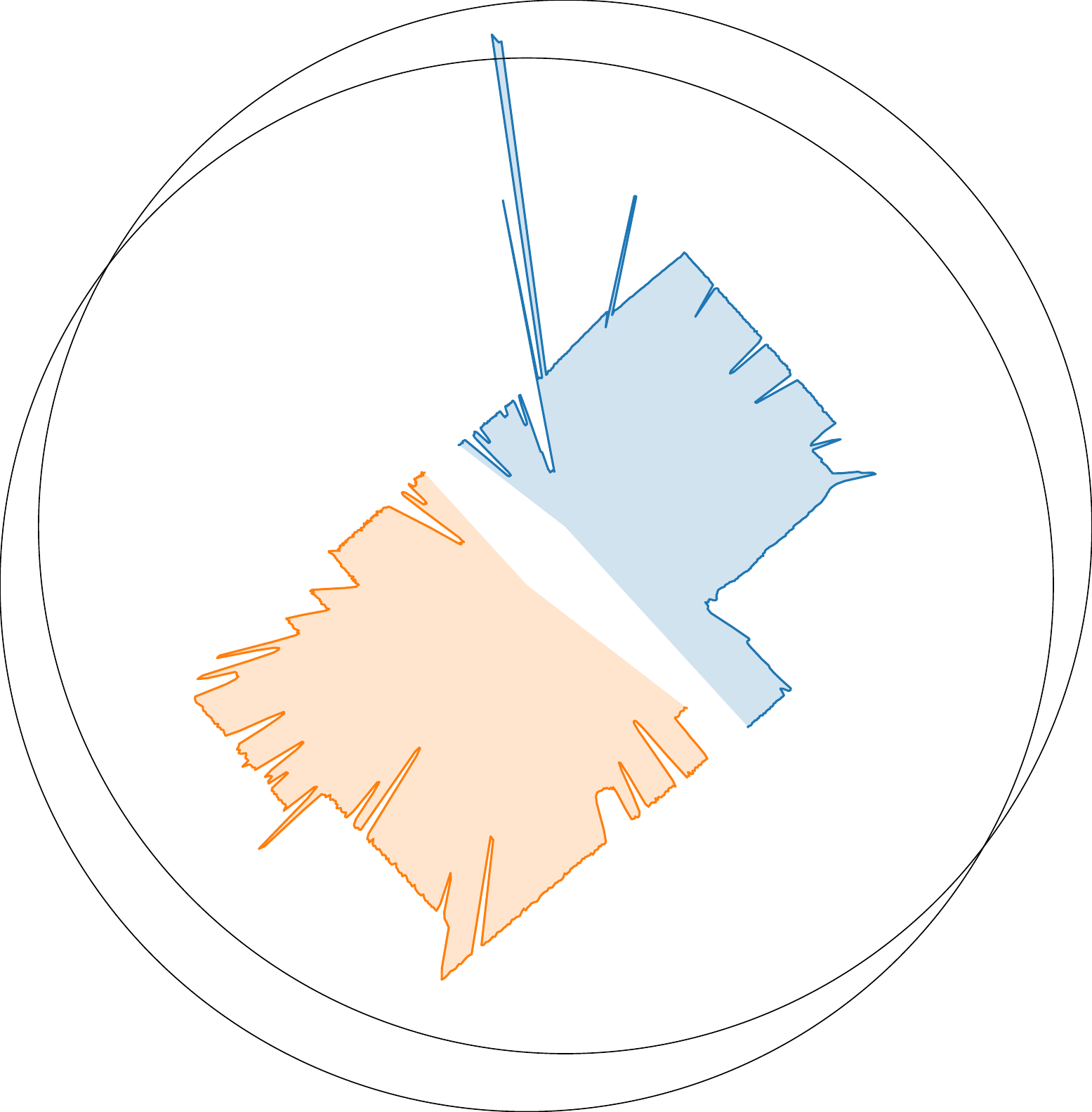}
    \\\vspace{3pt}
    \includegraphics[width=0.31\linewidth]{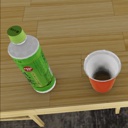}
    \hfill
    \includegraphics[width=0.31\linewidth]{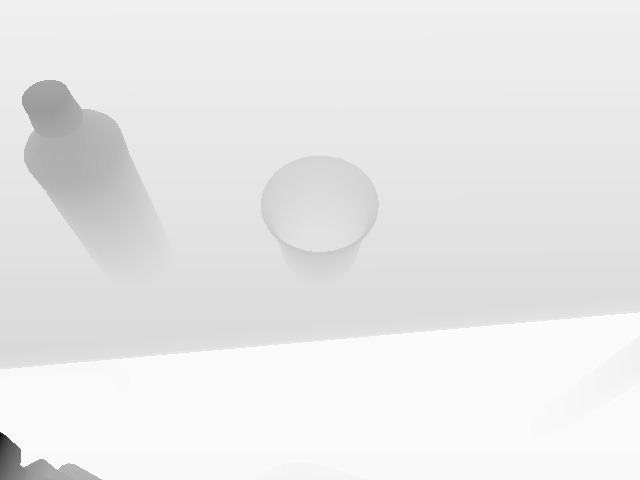}
    \hfill
    \includegraphics[width=0.31\linewidth]{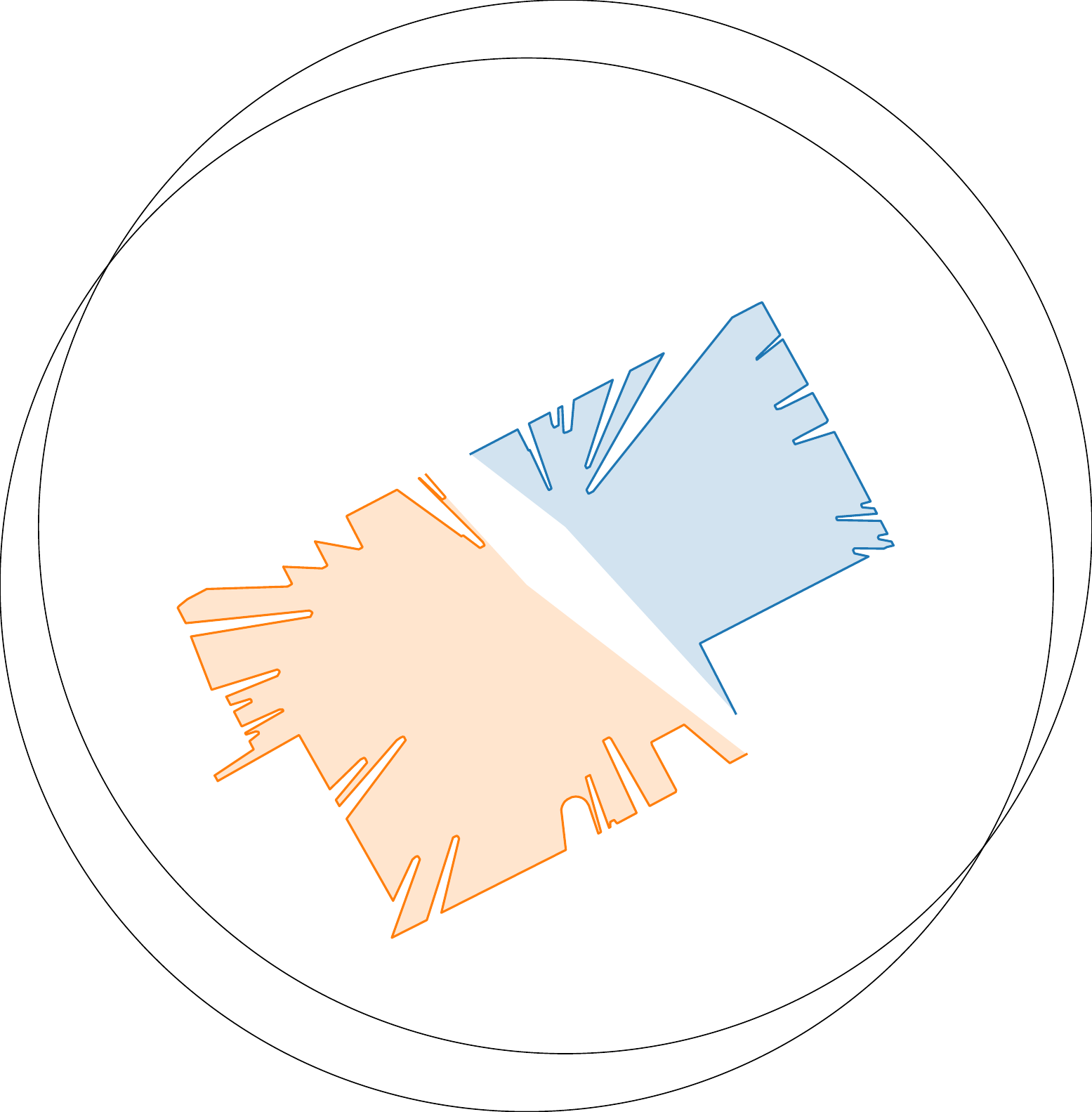}
    \\\vspace{10pt}
    \includegraphics[width=0.31\linewidth]{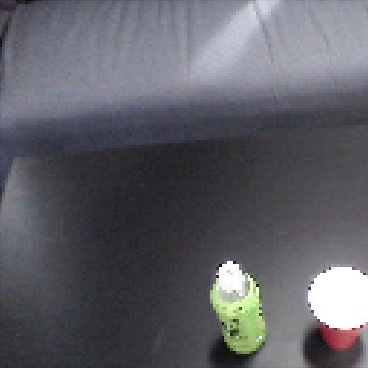}
    \hfill
    \includegraphics[width=0.31\linewidth]{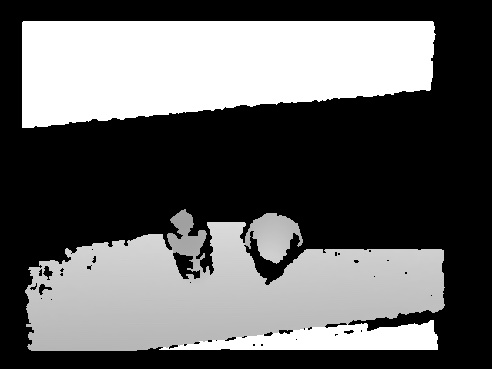}
    \hfill
    \includegraphics[width=0.31\linewidth]{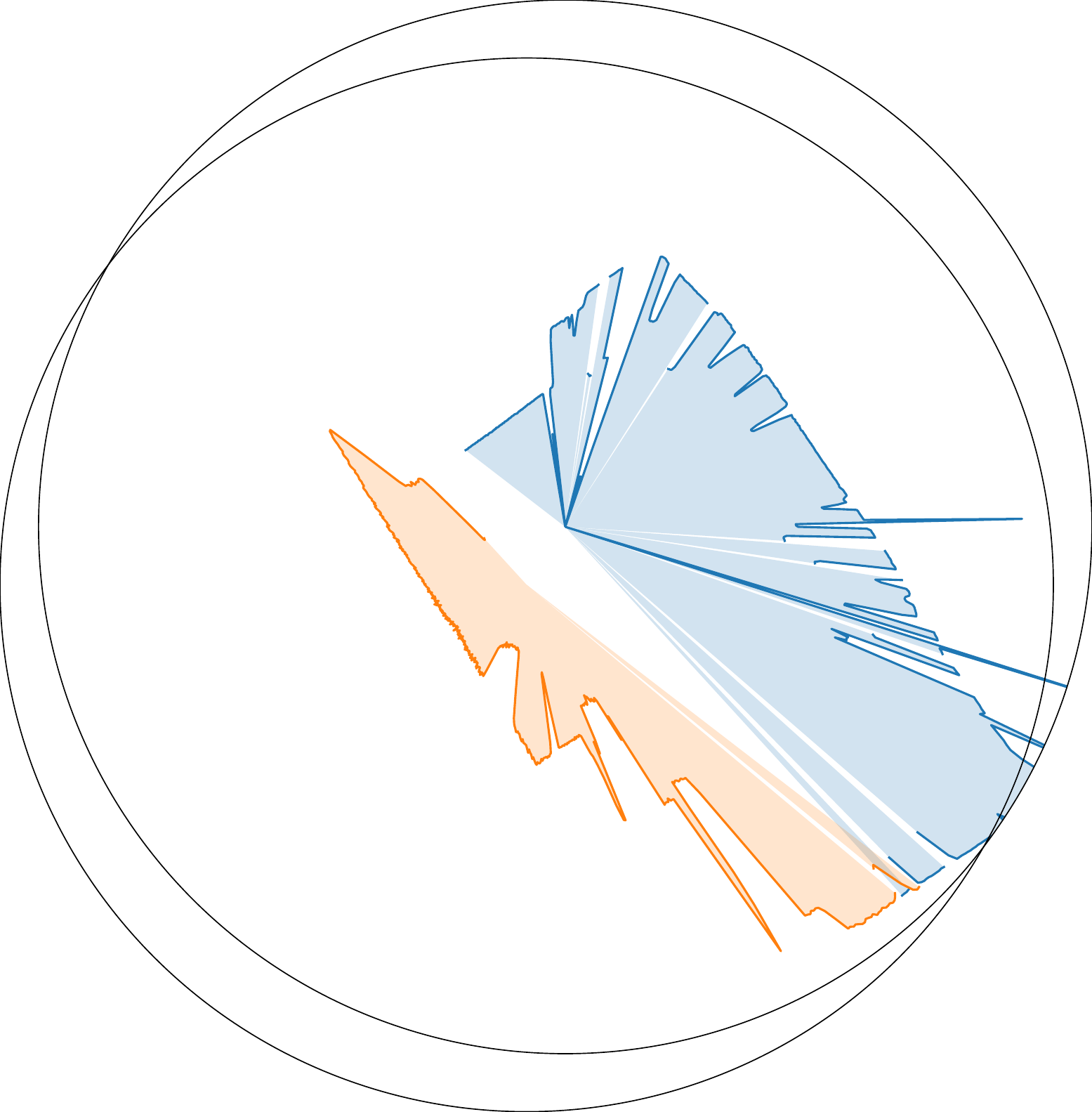}
    \\\vspace{3pt}
    \includegraphics[width=0.31\linewidth]{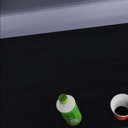}
    \hfill
    \includegraphics[width=0.31\linewidth]{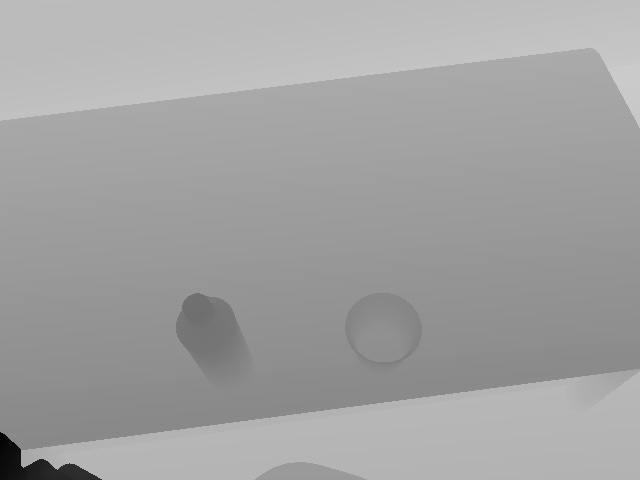}
    \hfill
    \includegraphics[width=0.31\linewidth]{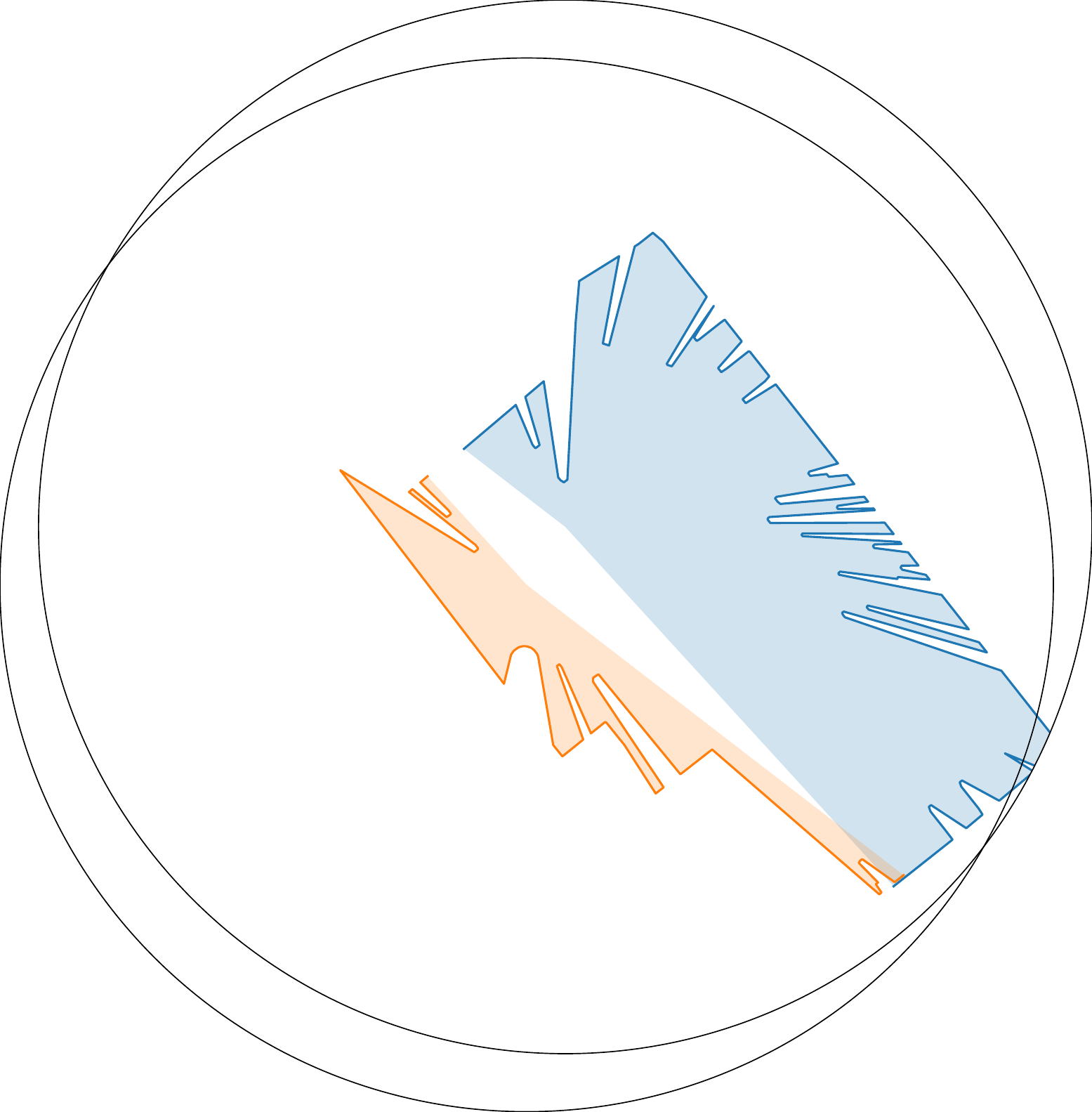}
    \caption{\revision{\textbf{Comparison between real and simulated sensor signals.} RGB, Depth and LiDAR signals from the real-world sensors (first and third rows) and from the simulated sensors (second and fourth rows) at two different locations of the \texttt{mockup\_apt} scene (location 1: first two rows, location 2: last two rows). We observed a smaller sensor sim-real gap for LiDAR than for RGB and Depth: RGB is heavily influenced by lighting conditions and camera settings while Depth has difficulty in capturing reflective surfaces.}}
    \label{fig:simrealsensors}
\end{figure}

\revision{We performed additional experiments to characterize the gap between simulation and the real world. In our experiments, we moved the robot to different locations in the \texttt{mockup\_apt} scene and collected real sensor signals: RGB images, depth maps, and LiDAR measurements. We then moved the robot in simulation to the same locations and collected virtual sensor signals. Some of the images are depicted in Fig.~\ref{fig:simrealsensors}. We observe that, thanks to our highly realistic models, \simulator provides high-fidelity sensor signals that approximate the real-world ones. However, some of the sources of noise in the real sensors are not modeled currently in \simulator, contributing to a sim-real gap, e.g., the poor dynamic range of the RGB camera, or the ``shadow'' effects in depth maps due to the projected light mechanism. This analysis indicates possible avenues to further close the sensor gap between simulation and the real world, e.g., by including sensor noise models in \simulator or by leveraging sim-to-real techniques (e.g., domain randomization, system identification).}

\subsection{Additional Experiments}
\revision{
In addition to \rlprimhist, we also evaluated \rlvmc and \rlprimhist in the real world and observed a similar trend of performance in the real world as in simulation: \rlvmc < \rlprim < \rlprimhist. \rlvmc still achieves zero success because of sparse reward and exploration difficulty during training. \rlprim has worse performance than \rlprimhist: on average, the robot with \rlprimhist successfully places 0.64 cups/bottles, whereas the one with \rlprim places 0. Qualitatively, \rlprim tends to get stuck in a repetitive action loop because the agent is unaware of its action history. This preliminary result offers us some confidence that \simulator can be a reliable test bed for future sim-to-real robotics research.
}

\section{Ethical Statement}

\benchmark aims to drive embodied AI solutions that fulfill human need. A primary ethical consideration of \benchmark is therefore the method used to determine such need. To understand which activities would be useful for humans, we survey 1,461 respondents on Amazon Mechanical Turk and collected 50 responses for each activity, aiming for a clear consensus signal. Though this is a large group, the results are still biased by the fact that the researchers, annotators, and data providers only represent a small population relative to potential users of such technology. We plan to address these biases by making \dataset open-source and invite a wider community to contribute to its knowledge base. 

Furthermore, compared to the U.S. population, the demographic representation in our survey skews white, male, non-disability status, mid-five figure incomes, and 30-40 age range with more representation on the higher side than the lower side of that range\textemdash closer to the demographics of Mechanical Turk workers. This also biases the survey population toward certain responses, creating potential ethical limitations. 

Specifically this bias may affect the survey's representation of people who would be affected most by autonomous agents~\cite{mckinsey2019automation}. We therefore asked several explicit questions such as ``Do you do household work for a living?'', ``If you do household work for a living, would you benefit from assistance?'',  ``Do you pay for someone else to do household work for you?''. Future work involves using these questions to direct benchmark development.


\newpage
\definecolor{main-color}{rgb}{0,0,0}
\definecolor{back-color}{rgb}{0.942, 0.942, 0.93}
\definecolor{string-color}{rgb}{0.3333, 0.5254, 0.345}
\definecolor{key-color}{rgb}{0.847, 0.1059, 0.3765}
\definecolor{pred-color}{rgb}{0.1176, 0.5333, 0.898}
\definecolor{term-color}{rgb}{0., 0.31, 0.251}

\lstdefinestyle{mystyle}
{
    language = C,
    basicstyle = {\tiny \ttfamily \color{main-color}},
    backgroundcolor = {\color{back-color}},
    stringstyle = {\color{string-color}},
    keywordstyle = [2]{\color{pred-color}},
    keywordstyle = [3]{\color{key-color}},
    keywordstyle = {\color{term-color}},
    otherkeywords = {?},
    morekeywords = [2]{ontop, inside, inroom},
    morekeywords = [3]{define, problem, domain, objects, init, goal, and, for_n_pairs, forall, :, exists, and},
}

\noindent\begin{minipage}[t]{0.85\textwidth}
\begin{lstlisting}[style = mystyle,
  caption={{\texttt{BakingSugarCookies}}}, label={lst:bddl3}]
(define (problem baking_sugar_cookies-0)
    (:domain omnigibson)

    (:objects
        flour.n.01_1 - flour.n.01
        granulated_sugar.n.01_1 - granulated_sugar.n.01
        raw_egg.n.01_1 - raw_egg.n.01
        vanilla.n.02_1 - vanilla.n.02
        melted__butter.n.01_1 - melted__butter.n.01
        sodium_carbonate.n.01_1 - sodium_carbonate.n.01
        baking_powder.n.01_1 - baking_powder.n.01
        electric_mixer.n.01_1 - electric_mixer.n.01
        mixing_bowl.n.01_1 - mixing_bowl.n.01
        sugar_cookie.n.01_1 sugar_cookie.n.01_2 sugar_cookie.n.01_3
        sugar_cookie.n.01_4 sugar_cookie.n.01_5 sugar_cookie.n.01_6 - sugar_cookie.n.01
        oven.n.01_1 - oven.n.01
        cookie_sheet.n.01_1 - cookie_sheet.n.01
        flour__sack.n.01_1 - flour__sack.n.01
        sugar__sack.n.01_1 - sugar__sack.n.01
        sodium_carbonate__jar.n.01_1 - sodium_carbonate__jar.n.01
        baking_powder__jar.n.01_1 - baking_powder__jar.n.01
        mason_jar.n.01_1 - mason_jar.n.01
        countertop.n.01_1 countertop.n.01_2 - countertop.n.01
        vanilla__bottle.n.01_1 - vanilla__bottle.n.01
        salt.n.02_1 - salt.n.02
        salt__shaker.n.01_1 - salt__shaker.n.01
        bowl.n.01_1 - bowl.n.01
        electric_refrigerator.n.01_1 - electric_refrigerator.n.01
        tablespoon.n.02_1 - tablespoon.n.02
        floor.n.01_1 - floor.n.01
        agent.n.01_1 - agent.n.01
    )
    
    (:init 
        (filled flour__sack.n.01_1 flour.n.01_1)
        (ontop flour__sack.n.01_1 countertop.n.01_1)
        (insource salt__shaker.n.01_1 salt.n.02_1)
        (ontop salt__shaker.n.01_1 countertop.n.01_1)
        (ontop tablespoon.n.02_1 countertop.n.01_1)
        (filled sugar__sack.n.01_1 granulated_sugar.n.01_1)
        (ontop sugar__sack.n.01_1 countertop.n.01_1)
        (inside raw_egg.n.01_1 bowl.n.01_1)
        (insource vanilla__bottle.n.01_1 vanilla.n.02_1)
        (ontop vanilla__bottle.n.01_1 countertop.n.01_1)
        (filled mason_jar.n.01_1 melted__butter.n.01_1)
        (ontop mason_jar.n.01_1 countertop.n.01_1)
        (filled sodium_carbonate__jar.n.01_1 sodium_carbonate.n.01_1)
        (ontop sodium_carbonate__jar.n.01_1 countertop.n.01_2)
        (filled baking_powder__jar.n.01_1 baking_powder.n.01_1)
        (ontop baking_powder__jar.n.01_1 countertop.n.01_2)
        (ontop electric_mixer.n.01_1 countertop.n.01_2)
        (attached mixing_bowl.n.01_1 electric_mixer.n.01_1)
        (inroom oven.n.01_1 kitchen)
        (inroom countertop.n.01_1 kitchen)
        (inroom countertop.n.01_2 kitchen)
        (inside bowl.n.01_1 electric_refrigerator.n.01_1)
        (inroom electric_refrigerator.n.01_1 kitchen)
        (ontop cookie_sheet.n.01_1 countertop.n.01_2)
        (inroom floor.n.01_1 kitchen) 
        (ontop agent.n.01_1 floor.n.01_1)
        (future sugar_cookie.n.01_4) 
        (future sugar_cookie.n.01_1) 
        (future sugar_cookie.n.01_2) 
        (future sugar_cookie.n.01_6) 
        (future sugar_cookie.n.01_3) 
        (future sugar_cookie.n.01_5) 
    )
    
    (:goal 
        (and 
            (real ?sugar_cookie.n.01_1) 
            (real ?sugar_cookie.n.01_2) 
            (real ?sugar_cookie.n.01_3) 
            (real ?sugar_cookie.n.01_4) 
            (real ?sugar_cookie.n.01_5) 
            (real ?sugar_cookie.n.01_6) 
            (forall 
                (?sugar_cookie.n.01 - sugar_cookie.n.01) 
                (and 
                    (cooked ?sugar_cookie.n.01) 
                    (ontop ?sugar_cookie.n.01 ?cookie_sheet.n.01_1)
                )
            )
        )
    )
)
\end{lstlisting}
\end{minipage}

\begin{minipage}[t]{.45\textwidth}
\begin{lstlisting}[style = mystyle, caption={{\texttt{CleanYourLaundryRoom}}}, label={lst:bddl4}]
(define 
  (problem clean_your_laundry_room_1)
  (:domain omnigibson)
    
  (:objects
    rag.n.01_1 - rag.n.01
    dryer.n.01_1 - dryer.n.01
    water.n.06_1 - water.n.06
    vinegar.n.01_1 - vinegar.n.01
    washer.n.03_1 - washer.n.03
    dust.n.01_1 - dust.n.01
    mold.n.05_1 - mold.n.05
    floor.n.01_1 - floor.n.01
    agent.n.01_1 - agent.n.01
  )
    
  (:init 
    (ontop rag.n.01_1 dryer.n.01_1) 
    (not 
      (covered water.n.06_1 rag.n.01_1)
    ) 
    (empty vinegar.n.01_1 washer.n.03_1) 
    (covered dust.n.01_1 dryer.n.01_1) 
    (covered mold.n.05_1 washer.n.03_1) 
    (onfloor agent.n.01_1 floor.n.01_1) 
    (filled water.n.06_1 bottle.n.01_1) 
    (onfloor bottle.n.01_1 floor.n.01_1)
    (inroom washer.n.03_1 laundry_room)
    (inroom floor.n.01_1 laundry_room)
  )
    
  (:goal 
    (and 
      (filled ?vinegar.n.01_1 ?washer.n.03_1) 
      (not 
        (covered ?dust.n.01_1 ?dryer.n.01_1)
      ) 
      (not 
        (covered ?mold.n.05_1 ?washer.n.03_1)
      )
    )
  )
)
\end{lstlisting}
\end{minipage}\hfill
\begin{minipage}[t]{.52\textwidth}
\begin{lstlisting}[style = mystyle, caption={{\texttt{CleanTheBottomOfAnIron}}}, label={lst:bddl5}]
(define (problem clean_the_bottom_of_an_iron-0)
    (:domain omnigibson)

    (:objects
        tarnish.n.02_1 - tarnish.n.02
        iron.n.04_1 - iron.n.04
        ironing_board.n.01_1 - ironing_board.n.01
        emery_paper.n.01_1 - emery_paper.n.01
        countertop.n.01_1 - countertop.n.01
        newspaper.n.03_1 - newspaper.n.03
        floor.n.01_1 - floor.n.01
        agent.n.01_1 - agent.n.01
    )
    
    (:init 
        (covered iron.n.04_1 tarnish.n.02_1)
        (toggled_on iron.n.04_1) 
        (ontop iron.n.04_1 ironing_board.n.01_1) 
        (ontop ironing_board.n.01_1 floor.n.01_1) 
        (ontop emery_paper.n.01_1 countertop.n.01_1)
        (ontop newspaper.n.03_1 ironing_board.n.01_1) 
        (inroom countertop.n.01_1 utility_room) 
        (inroom floor.n.01_1 utility_room) 
        (ontop agent.n.01_1 floor.n.01_1)
    )
    
    (:goal 
        (and
            (not 
                (covered ?iron.n.04_1 ?tarnish.n.02_1)
            )
        )
    )
)
\end{lstlisting}
\end{minipage}

\begin{minipage}[t]{.45\textwidth}
\begin{lstlisting}[style = mystyle, caption={{\texttt{PackingLunch}}}, label={lst:bddl1}]
(define 
  (problem packing_lunches_1)
  (:domain igibson)
    
  (:objects
    shelf.n.01_1 - shelf.n.01
    water.n.06_1 - water.n.06
    countertop.n.01_1 - countertop.n.01
    apple.n.01_1 - apple.n.01
    electric_refrigerator.n.01_1 - 
        electric_refrigerator.n.01
    hamburger.n.01_1 - hamburger.n.01
    basket.n.01_1 - basket.n.01
  )
    
  (:init 
    (ontop water.n.06_1 countertop.n.01_1)
    (inside apple.n.01_1 
        electric_refrigerator.n.01_1)
    (inside hamburger.n.01_1 
        electric_refrigerator.n.01_1)
    (ontop basket.n.01_1 countertop.n.01_1) 
    (inroom countertop.n.01_1 kitchen) 
    (inroom electric_refrigerator.n.01_1 
        kitchen) 
    (inroom shelf.n.01_1 kitchen)
  )
    
  (:goal 
    (and 
      (for_n_pairs
        (1)
        (?hamburger.n.01 - hamburger.n.01)
        (?basket.n.01 - basket.n.01)
        (inside ?hamburger.n.01 ?basket.n.01)
      )
      (for_n_pairs 
        (1) 
        (?basket.n.01 - basket.n.01) 
        (?water.n.06 - water.n.06) 
        (inside ?water.n.06 ?basket.n.01)
      ) 
      (for_n_pairs 
        (1) 
        (?basket.n.01 - basket.n.01) 
        (?apple.n.01 - apple.n.01) 
        (inside ?apple.n.01 ?basket.n.01)
      ) 
      (forall 
        (?basket.n.01 - basket.n.01) 
        (ontop ?basket.n.01 ?countertop.n.01_1) 
      )
    )
  )
)
\end{lstlisting}
\end{minipage}\hfill
\begin{minipage}[t]{.47\textwidth}
\begin{lstlisting}[style=mystyle, caption={{\texttt{ServingHorsDoeuvres}}}, label={lst:bddl2}]
(define 
  (problem serving_hors_d_oeuvres_1)
  (:domain igibson)

  (:objects
    tray.n.01_1 tray.n.01_2 - tray.n.01
    countertop.n.01_1 - countertop.n.01
    oven.n.01_1 - oven.n.01
    sausage.n.01_1 sausage.n.01_2 - sausage.n.01
    cherry.n.03_1 cherry.n.03_2 - cherry.n.03
    electric_refrigerator.n.01_1 - 
        electric_refrigerator.n.01
  )
    
  (:init 
    (ontop tray.n.01_1 countertop.n.01_1)
    (ontop tray.n.01_2 countertop.n.01_1)
    (inside sausage.n.01_1 oven.n.01_1)
    (inside sausage.n.01_2 oven.n.01_1)
    (inside cherry.n.03_1 
        electric_refrigerator.n.01_1)
    (inside cherry.n.03_2 
        electric_refrigerator.n.01_1)
    (inroom oven.n.01_1 kitchen)
    (inroom electric_refrigerator.n.01_1 
        kitchen)
    (inroom countertop.n.01_1 kitchen)
  )

  (:goal
    (and
      (exists
        (?tray.n.01 - tray.n.01)
        (and
          (forall
            (?sausage.n.01 - sausage.n.01)
            (ontop ?sausage.n.01 ?tray.n.01)
          )
          (forall
            (?cherry.n.03 - cherry.n.03)
            (not
              (ontop ?cherry.n.03 ?tray.n.01)
            )
          )
        )
      )
      (exists
        (?tray.n.01 - tray.n.01)
        (and
          (forall
            (?cherry.n.03 - cherry.n.03)
            (ontop ?cherry.n.03 ?tray.n.01)
          )
          (forall
            (?sausage.n.01 - sausage.n.01)
            (not
              (ontop ?sausage.n.01 ?tray.n.01)
            )
          )
        )
      )
    )
  )
)
\end{lstlisting}
\end{minipage}

\end{document}